\documentclass{article}

\PassOptionsToPackage{numbers, compress}{natbib}

\usepackage[eandd, final]{neurips_2026}

\usepackage[utf8]{inputenc} 
\usepackage[T1]{fontenc}    
\usepackage{hyperref}       
\usepackage{url}            
\usepackage{booktabs}       
\usepackage{amsfonts}       
\usepackage{nicefrac}       
\usepackage{microtype}     
\usepackage[table]{xcolor}        
\usepackage{graphicx}
\usepackage{pifont}
\usepackage{enumitem}
\usepackage{wrapfig}
\usepackage{multicol}
\usepackage{multirow}

\usepackage[most]{tcolorbox}
\tcbuselibrary{breakable, skins}
\usepackage{enumitem}

\newtcolorbox{promptbox}[1][]{
  breakable, enhanced,
  skin first=enhanced,
  skin middle=enhanced,
  skin last=enhanced,
  colback=gray!5, colframe=black!60,
  boxrule=0.4pt, arc=2pt,
  left=6pt, right=6pt, top=4pt, bottom=4pt,
  pad at break*=2mm,
  fonttitle=\bfseries\small,
  fontupper=\small,
  title={#1},
}

\definecolor{sparkleS}{HTML}{7CA7D8}  
\definecolor{sparkleP}{HTML}{7EAF5D}  
\definecolor{sparkleA}{HTML}{D16E6A}  
\definecolor{sparkleR}{HTML}{ECB575} 
\definecolor{sparkleK}{HTML}{7CA7D8} 
\definecolor{sparkleL}{HTML}{7EAF5D} 
\definecolor{sparkleE}{HTML}{E8746C}  

\newcommand{\sparklename}{%
  {\color{sparkleS}S}%
  {\color{sparkleP}p}%
  {\color{sparkleA}a}%
  {\color{sparkleR}r}%
  {\color{sparkleK}k}%
  {\color{sparkleL}l}%
  {\color{sparkleE}e}%
}

\definecolor{baitbox_red}{RGB}{255, 136, 120}    
\definecolor{baitbox_green}{RGB}{136, 229, 178}   
\definecolor{shallow_blue}{RGB}{220,235,250}  

\title{%
  \raisebox{-0.15\height}{\includegraphics[height=1.3em]{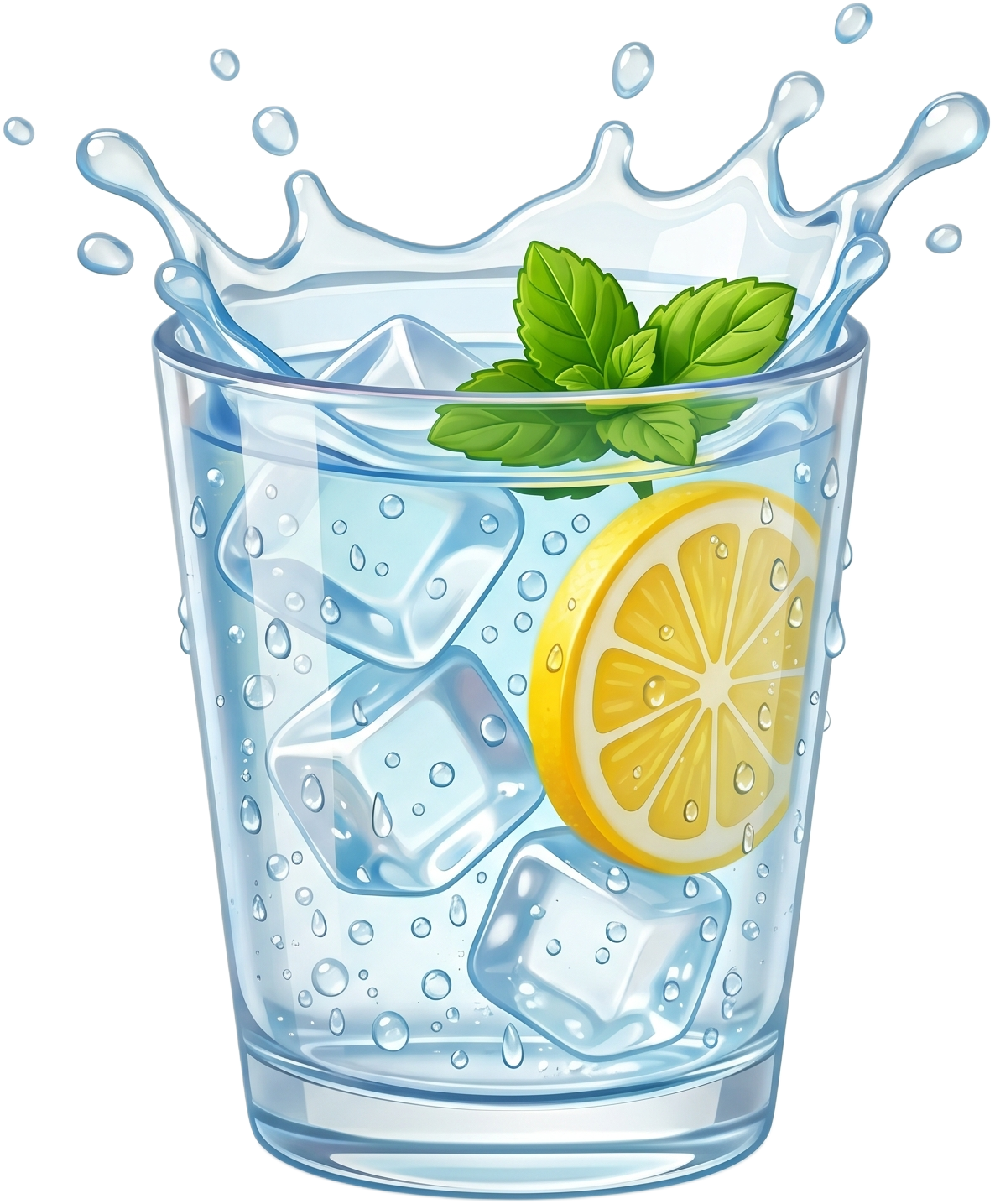}}%
  \hspace{0em}\sparklename: Realizing Lively Instruction-Guided Video Background Replacement via Decoupled Guidance%
}

\author{%
  Ziyun Zeng\quad
  Yiqi Lin\quad
  Guoqiang Liang\quad
  Mike Zheng Shou\textsuperscript{\ding{41}} \\
  Show Lab, National University of Singapore \\
  \small{\textsuperscript{\ding{41}}Corresponding Author}
}

\begin{document}

\maketitle

\begin{abstract}
In recent years, open-source efforts like Se\~{n}orita-2M~\cite{senorita2m} have propelled video editing toward natural language instruction. However, current publicly available datasets predominantly focus on local editing or style transfer, which largely preserve the original scene structure and are easier to scale. In contrast, \emph{Background Replacement}, a task central to creative applications such as film production and advertising, requires synthesizing entirely new, temporally consistent scenes while maintaining accurate foreground-background interactions, making large-scale data generation significantly more challenging. Consequently, this complex task remains largely underexplored due to a scarcity of high-quality training data. This gap is evident in poorly performing state-of-the-art models, \emph{e.g.,} Kiwi-Edit~\cite{kiwiedit}, because the primary open-source dataset that contains this task, \emph{i.e.,} OpenVE-3M~\cite{openve3m}, frequently produces static, unnatural backgrounds.
In this paper, we trace this quality degradation to \emph{a lack of precise background guidance} during data synthesis. Accordingly, we design a scalable pipeline that generates foreground and background guidance in a decoupled manner with strict quality filtering. Building on this pipeline, we introduce \emph{Sparkle}, a dataset of $\sim$140K video pairs spanning five common background-change themes, alongside \emph{Sparkle-Bench}, the largest evaluation benchmark tailored for background replacement to date. Experiments demonstrate that our dataset and the model trained on it achieve substantially better performance than all existing baselines on both OpenVE-Bench and Sparkle-Bench. Our proposed dataset, benchmark, model, as well as the full data generation pipeline, are fully open-sourced at \url{https://showlab.github.io/Sparkle/}.
\end{abstract}
\vspace{-1em}
\section{Introduction}
\vspace{-0.5em}
\label{sec:introduction}

\begin{figure}[t]
    \centering
    \includegraphics[width=1.\columnwidth]{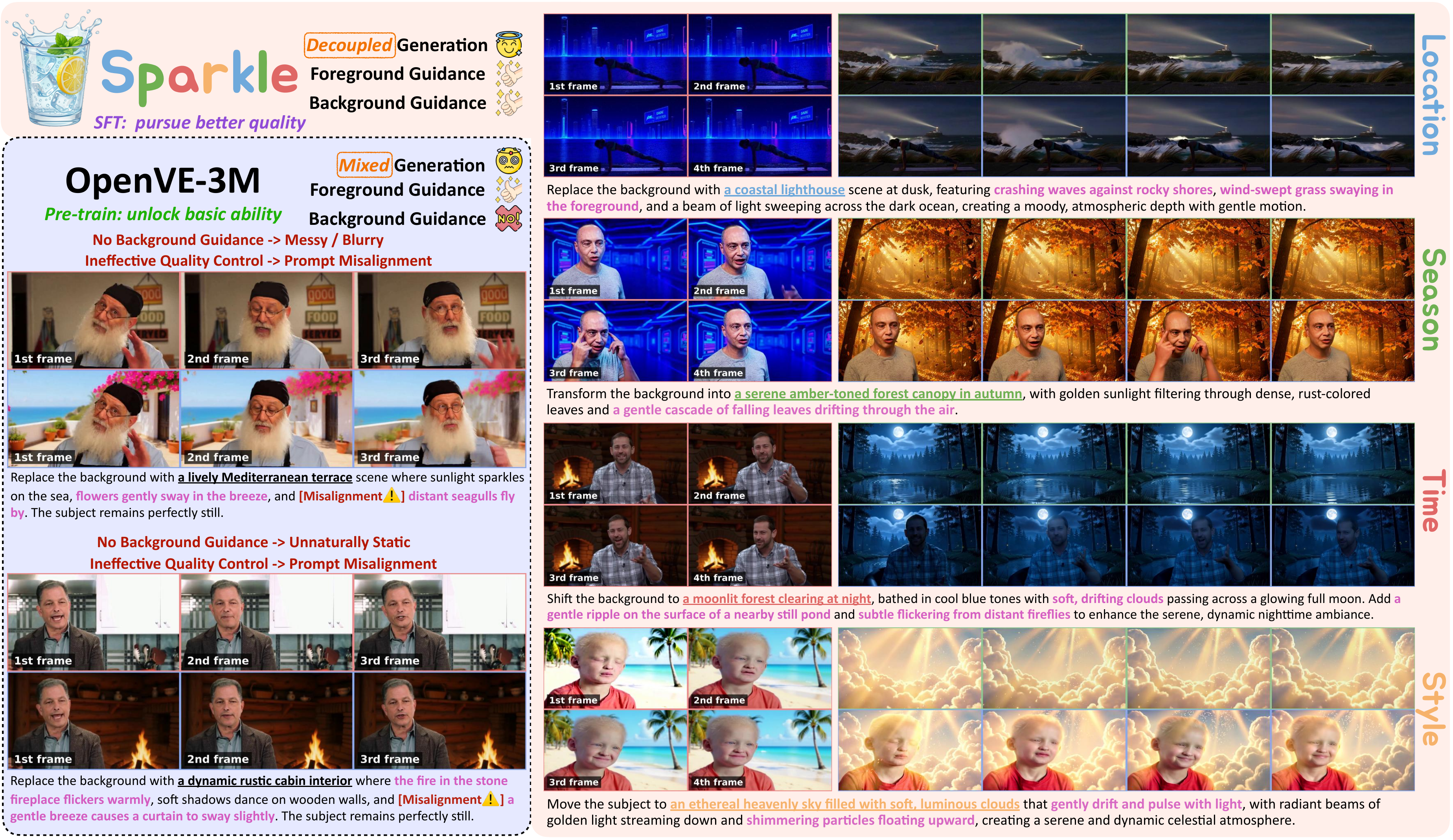}
    \caption{Data comparison between OpenVE-3M~\cite{openve3m} and our proposed \emph{Sparkle}. \textbf{Left:} Relying solely on foreground guidance, OpenVE-3M frequently suffers from severe background structural collapse. \textbf{Right:} \emph{Sparkle} curates foreground-compatible background videos independently. The final synthesis utilizes dual guidance from both the background and the foreground (tracked by our high-precision BAIT algorithm) to ensure dynamic realism. Zoom in for subtle dynamics like crashing waves.}
    \label{fig:teaser}
\vspace{-2em}
\end{figure}

Over the past few years, the visual generation community has evolved rapidly. Within the image domain, significant breakthroughs have been achieved in editing. Open-source models, \emph{e.g.,} Qwen-Image-Edit~\cite{qwenimage} and FLUX.2-klein-9B~\cite{flux2klein}, have gradually narrowed the performance gap with commercial models like Nano Banana 2~\cite{nanobanana2} and GPT-Image-2~\cite{gptimage2}. As a natural extension of image synthesis, video editing has attracted increasing attention from researchers in recent months, and it is emerging as a promising direction that could be highly beneficial for advancing world understanding and inspiring human creativity. Unlike the traditional condition-driven editing paradigm that requires users to prepare depth videos or other auxiliary inputs, \emph{e.g.,} VACE~\cite{vace}, the research community is currently making significant efforts to adapt the success of instruction-guided image editing techniques to video editing, offering a more user-friendly and easily deployable alternative.

Among the various explorations, establishing a robust data infrastructure remains a critical priority for this nascent field. Recently, several works have introduced high-quality video editing data. For instance, Se\~{n}orita-2M~\cite{senorita2m}, ReCo~\cite{reco}, and Ditto-1M~\cite{ditto} provide diverse edits. However, the majority of these datasets focus exclusively on object manipulation and global style transfer. Consequently, they neglect the highly challenging background replacement task requiring large-scale area re-creation while preserving the foreground figures and objects, a capability that is in high demand across numerous real-world applications like film post-production and advertising.

Recently, OpenVE-3M~\cite{openve3m}, the largest open-source video editing dataset to date, became the first to explicitly incorporate background replacement as a supported task. The derivative models, \emph{e.g.}, OpenVE-Edit~\cite{openve3m} and Kiwi-Edit~\cite{kiwiedit}, unlock basic video background replacement capability. However, despite their specialized training, these models struggle to surpass 50\% of the maximum score (\emph{i.e.,} 2.5/5.0) on OpenVE-Bench under the rigorous Gemini-2.5-Pro evaluation. Furthermore, the generated videos frequently suffer from rigid compositing, unnaturally blending dynamic foreground subjects with entirely static backgrounds, and sometimes fail to preserve the foreground subjects, thereby falling significantly short of acceptable visual quality.

To investigate the root cause of these stale background edits, we conducted an in-depth analysis of OpenVE-3M's data pipeline. We observe that it directly feeds the background-replaced initial frame into Wan2.1-Fun-V1.1-14B-Control~\cite{wan} to generate the full video, where the overall motion control signal solely comes from a foreground Canny edge video generated via a single-pass Grounded SAM2 tracking. As illustrated in Figure~\ref{fig:teaser} (left), this pipeline suffers from two primary issues:
\begin{itemize}[leftmargin=1em,itemsep=0.2ex,topsep=0pt]
\item \textbf{Absence of Background Guidance}. This is the primary cause of low-quality background edits. Without explicit background guidance, the model typically ignores background dynamics entirely, \emph{e.g.,} the bottom-left video. In more severe cases, the background structure collapses, resulting in messy or blurry artifacts, \emph{e.g.,} the top-left video. 
\item \textbf{Prompt Misalignment}. Because OpenVE-3M lacks quality filtering, the edited initial frames frequently fail to align with the prompts. For instance, the top-left video completely omits the flying seagulls, and the bottom-left video lacks a curtain entirely, let alone the required dynamics.
\end{itemize}
Furthermore, the single-pass foreground tracking approach is susceptible to \textbf{Entity Loss}, which degrades the foreground guidance quality. As demonstrated in Figure~\ref{fig:teaser} (left), this tracking deficiency fails to preserve fine-grained temporal details. For instance, in the third frame of the top-left video, the subject's originally open hand is erroneously rendered as a closed fist in the edited frame.

Based on these observations, we propose a scalable pipeline designed to synthesize high-quality and lively background replacement data illustrated in Figure~\ref{fig:teaser} (right). Its unique properties are as follows:
\begin{itemize}[leftmargin=1em,itemsep=0.2ex,topsep=0pt]
    \item \textbf{Individual Lively Background Generation.} We abandon the mixed generation paradigm that directly generates edited videos from a composite foreground-background frame. Instead, we propose a novel method that first gathers pure background images compatible with the original foreground. These images are subsequently animated using an I2V model. By omitting the foreground, the model focuses exclusively on background dynamics, producing vivid videos that accurately capture subtle motions (\emph{e.g.,} crashing waves, falling leaves, and drifting clouds).
    \item \textbf{High-Precision Foreground Tracking (BAIT).} To overcome the limitations of coarse, single-pass tracking, we propose \emph{Bbox-Anchor-In-Temporal} (BAIT), a two-stage approach for fine-grained foreground extraction. This pipeline performs VLM-based grounding on sparsely sampled frames, followed by multi-pass dense tracking via SAM3~\cite{sam3}. A voting mechanism then aggregates the resulting masks, ensuring high precision through consensus across diverse temporal anchors.
    \item \textbf{High-Quality Background Replacement via Decoupled Guidance.} Instead of simply cutting out the foreground tracked by BAIT and pasting it onto the new background, we separately extract Canny edges from both the prepared foreground and background. We then regenerate the background-replaced video using a control model. This decoupled approach effectively prevents artifacts such as harsh cutout contours, ensuring exceptional visual quality.
    \item \textbf{Rigorous Quality Filtering.} Inspired by the recent success of image reward models, we apply EditScore~\cite{editreward} after every operation involving content modification (\emph{e.g.,} background generation and final video synthesis). This rigorous filtering significantly suppresses prompt misalignment.
\end{itemize}
Building upon this data pipeline, we introduce the \emph{Sparkle} dataset, comprising $\sim$140K high-quality video pairs tailored for the background replacement task. \emph{Sparkle} encompasses five themes and 21 subthemes across $\sim$100 distinct scenes. Under the OpenVE-Bench evaluation protocols, its data quality significantly surpasses that of OpenVE-3M. Furthermore, it maintains a balanced difficulty level optimal for model training, as evidenced by the substantial performance gains observed in a \emph{Sparkle}-tuned general video editor, \emph{i.e.,} Kiwi-Edit~\cite{kiwiedit}. Additionally, we propose \emph{Sparkle-Bench}, the largest background replacement benchmark to date, covering 458 videos across $\sim$100 scenes. This benchmark is accompanied by a fine-grained six-dimensional evaluation protocol. We believe our dataset, benchmark, and model will facilitate more comprehensive research in this field.
\vspace{-1em}
\section{Related Work}
\vspace{-0.5em}

\noindent \textbf{Instruction-Guided Video Editing Datasets.} As instruction-guided video editing is a rapidly emerging research area, the community has made significant strides in establishing its data infrastructure over the past year. Current data synthesis paradigms for instruction-video pairs can be broadly categorized into two approaches: 
(\textbf{i}) \emph{One-step V2V Generation}. This approach is primarily applied to relatively simple tasks, such as object removal. For instance, Se\~{n}orita-2M~\cite{senorita2m} trains a dedicated video remover that directly operates on source videos to generate object removal data. Similarly, OpenVE-3M~\cite{openve3m} adopts DiffuEraser~\cite{diffueraser} to erase target objects within source videos. 
(\textbf{ii}) \emph{Two-step I2I + I2V Generation}. This represents a more generalized paradigm applicable to complex tasks, such as object swapping, local modification, or global style transfer. Recent datasets, including InsViE-1M~\cite{insvie}, Se\~{n}orita-2M~\cite{senorita2m}, Ditto-1M~\cite{ditto}, and OpenVE-3M~\cite{openve3m}, adopt this pipeline for both local and global manipulations. Typically, the first frame of the source video is extracted and processed by an image editing or inpainting model. Subsequently, an in-context video generator leverages this edited frame, along with auxiliary conditions such as depth maps, to synthesize the final edited video.

The aforementioned paradigms excel at local manipulation and style transfer because they avoid the large-scale scene re-creation and strict foreground preservation required for background replacement. This complexity leads to the scarcity of high-quality data for this task. OpenVE-3M attempted to address this gap via the I2I + I2V paradigm. It uses FLUX.1-Kontext~\cite{flux1kontext} to replace the first frame's background and synthesizes the full video with Wan2.1-Fun-V1.1-14B-Control~\cite{wan}, guided by foreground Canny edges tracked by Grounded SAM2~\cite{groundedsam}. While this preserves the foreground, it suffers from severe background structural collapse as discussed in Section~\ref{sec:introduction}, resulting in sub-optimal data quality. In contrast, we introduce a novel \emph{decoupled} generation paradigm tailored specifically for background replacement. By independently generating precise foreground and background guidance, our approach maintains control over subtle motions. Consequently, the \emph{Sparkle} dataset and its derivative model achieve significant quality improvements over the OpenVE-3M baseline, fully demonstrating the effectiveness of our pipeline.

\noindent \textbf{Video Editing Models.} Traditional video editing models typically rely on auxiliary control signals. For example, VACE~\cite{vace} requires inputs such as Canny edges or depth maps to execute an edit. Following the introduction of high-quality instruction-guided video editing datasets~\cite{senorita2m,ditto,insvie,openve3m}, the paradigm has rapidly shifted toward natural language-driven editing, which eliminates the need for explicit auxiliary conditions. Several notable models have recently emerged in this space, \emph{e.g.,} InstructX~\cite{instructx}, UniVideo~\cite{univideo}, and Kiwi-Edit~\cite{kiwiedit}. Nevertheless, due to the scarcity of high-quality background replacement data, existing models struggle with this specific task. They often inherit the data deficiencies of their upstream training sets, \emph{e.g.,} OpenVE-3M, resulting in stale and rigid edits. To validate our data pipeline, we select a representative medium-sized model, \emph{i.e.,} Kiwi-Edit, and fine-tune it on the proposed \emph{Sparkle} dataset. \emph{We intentionally avoid any structural modifications to the model architecture to ensure that all performance gains stem purely from the enhanced data quality.} Experimental results show that the \emph{Sparkle}-tuned Kiwi-Edit, namely \emph{Kiwi-Sparkle}, significantly outperforms the baseline, firmly validating the high quality and effectiveness of our curated dataset.
\vspace{-1em}
\section{Methodology}
\vspace{-0.5em}

In this section, we detail the five-stage data pipeline used to construct the proposed \emph{Sparkle} dataset, as illustrated in Figure~\ref{fig:framework}. This sequential process integrates rigorous data filtering across all stages, encompassing source video collection, independent background generation, high-precision foreground tracking, and decoupled guidance-driven background replacement.

\vspace{-0.5em}
\subsection{Source Video Collection}
\vspace{-0.5em}
\label{subsec:source_video_collection}

To efficiently harvest a diverse corpus for background replacement, we sample source and edited videos from OpenVE-3M at 2FPS. We then evaluate the paired frames using EditScore~\cite{editscore}, discarding videos with an average frame-level overall score below 8. We hypothesize that these remaining videos are more amenable to high-quality manipulation via current open-source toolkits. This initial filtering stage yields a preliminary pool of $\sim$940K source videos.

Since current open-source models struggle to synchronize the camera movement of the edited video with that of the source video, we restrict our scope to fixed-camera videos, enabling natural background detachment. To efficiently handle the large video volume, we employ a coarse-to-fine filtering approach (Figure~\ref{fig:framework}, Stage 1). The coarse stage detects camera movement via optical flow computed by Unimatch~\cite{unimatch} and homography matrix estimation. Due to space constraints, we defer the algorithmic details to Appendix~\ref{sec:appendix_source_video_collection}. This process rapidly reduces the source pool from $\sim$940K to $\sim$260K. To address cases missed by the coarse stage, we further implement a fine-grained VLM filter. Specifically, we utilize Qwen3-VL-32B~\cite{qwen3vl} to detect residual camera movement across the entire video, requiring the model to articulate its reasoning before judging to ensure high accuracy. This rigorous step further reduces the candidate pool from $\sim$260K to $\sim$224K.

\vspace{-0.5em}
\subsection{Preliminary Background Replacement}
\vspace{-0.5em}

To generate diverse editing prompts, we first reuse existing prompts from OpenVE-3M's background replacement tasks, establishing a robust baseline for direct quality comparison. Next, based on a systematic review of existing datasets, we leverage Gemini-2.5-Pro to hierarchically categorize scene types into four themes (\emph{Location}, \emph{Season}, \emph{Time}, and \emph{Style}). Each theme comprises 4--6 subthemes, with $\sim$10 specific scenes per subtheme. The statistical distribution of these categories is illustrated in Figure~\ref{fig:sparkle_stat} and will be discussed later. Finally, Qwen3-VL-32B formulates comprehensive editing instructions for all source videos. To ensure accurate visual comprehension, it first describes the original scene before randomly selecting a target subtheme and scene to generate the final prompt.

\begin{figure}[t]
    \centering
    \includegraphics[width=1.\columnwidth]{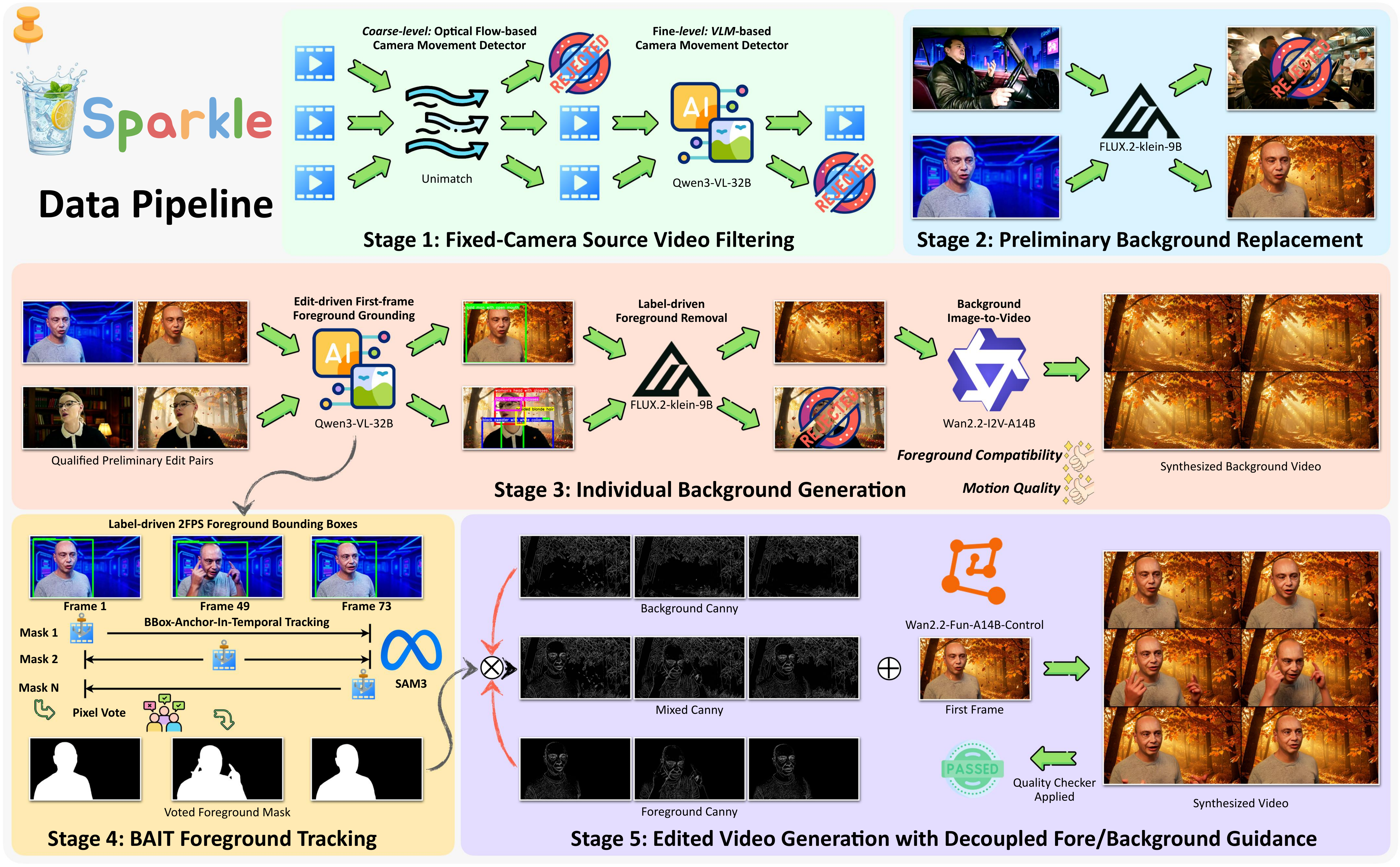}
    \caption{The \emph{Sparkle} data pipeline. First, only fixed-camera videos are retained to enable independent background generation. After preliminary first-frame background replacement, a VLM identifies the foreground, which is then removed to isolate a pure background image. An I2V model animates this image into a background video. Concurrently, our BAIT algorithm precisely tracks the foreground. Finally, decoupled foreground and background Canny edges guide video synthesis, conditioned on the edited first frame. EditScore~\cite{editscore} filters low-quality outputs after every modification.}
    \label{fig:framework}
\vspace{-1.5em}
\end{figure}

Next, we perform a preliminary background replacement by leveraging FLUX.2-klein-9B~\cite{flux2klein} to edit the first frame of the source video according to the prompt. Because the editing process can occasionally fail, \emph{e.g.,} missing required background elements, we employ an image editing reward model, \emph{i.e.,} EditScore~\cite{editscore}, to evaluate the output quality. We filter out any edits with an overall score below 8, as this typically indicates prompt misalignment or poor visual fidelity. The overall workflow is illustrated in Figure~\ref{fig:framework}, Stage 2. These successfully edited frames then serve as the initial condition for the final video synthesis.

\vspace{-0.5em}
\subsection{Individual Background Generation}
\vspace{-0.5em}

Although we obtain high-quality edited initial frames in the previous stage, directly synthesizing the video using a control model guided solely by the foreground inevitably leads to structural collapse or motion loss within the background, thereby significantly degrading overall visual quality. This degradation occurs because control models, \emph{e.g.,} Wan2.1-Fun-V1.1-14B-Control~\cite{wan}, are prone to over-concentrating on the foreground when explicit background guidance is absent.

To address this limitation, we propose a novel pipeline to completely detach the foreground from the background, enabling decoupled guidance. As shown in Figure~\ref{fig:framework}, Stage 3, the process begins with edit-driven foreground grounding. Qwen3-VL-32B compares the original and preliminarily edited first frames to identify foreground elements to preserve. These labels are translated into removal instructions, \emph{e.g.,} ``Remove the bald man'', for FLUX.2-klein-9B to erase the foreground from the edited first frame. This operation ensures \emph{foreground compatibility}, as the isolated background derives directly from the composite frame. To guarantee a perfectly clean background, we apply EditScore~\cite{editscore} after each removal, using a stricter threshold of 8.5 to discard sub-optimal outputs.

Finally, we use Qwen3-VL-32B to extract the target background caption from the editing prompt. We then feed the isolated background image into an I2V model, \emph{i.e.,} Wan2.2-I2V-A14B, utilizing the extracted caption as the textual condition. To accelerate this time-consuming process, we employ a four-step distilled version~\cite{lightx2v}, as we observed no significant quality degradation for this task. Unhindered by foreground elements, the model focuses entirely on rendering the required background dynamics, \emph{e.g.,} swaying grass, thereby generating a high-quality, motion-centric background video.

\vspace{-0.5em}
\subsection{Bbox-Anchor-In-Temporal (BAIT) Foreground Tracking}
\vspace{-0.5em}

\begin{figure}[t]
    \centering
    \includegraphics[width=.9\columnwidth]{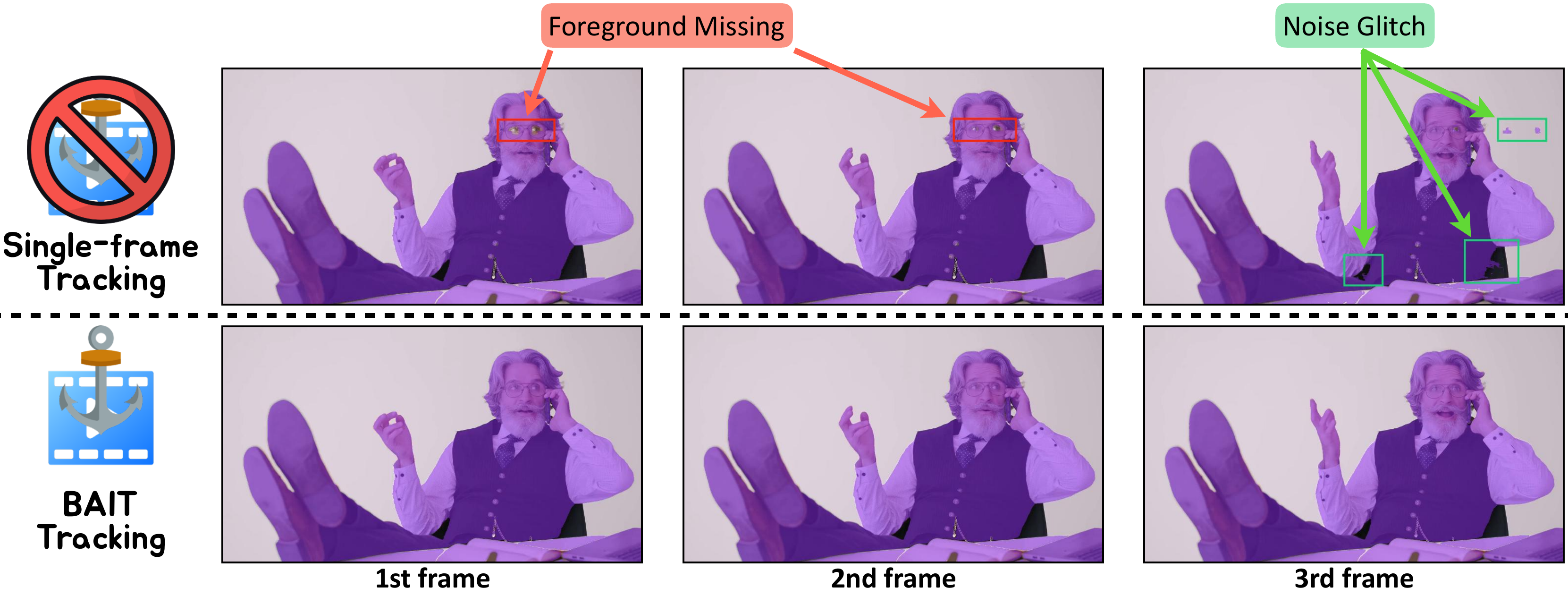}
    \vspace{-1em}
    \caption{Visual comparison between single-frame tracking (top) and our BAIT (bottom). The red and green boxes highlight \textcolor{baitbox_red}{\textbf{foreground missing}} and \textcolor{baitbox_green}{\textbf{noise glitches}} in single-frame tracking, respectively.}
    \label{fig:bait}
\vspace{-1.5em}
\end{figure}

As discussed in Section~\ref{sec:introduction}, the single-pass tracking employed by OpenVE-3M is susceptible to entity loss, which leads to occasional visual inconsistencies between the source and edited frames. Therefore, in addition to our independent background generation approach, we propose a high-precision foreground tracking algorithm termed \emph{Bbox-Anchor-In-Temporal} (BAIT).

To begin, we prompt Qwen3-VL-32B to conduct a second round of grounding on frames sampled at 2FPS, tracing the foreground labels obtained in Figure~\ref{fig:framework}, Stage 3, to extract precise bounding boxes. These bounding boxes at various timestamps serve as explicit temporal anchors. Next, utilizing these boxes as visual prompts, we employ SAM3~\cite{sam3} to perform $N$ isolated forward and backward tracking passes, where $N$ denotes the total number of sampled frames. Finally, we apply a pixel-wise voting mechanism across the resulting $N$ video masks: a pixel is assigned to the final foreground mask only if a majority consensus is reached, \emph{i.e.,} predicted as foreground by more than half of the masks; otherwise, it is classified as background. The whole process is illustrated in Figure~\ref{fig:framework}, Stage 4.

Figure~\ref{fig:bait} illustrates the advantages of leveraging consensus across multiple temporal anchors. The top row demonstrates single-pass tracking initialized from a single frame's bounding boxes, which frequently encounters \emph{foreground missing} (the incompletely tracked glasses in red boxes) and \emph{noise glitches} (artifact spots on the background in green boxes). By employing our proposed BAIT algorithm, these artifacts are effectively suppressed, resulting in clean and precise foreground masks.

\vspace{-0.5em}
\subsection{Edited Video Generation with Decoupled Guidance}
\vspace{-0.5em}

With the BAIT tracking results, we can derive decoupled guidance signals from background videos and foreground objects to synthesize the edited video. We adopt Wan2.2-Fun-A14B-Control~\cite{wan} as our video generator, which supports three guidance types: Canny, Pose, and Depth. We observe that: (\textbf{i}) \emph{Pose is unsuitable for background generation}, as it primarily models human body structures. (\textbf{ii}) \emph{Depth struggles to capture fine-grained background dynamics}, particularly for elements with similar depth values, and incurs substantial computational overhead. (\textbf{iii}) \emph{Canny better preserves dynamic details} through distinguishable edge structures while being significantly more computationally efficient. Therefore, we select Canny as our guidance signal for scalable data generation.

Specifically, we extract Canny edges from the source and background videos using Lineart~\cite{lineart}, and combine them according to the foreground mask generated by BAIT. Within the foreground contour, we utilize the Canny edges from the source video; otherwise, we use the Canny edges from the background video. This process yields a high-quality, comprehensive control video derived from decoupled foreground and background guidance. This guidance, along with the edited first frame from Figure~\ref{fig:framework}, Stage 2, is fed into the control model, \emph{i.e.,} Wan2.2-Fun-A14B-Control~\cite{wan}, to synthesize the final background-replaced video. Lastly, we uniformly sample four frames while excluding the first frame from the synthesized video  (which was already evaluated in Stage 2) and compute the average overall score via EditScore. We discard videos with an average score below 8. Figure~\ref{fig:framework}, Stage 5 illustrates the full workflow. Compared to the naive foreground copy-and-paste shortcut, this regeneration paradigm effectively avoids artifacts such as harsh cutout contours, ensuring the synthesized videos maintain high quality.

\vspace{-0.5em}
\subsection{Dataset Statistics}
\vspace{-0.5em}

\begin{wrapfigure}{r}{0.4\columnwidth}
\vspace{-2em}
    \centering
    \includegraphics[width=1.\linewidth]{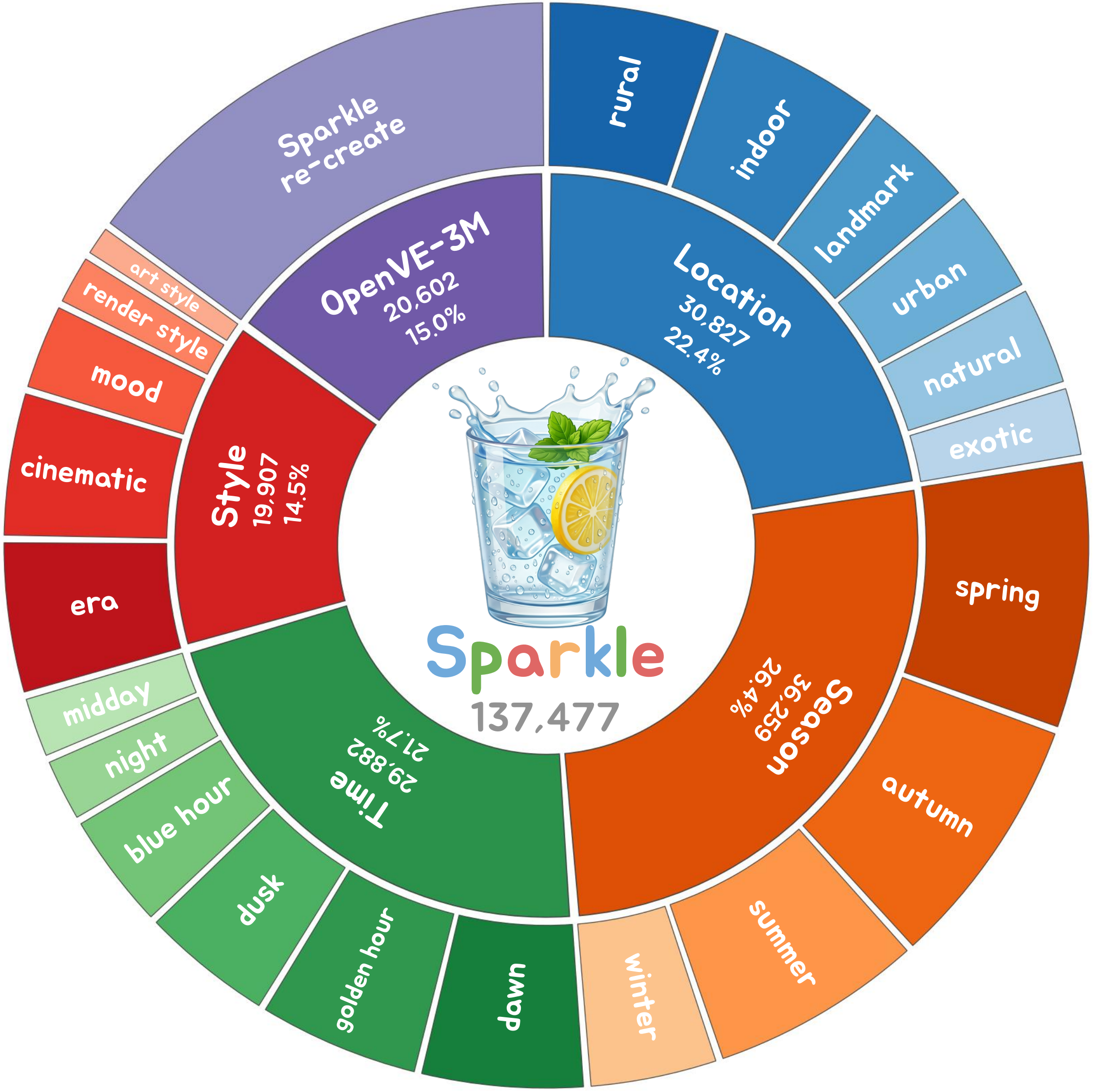}
    \caption{\emph{Sparkle} statistical distribution.}
    \label{fig:sparkle_stat}
\vspace{-2em}
\end{wrapfigure}
Building upon the aforementioned pipeline, we curated \emph{Sparkle}, comprising $\sim$140K videos across five relatively balanced themes and 22 subthemes across $\sim$100 diverse scenes, as shown in Figure~\ref{fig:sparkle_stat}. Notably, our \emph{Style} theme differs from simple global style transfer by requiring the foreground to remain entirely intact while modifying only the background. This challenging constraint for existing models results in a lower yield of qualified videos compared to other themes. In summary, \emph{Sparkle} covers diverse background replacement scenarios at a modest scale, making it highly suitable for capability refinement following large-scale pre-training. As demonstrated in our Experiments, lightweight fine-tuning on \emph{Sparkle} yields significant improvements, firmly validating the substantial benefit of our data pipeline and the dataset.

\vspace{-0.5em}
\subsection{Sparkle-Bench}
\vspace{-0.5em}
\label{subsec:sparkle_bench}

Beyond the dataset, we also introduce \emph{Sparkle-Bench}, a benchmark tailored specifically for background replacement. To ensure an appropriate level of difficulty, we construct this benchmark using candidate videos that passed the first four stages of our pipeline but failed the final quality check in Stage 5. These videos provide ideal evaluation targets: having passed most checks, their lower synthesis scores indicate they are challenging yet viable for manipulation. Through rigorous manual inspection, we selected 4-5 appropriately challenging videos per subtheme. This yields 458 videos covering 97 distinct scenes across 21 subthemes as shown in Table~\ref{tab:sparkle_bench_statistics} (the OpenVE-3M subtheme from Sparkle is excluded, as it does not correspond to a specific editing type). As the largest benchmark of its kind to date, we believe \emph{Sparkle-Bench} offers the community comprehensive evaluation insights.

\begin{wraptable}{r}{0.45\columnwidth}
\setlength{\tabcolsep}{4pt}
\centering
\vspace{-1.5em}
\caption{Statistics of \emph{Sparkle-Bench}.}
\resizebox{1.\linewidth}{!}{
\begin{tabular}{lcccc}
\toprule
\textbf{Theme} & \textbf{Subtheme} & \textbf{Scene} & \textbf{Vid / Scene} & \textbf{Videos} \\
\midrule
Location        & 6        & 27    & 4           & 108    \\
Season          & 4        & 24    & 5           & 120    \\
Time            & 6        & 24    & 5           & 120    \\
Style           & 5        & 22    & 5           & 110    \\
Total           & 21       & 97    & -           & 458    \\
\bottomrule
\end{tabular}}
\label{tab:sparkle_bench_statistics}
\vspace{-0.5em}
\end{wraptable}
Regarding the evaluation metrics, we find the conventional OpenVE-Bench protocol somewhat coarse. Therefore, we propose a set of six-dimensional criteria (each scored on a 1-to-5 scale) spanning three perspectives, specifically tailored for background replacement: \textbf{(i)} \emph{Global Assessment}, which includes \emph{Instruction Compliance} to measure overall prompt adherence, and \emph{Overall Visual Quality} to encompass global video quality and foreground-background harmonization (with specific consideration given to lighting and shadow adjustments); \textbf{(ii)} \emph{Foreground Assessment}, which includes \emph{Foreground Integrity} to assess whether the foreground is preserved intact, and \emph{Foreground Motion Consistency} to evaluate whether the preserved foreground subjects behave consistently with the source videos; and \textbf{(iii)} \emph{Background Assessment}, which includes \emph{Background Dynamics} to measure the dynamic realism of the background (\emph{i.e.,} whether it accurately produces the required motion), and \emph{Background Visual Quality} to determine if the replaced background maintains a high aesthetic standard. Following OpenVE-Bench, we constrain the scores of the other five dimensions to be no higher than \emph{Instruction Compliance}, thereby emphasizing instruction-following. Please refer to Appendix~\ref{sec:appendix_detailed_evaluation_protocol} for more details.
\vspace{-1em}
\section{Experiments}
\vspace{-0.5em}

\vspace{-0.5em}
\subsection{Experimental Setup}
\vspace{-0.5em}

\begin{table}[t]
\centering
\caption{Data quality assessment. We randomly sample 500 videos per theme to represent the overall distribution. \textbf{\textcolor{gray}{Gray}} numbers denote the scores of OpenVE-3M raw edits, which share the same source videos and prompts as our OpenVE-3M subset and can be directly compared. \textbf{\textcolor{red!60}{Red}} numbers denote the absolute gain compared to the OpenVE-3M baseline.}
\resizebox{.9\columnwidth}{!}{
\begin{tabular}{lcccccc}
\toprule
\textbf{Dimensions}                       & \multicolumn{2}{c}{\textbf{OpenVE-3M}} & \textbf{Location} & \textbf{Season} & \textbf{Time} & \textbf{Style} \\
\midrule
Instruction Compliance           & \textcolor{gray}{3.34}          & 3.82 \textcolor{red!60}{(+14\%)}         & 4.09 \textcolor{red!60}{(+22\%)}     & 4.04 \textcolor{red!60}{(+21\%)}   & 4.03 \textcolor{red!60}{(+21\%)} & 3.85 \textcolor{red!60}{(+15\%)}  \\
Consistency \& Detail   Fidelity & \textcolor{gray}{2.91}          & 3.62 \textcolor{red!60}{(+24\%)}          & 3.81 \textcolor{red!60}{(+31\%)}     & 3.75 \textcolor{red!60}{(+29\%)}   & 3.70 \textcolor{red!60}{(+27\%)} & 3.65 \textcolor{red!60}{(+25\%)}  \\
Visual Quality \& Stability      & \textcolor{gray}{3.01}          & 3.68 \textcolor{red!60}{(+22\%)}         & 3.75 \textcolor{red!60}{(+25\%)}     & 3.81 \textcolor{red!60}{(+27\%)}   & 3.80 \textcolor{red!60}{(+26\%)} & 3.65 \textcolor{red!60}{(+21\%)}  \\
Average Score                    & \textcolor{gray}{3.09}          & 3.71 \textcolor{red!60}{(+20\%)}         & 3.88 \textcolor{red!60}{(+26\%)}     & 3.86 \textcolor{red!60}{(+25\%)}   & 3.84 \textcolor{red!60}{(+24\%)} & 3.72 \textcolor{red!60}{(+20\%)} \\
\bottomrule
\end{tabular}}
\label{tab:data_quality}
\vspace{-1.5em}
\end{table}

Since the primary focus of this paper is the data pipeline and the resulting \emph{Sparkle} dataset, we perform a lightweight fine-tuning on a general video editing model, \emph{i.e.,} Kiwi-Edit~\cite{kiwiedit}, without any architectural modifications. By doing so, we demonstrate that the observed performance gains stem purely from the superior quality of our data. Specifically, we fine-tune the model on the proposed \emph{Sparkle} dataset for 10K steps with a batch size of 128, namely \emph{Kiwi-Sparkle}. All other training configurations remain identical to those detailed in the official Kiwi-Edit repository.

For evaluation, we primarily adopt the OpenVE-Bench protocol to ensure continuous comparison with the OpenVE-3M baseline. This protocol evaluates \emph{Instruction Compliance}, \emph{Consistency \& Detail Fidelity}, and \emph{Visual Quality \& Stability} on a 1-to-5 scale. Following OpenVE-3M~\cite{openve3m}, we cap the latter two scores at the \emph{Instruction Compliance} score. This constraint prevents score hacking, where models might inflate visual quality at the expense of accurate instruction following. For \emph{Sparkle-Bench}, we utilize the six-dimensional metrics detailed in Section~\ref{subsec:sparkle_bench}. Across all benchmarks, we employ Gemini-2.5-Pro as the evaluator due to its exceptional video understanding capabilities.

\vspace{-0.5em}
\subsection{Main Results}
\vspace{-0.5em}

\noindent \textbf{Data Quality Assessment.} We evaluate the data quality of \emph{Sparkle} by randomly sampling 500 videos per theme due to quota constraints. For the OpenVE-3M subset, our recreated videos share the exact source videos and prompts with the original dataset, enabling rigorous direct comparison. As shown in Table~\ref{tab:data_quality}, the OpenVE-3M baseline (gray numbers) scores poorly across all dimensions, explaining why its derivative models struggle to surpass 2.5/5.0 in downstream evaluations. In contrast, \emph{Sparkle} achieves average gains of over 20\% in both the OpenVE-3M subset and the remaining four themes. The particularly significant improvements in \emph{Consistency \& Detail Fidelity} and \emph{Visual Quality \& Stability} indicate that while OpenVE-3M suffers from severe structural degradation due to its sole reliance on foreground guidance, our decoupled guidance paradigm effectively mitigates these issues, substantially enhancing overall quality.

\begin{table}[t]
\setlength{\tabcolsep}{8pt}
\centering
\caption{Scores for the background replacement task on OpenVE-Bench. Ins, Cons, and VQ stand for \emph{Instruction Compliance}, \emph{Consistency \& Detail Fidelity}, and \emph{Visual Quality \& Stability}, respectively.}
\resizebox{.9\columnwidth}{!}{
\begin{tabular}{lcccccccc}
\toprule
\textbf{Model} & \textbf{Params.} & \textbf{Res.} & \textbf{Frames} & \textbf{Public Access} & \textbf{Overall} & \textbf{Ins.} & \textbf{Cons.} & \textbf{VQ.} \\
\midrule
InsViE~\cite{insvie}         & 2B               & 480P          & 25              & Model \& Data         & 1.02             & 1.03          & 1.02                   & 1.02                 \\
DITTO~\cite{ditto}          & 14B              & 480P          & 81              & Model \& Data         & 1.55             & 1.66          & 1.56                   & 1.44                 \\
ICVE~\cite{icve}           & 13B              & 480P          & 81              & Model-Only            & 1.87             & 2.15          & 1.83                   & 1.63                 \\
OmniVideo2~\cite{omnivideo2}     & A14B             & 480P          & 41              & Model-Only            & 2.04             & 2.19          & 2.03                   & 1.90                 \\
Lucy-Edit-1.1~\cite{lucyedit}  & 5B               & 720P          & 81              & Model-Only            & 2.10             & 2.41          & 2.03                   & 1.86                 \\
OpenVE-Edit~\cite{openve3m}    & 5B               & 720P          & -               & Data-Only             & 2.36             & -             & -                      & -                    \\
Kiwi-Edit~\cite{kiwiedit}      & 5B               & 720P          & 81              & Model \& Data         & 2.58             & 2.81          & 2.58                   & 2.36                 \\
Runway Aleph~\cite{runway_aleph}   & -                & 720P          & -               & Proprietary           & 2.62             & -             & -                      & -                    \\
UniVideo~\cite{univideo}       & 13B              & 480P          & 81              & Model-Only            & 2.74             & 3.12          & 2.64                   & 2.46                 \\
\midrule
\rowcolor{shallow_blue}
Kiwi-Sparkle (Ours)   & 5B               & 720P          & 81              & Model \& Data         & \textbf{3.29}             & \textbf{3.51}          & \textbf{3.15}                   & \textbf{3.22}                \\
\bottomrule
\end{tabular}}
\label{tab:openve_bench}
\vspace{-0.5em}
\end{table}

\noindent \textbf{Performance on OpenVE-Bench.} Although Table~\ref{tab:data_quality} demonstrates that our \emph{Sparkle} dataset inherently possesses high data quality, if the editing pairs are too difficult for a model to learn from, the impact of our technical contributions would be diminished. To investigate this, we fine-tuned a medium-sized general video editor, \emph{i.e.,} Kiwi-Edit, on \emph{Sparkle} (referred to as \emph{Kiwi-Sparkle}) and evaluated its background replacement performance on OpenVE-Bench. The results are presented in Table~\ref{tab:openve_bench}. Notably, even the best open-source models trained on proprietary internal data, \emph{e.g.,} UniVideo~\cite{univideo}, fail to reach the 60\% score threshold (3.0/5.0), highlighting the severe scarcity of high-quality background replacement data within the current community. Conversely, \emph{Kiwi-Sparkle} exhibits a remarkable boost compared to existing baselines, achieving a 28\% overall gain from Kiwi-Edit and outperforming competitors with $3\times$ more parameters, \emph{e.g.,} UniVideo and OmniVideo2. This demonstrates that \emph{Sparkle} not only significantly improves instruction compliance and the visual quality of such edits, but also maintains a well-balanced difficulty level that allows general video editors to effectively inherit its knowledge, thereby making a timely contribution to the field.

\noindent \textbf{Performance on Sparkle-Bench.} Table~\ref{tab:sparkle_bench} presents the overall scores across the four themes on \emph{Sparkle-Bench}. Encompassing $\sim$100 diverse scenes, this benchmark demands broader background replacement capabilities. We observe that models specifically enhanced for background editing demonstrate greater robustness. For instance, Lucy-Edit-1.1~\cite{lucyedit} achieves better performance here than on OpenVE-Bench. Conversely, general models like UniVideo suffer from degraded performance. Their low \emph{Background Dynamics} scores suggest a deficit of high-quality training data, resulting in poorly animated backgrounds. In contrast, our \emph{Kiwi-Sparkle} exhibits strong performance, significantly improving both instruction-following and the generation quality of the foreground and background, solidifying our contribution. Please refer to Appendix~\ref{subsec:appendix_full_results_on_sparkle_bench} for specific scores of the four themes.

\begin{table}[t]
\setlength{\tabcolsep}{8pt}
\centering
\caption{Scores on \emph{Sparkle-Bench}. Abbreviations: \emph{Instruction Compliance} (Ins), \emph{Overall Visual Quality} (Vis), \emph{Foreground Integrity} (FgIn), \emph{Foreground Motion Consistency} (FgMo), \emph{Background Dynamics} (BgDy), and \emph{Background Visual Quality} (BgVi).}
\resizebox{.9\columnwidth}{!}{
\begin{tabular}{lcccccccc}
\toprule
\multirow{2}{*}{\textbf{Model}} & \multirow{2}{*}{\textbf{Configuration}} & \multirow{2}{*}{\textbf{Overall}} & \multicolumn{2}{c}{\textbf{Global}} & \multicolumn{2}{c}{\textbf{Foreground}} & \multicolumn{2}{c}{\textbf{Background}} \\
\cmidrule(l){4-9}
                                &                                         &                                   & Ins.             & Vis.             & FgIn.              & FgMo.              & BgDy.              & BgVi.              \\
\midrule
InsViE~\cite{insvie}                          & 2B-25F@480P                             & 1.05                              & 1.08             & 1.03             & 1.05               & 1.08               & 1.04               & 1.05               \\
ICVE~\cite{icve}                            & 13B-81F@480P                            & 1.95                              & 2.25             & 1.49             & 2.13               & 2.24               & 1.58               & 1.99               \\
DITTO~\cite{ditto}                           & 14B-81F@480P                            & 2.01                              & 2.15             & 1.94             & 1.94               & 2.11               & 1.77               & 2.13               \\
OmniVideo2~\cite{omnivideo2}                      & A14B-41F@480P                           & 2.35                              & 2.58             & 2.17             & 2.43               & 2.48               & 1.92               & 2.55               \\
UniVideo~\cite{univideo}                        & 13B-81F@480P                            & 2.41                              & 2.71             & 1.82             & 2.59               & 2.71               & 2.02               & 2.62               \\
Kiwi-Edit~\cite{kiwiedit}                       & 5B-81F@720P                             & 2.54                              & 2.92             & 2.15             & 2.86               & 2.90               & 1.57               & 2.84               \\
Lucy-Edit-1.1~\cite{lucyedit}                   & 5B-81F@720P                             & 2.74                              & 3.06             & 2.23             & 2.78               & 3.04               & 2.46               & 2.83               \\
\midrule
\rowcolor{shallow_blue}
Kiwi-Sparkle (Ours)                    & 5B-81F@720P                             & \textbf{3.81}                              & \textbf{4.10}             & \textbf{3.40}             & \textbf{3.77}               & \textbf{4.05}               & \textbf{3.54}               & \textbf{3.99}     \\
\bottomrule
\end{tabular}}
\label{tab:sparkle_bench}
\vspace{-1.5em}
\end{table}

\vspace{-1em}
\subsection{Ablation Studies}
\vspace{-0.5em}

\begin{table}[t]
\setlength{\tabcolsep}{4pt}
\centering
\caption{Comparison between the \emph{Copy-and-Paste} and our \emph{Decoupled} generation paradigms for video synthesis. \textbf{\textcolor{red!60}{Red}} numbers denote the absolute gain compared to the Copy-and-Paste baseline.}
\resizebox{1.\columnwidth}{!}{
\begin{tabular}{lcccccccccc}
\toprule
\multirow{2}{*}{\textbf{Dimensions}}      & \multicolumn{2}{c}{\textbf{OpenVE-3M}} & \multicolumn{2}{c}{\textbf{Location}} & \multicolumn{2}{c}{\textbf{Season}} & \multicolumn{2}{c}{\textbf{Time}} & \multicolumn{2}{c}{\textbf{Style}} \\
\cmidrule(l){2-11}
                                 & Copy          & Decouple        & Copy          & Decouple       & Copy         & Decouple      & Copy        & Decouple     & Copy        & Decouple      \\
\midrule
Instruction Compliance           & 3.12        & 3.82 \textcolor{red!60}{(+22\%)}            & 3.15        & 4.09 \textcolor{red!60}{(+30\%)}           & 3.10       & 4.04 \textcolor{red!60}{(+30\%)}          & 2.88      & 4.03 \textcolor{red!60}{(+40\%)}         & 3.26      & 3.85 \textcolor{red!60}{(+18\%)}          \\
Consistency \& Detail   Fidelity & 2.57        & 3.62 \textcolor{red!60}{(+41\%)}            & 2.45        & 3.81 \textcolor{red!60}{(+56\%)}           & 2.46       & 3.75 \textcolor{red!60}{(+52\%)}          & 2.20      & 3.70 \textcolor{red!60}{(+68\%)}         & 2.61      & 3.65 \textcolor{red!60}{(+40\%)}         \\
Visual Quality \& Stability      & 2.36        & 3.68 \textcolor{red!60}{(+56\%)}           & 2.14        & 3.75 \textcolor{red!60}{(+75\%)}           & 2.18       & 3.81 \textcolor{red!60}{(+75\%)}          & 1.77      & 3.80 \textcolor{red!60}{(+115\%)}         & 2.28      & 3.65 \textcolor{red!60}{(+60\%)}         \\
Average Score                    & 2.68        & 3.71 \textcolor{red!60}{(+38\%)}            & 2.58        & 3.88 \textcolor{red!60}{(+50\%)}           & 2.58       & 3.86 \textcolor{red!60}{(+50\%)}          & 2.28      & 3.84 \textcolor{red!60}{(+68\%)}         & 2.72      & 3.72 \textcolor{red!60}{(+37\%)}  \\
\bottomrule
\end{tabular}}
\label{tab:ab_cp}
\end{table}

\noindent \textbf{Comparison to Copy-and-Paste Video Synthesis.} A shortcut for final synthesis is directly pasting the foreground onto the background. However, this naive approach introduces artifacts like harsh contours and ignores crucial shadow adjustments in light-sensitive scenarios, \emph{e.g.,} time-oriented editing, resulting in inharmonious compositions. To validate this, we evaluated 500 copy-and-pasted videos per theme using sources from Table~\ref{tab:data_quality}. As Table~\ref{tab:ab_cp} shows, this rigid paradigm severely degrades overall quality, whereas our approach achieves a notable 115\% visual quality gain over this baseline in the \emph{Time} theme. These results demonstrate that our decoupled paradigm, which leverages Canny edges and the edited first frame for full-video regeneration, effectively ensures dynamic realism and harmonious environmental integration, yielding significantly higher-quality outputs.

\begin{table}[t]
\setlength{\tabcolsep}{1pt}
\centering
\caption{Comparison of video quality using \emph{Foreground-only} guidance (FG-Only) versus our \emph{Decoupled} guidance (FG+BG). \textbf{\textcolor{red!60}{Red}} numbers denote the absolute gain compared to the FG-Only baseline.}
\resizebox{1.\columnwidth}{!}{
\begin{tabular}{lcccccccccc}
\toprule
\multirow{2}{*}{\textbf{Dimensions}}      & \multicolumn{2}{c}{\textbf{OpenVE-3M}} & \multicolumn{2}{c}{\textbf{Location}} & \multicolumn{2}{c}{\textbf{Season}} & \multicolumn{2}{c}{\textbf{Time}} & \multicolumn{2}{c}{\textbf{Style}} \\
\cmidrule(l){2-11}
                                 & FG-Only            & FG+BG         & FG-Only           & FG+BG         & FG-Only          & FG+BG        & FG-Only         & FG+BG       & FG-Only          & FG+BG       \\
\midrule
Instruction Compliance           & 3.55          & 3.82 \textcolor{red!60}{(+\phantom{1}8\%)}          & 3.71         & 4.09 \textcolor{red!60}{(+10\%)}          & 3.56        & 4.04 \textcolor{red!60}{(+13\%)}         & 3.62       & 4.03 \textcolor{red!60}{(+11\%)}        & 3.45        & 3.85 \textcolor{red!60}{(+12\%)}        \\
Consistency \& Detail   Fidelity & 3.29          & 3.62 \textcolor{red!60}{(+10\%)}          & 3.40         & 3.81 \textcolor{red!60}{(+12\%)}          & 3.22        & 3.75 \textcolor{red!60}{(+16\%)}         & 3.37       & 3.70 \textcolor{red!60}{(+10\%)}        & 3.22        & 3.65 \textcolor{red!60}{(+13\%)}        \\
Visual Quality \& Stability      & 3.25          & 3.68 \textcolor{red!60}{(+13\%)}          & 3.30         & 3.75 \textcolor{red!60}{(+14\%)}          & 3.08        & 3.81 \textcolor{red!60}{(+24\%)}         & 3.28       & 3.80 \textcolor{red!60}{(+16\%)}        & 3.23        & 3.65 \textcolor{red!60}{(+13\%)}        \\
Average Score                    & 3.36          & 3.71 \textcolor{red!60}{(+10\%)}          & 3.47         & 3.88 \textcolor{red!60}{(+12\%)}          & 3.29        & 3.86 \textcolor{red!60}{(+17\%)}         & 3.42       & 3.84 \textcolor{red!60}{(+12\%)}        & 3.30        & 3.72 \textcolor{red!60}{(+13\%)}       \\
\bottomrule
\end{tabular}}
\label{tab:ab_fg_only}
\vspace{-1em}
\end{table}

\noindent \textbf{Effectiveness of BAIT, Quality Control, and Background Guidance.} To validate each of our main contributions, we conducted a rigorous video quality comparison using the same 500 source videos and prompts as in Table~\ref{tab:data_quality}, utilizing only foreground Canny edges to control the overall video generation in the final stage, while all other components, \emph{e.g.,} the Wan2.2-Fun-A14B-Control video generator, remain identical. As shown in the FG-Only columns of Table~\ref{tab:ab_fg_only}, the average scores of these videos already exhibit a remarkable improvement over those of OpenVE-3M presented in Table~\ref{tab:data_quality}. Since background guidance is omitted in this setting, these gains stem purely from our more precise BAIT foreground tracking and the strict quality control that prevents prompt misalignment, thereby demonstrating their effectiveness. Furthermore, when introducing background guidance, \emph{i.e.,} the FG+BG columns, the average quality improves substantially. As the only difference between FG-Only and FG+BG is whether background guidance is used during final video synthesis, the resulting 10\%–17\% average gains can be attributed solely to the background guidance, without confounding factors. These remarkable gains indicate that the structural collapse issue has been significantly mitigated by our novel decoupled data generation paradigm.

\begin{table*}[t]
\centering

\begin{minipage}[t][3.2cm][t]{0.49\textwidth}
\centering
\vspace{0pt}
\caption{Performance of Kiwi-Edit trained on different \emph{Sparkle} corpus. \textbf{\textcolor{gray}{Gray}} numbers denote the Kiwi-Edit baseline. \textbf{\textcolor{red!60}{Red}} numbers indicate the absolute gain.}
\label{tab:ab_openve_only}

\vfill

\resizebox{\linewidth}{!}{
\begin{tabular}{lccc}
\toprule
\textbf{Dimensions} & \textbf{Kiwi-Edit} & \textbf{w/ OpenVE-3M} & \textbf{w/ Full Dataset} \\
\midrule
Instruction Compliance & \textcolor{gray}{2.81} & 3.24 \textcolor{red!60}{(+15\%)} & 3.51 \textcolor{red!60}{(+25\%)} \\
Consistency \& Detail Fidelity & \textcolor{gray}{2.58} & 2.92 \textcolor{red!60}{(+13\%)} & 3.15 \textcolor{red!60}{(+22\%)} \\
Visual Quality \& Stability & \textcolor{gray}{2.36} & 2.95 \textcolor{red!60}{(+25\%)} & 3.22 \textcolor{red!60}{(+36\%)} \\
Average Score & \textcolor{gray}{2.58} & 3.04 \textcolor{red!60}{(+18\%)} & 3.29 \textcolor{red!60}{(+28\%)} \\
\bottomrule
\end{tabular}}
\end{minipage}
\hfill
\begin{minipage}[t][3.2cm][t]{0.49\textwidth}
\centering
\vspace{0pt}
\setlength{\tabcolsep}{2pt}
\caption{OpenVE-Bench scores of Kiwi-Edit and Lucy-Edit-1.1 trained on the \emph{Sparkle} OpenVE-3M subset. \textbf{\textcolor{gray}{Gray}} and \textbf{\textcolor{red!60}{Red}} numbers denote the baseline and the absolute gain.}
\label{tab:ab_model_generalizability}

\vfill

\resizebox{\linewidth}{!}{
\begin{tabular}{lccccc}
\toprule
\textbf{Model} & \textbf{w/ OpenVE-3M} & \textbf{Overall} & \textbf{Ins.} & \textbf{Cons.} & \textbf{VQ.} \\
\midrule
\multirow{2}{*}{Kiwi-Edit} & \ding{55} & \textcolor{gray}{2.58} & \textcolor{gray}{2.81} & \textcolor{gray}{2.58} & \textcolor{gray}{2.36} \\
& \ding{51} & 3.04 \textcolor{red!60}{(+18\%)} & 3.24 \textcolor{red!60}{(+15\%)} & 2.92 \textcolor{red!60}{(+13\%)} & 2.95 \textcolor{red!60}{(+25\%)} \\
\midrule
\multirow{2}{*}{Lucy-Edit-1.1} & \ding{55} & \textcolor{gray}{2.10} & \textcolor{gray}{2.41} & \textcolor{gray}{2.03} & \textcolor{gray}{1.86} \\
& \ding{51} & 2.65 \textcolor{red!60}{(+26\%)} & 2.93 \textcolor{red!60}{(+22\%)} & 2.54 \textcolor{red!60}{(+25\%)} & 2.48 \textcolor{red!60}{(+33\%)} \\
\bottomrule
\end{tabular}}
\end{minipage}
\vspace{-0.5em}
\end{table*}

\noindent \textbf{Data Generalizability.} We further evaluate whether the four proposed themes beyond the OpenVE-3M subset are diverse enough to yield universal improvements across most background editing scenarios by comparing a Kiwi-Edit model fine-tuned exclusively on the OpenVE-3M subset against one fine-tuned on the full dataset. The results are presented in Table~\ref{tab:ab_openve_only}. Although the high-quality data within our OpenVE-3M subset already yields a clear gain over the untuned baseline, training on the full dataset, which incorporates broader data not explicitly tailored for this benchmark, achieves a more significant gain (28\% vs 18\%) compared to the subset-only model. These encouraging results demonstrate that \emph{Sparkle} maintains a high level of diversity capable of handling a broad spectrum of background replacements, thereby facilitating generalized performance improvements.

\noindent \textbf{Model Generalizability.} Beyond \emph{data} generalizability, we also test if \emph{Sparkle} has universal benefits across different \emph{models} by fine-tuning Lucy-Edit-1.1~\cite{lucyedit} on the OpenVE-3M subset of \emph{Sparkle}, enabling a direct comparison on OpenVE-Bench with Kiwi-Edit trained on the same subset in Table~\ref{tab:ab_openve_only}. The results are reported in Table~\ref{tab:ab_model_generalizability}. Encouragingly, \emph{Sparkle} yields larger gains for the less capable Lucy-Edit-1.1. Fine-tuning on the OpenVE-3M subset alone raises its score to 2.65, a 26\% improvement, surpassing vanilla Kiwi-Edit trained on millions of videos. These results demonstrate \emph{Sparkle}’s broad effectiveness for general video editors and prove its practical utility.

\begin{figure}[t]
    \centering
    \includegraphics[width=.9\columnwidth]{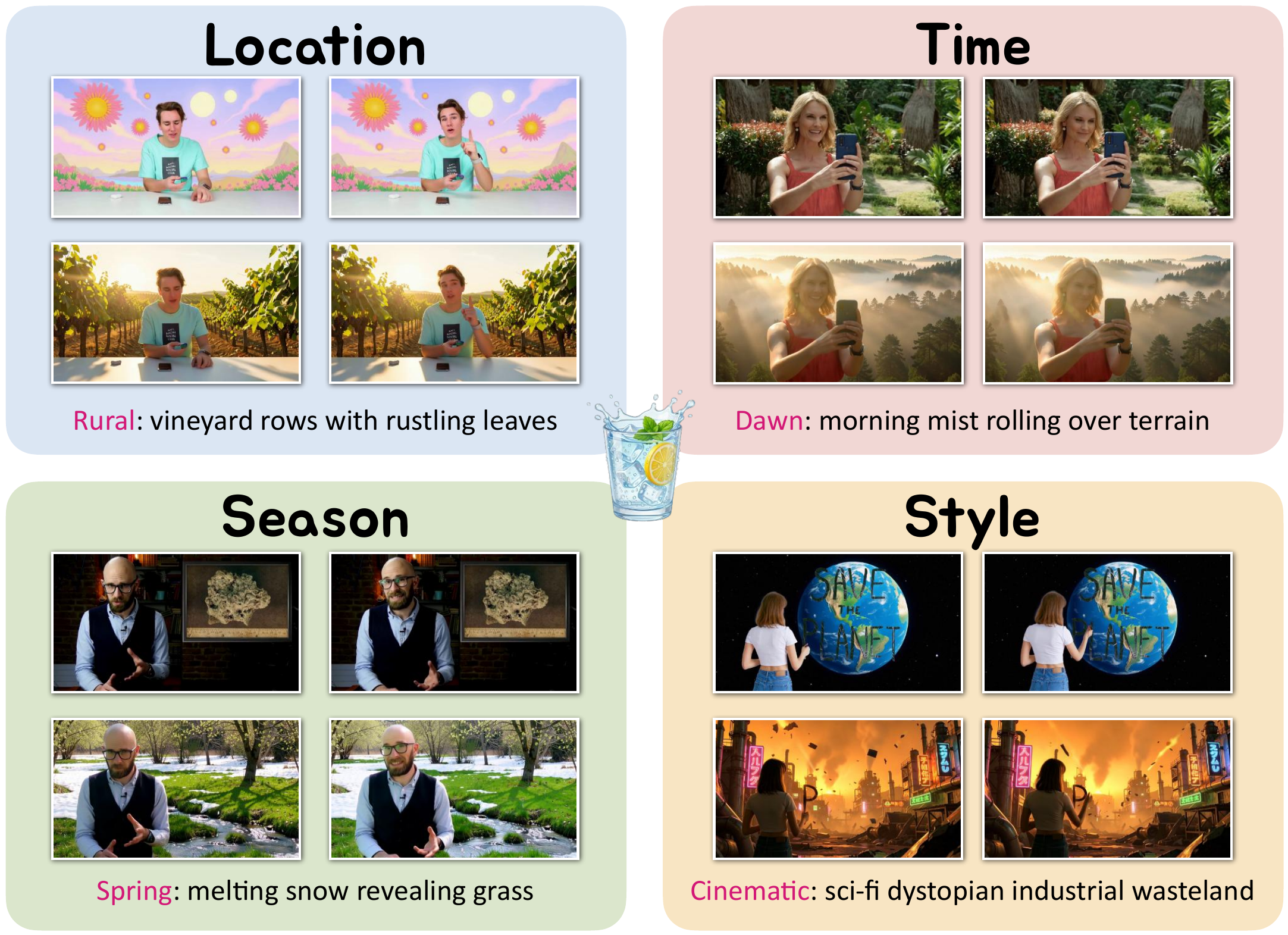}
    \caption{Categorical overview of \emph{Sparkle}. In the subcaptions, purple text indicates subthemes, while black text specifies the corresponding scenes.}
    \label{fig:vis_category}
\vspace{-1em}
\end{figure}

\noindent \textbf{Visualization.} Due to space constraints, we only show a categorical overview of \emph{Sparkle} in Figure~\ref{fig:vis_category} to illustrate its data organization, and provide extensive visualizations for all main experiments in Appendix~\ref{sec:appendix_visualization}. These encompass qualitative comparisons of the original OpenVE-3M against our recreated data (Table~\ref{tab:data_quality}), ablation results evaluating edited videos synthesized using the copy-and-paste paradigm (Table~\ref{tab:ab_cp}) and those using foreground-only guidance (Table~\ref{tab:ab_fg_only}), visual comparisons between the outputs of Kiwi-Edit and \emph{Kiwi-Sparkle} on OpenVE-Bench (Table~\ref{tab:openve_bench}) and \emph{Sparkle-Bench} (Table~\ref{tab:sparkle_bench}), demonstrations of \emph{Kiwi-Sparkle}'s efficacy as a foreground tracker via a trigger phrase, and a failure analysis for \emph{Kiwi-Sparkle}.

\vspace{-1em}
\section{Conclusion}
\vspace{-0.5em}

In this paper, we analyze the limitations of existing video background replacement data, pinpointing how the conventional mixed generation paradigm leads to stale edits. To address this, we propose a novel 5-stage decoupled generation paradigm. By combining our precise BAIT tracking for clean foreground detachment with compatible background video generation, we synthesize high-quality edits via decoupled guidance. Under strict quality control across all stages, we curate the \emph{Sparkle} dataset, which exhibits significant quality boosts over existing data. Furthermore, we introduce \emph{Sparkle-Bench}, encompassing $\sim$100 diverse scenes across 458 videos to advance comprehensive evaluation. Finally, our derivative model, \emph{Kiwi-Sparkle}, demonstrates remarkable gains over existing baselines. We believe this robust infrastructure (dataset, benchmark, and model) will greatly facilitate future research in this demanding area.

\section*{Acknowledgements and Disclosures}
This research is supported by the National Research Foundation, Singapore, under its AI Singapore Programme (AISG Award No: AISG3-RP-2022-030).

\bibliography{reference}
\bibliographystyle{plainnat}

\newpage
\appendix
\section{Coarse Camera Movement Filtering}
\label{sec:appendix_source_video_collection}

Since processing a large volume of source videos using the fine-grained VLM filter introduced in Section~\ref{subsec:source_video_collection} is unacceptably time-consuming, we propose a preliminary coarse filtering approach to efficiently eliminate vast numbers of unqualified videos. As illustrated in Figure~\ref{fig:framework}, Stage 1, we first utilize Unimatch~\cite{unimatch} to compute the optical flow of the source videos at 2 FPS. Subsequently, we apply the coarse-grained filtering strategy as follows:

\emph{Preliminary.} When camera movement occurs between two consecutive frames, the background pixels satisfy:
\begin{equation}
    [x'; y'; 1] \sim H[x; y; 1]
\end{equation}
where $(x,y)$ denotes a point in the first frame, $(x',y')$ is its corresponding position in the second frame, and $H \in \mathbb{R}^{3 \times 3}$ represents the homography matrix. Alternatively, given the optical flow $(u,v)$ computed at $(x,y)$ between the two frames, the anticipated movement is:
\begin{equation}
    (x', y') \approx (x+u, y+v)
\end{equation}
Consequently, using the source points $(x,y)$ and their flow-derived destinations $(x', y')$, we apply the RANSAC~\cite{ransac} algorithm to estimate a robust homography matrix $H$ that models the dominant transformation. We then apply $H$ to compute the transformed coordinates for each point. If the transformed position aligns with the optical flow estimation, the corresponding pixel is classified as background. We define $r$ as the ratio of points satisfying this transformation, where an empirical $r \ge 50\%$ indicates a high probability of global camera movement. Furthermore, we calculate the motion magnitude $m = \sqrt{u^2+v^2}$ at each pixel. We conclude that camera movement exists between the sampled frames only if both $r \ge 50\%$ and the average motion magnitude satisfies $m \ge 1$. If all consecutive sampled frames within a video are determined to be free of camera movement, the sequence is classified as static-camera and retained. This efficient process drastically reduces the overall source videos from 940K to 260K.

\section{Detailed Evaluation Protocol on \emph{Sparkle-Bench}}
\label{sec:appendix_detailed_evaluation_protocol}

In Section~\ref{subsec:sparkle_bench}, we introduce our six-dimensional criteria spanning three perspectives on the proposed \emph{Sparkle-Bench}. As previously stated, we utilize Gemini-2.5-Pro as the scorer due to its exceptional video understanding capabilities. The detailed evaluation criteria are outlined below:

\begin{promptbox}[Background Replacement Scoring Prompt]
You are a data rater specializing in grading video background replacement. You will be given two videos (source and edited) and the editing instruction. Your task is to evaluate the result on a 5-point scale across six dimensions:

\medskip
\noindent\textbf{Instruction Compliance}
\begin{enumerate}[leftmargin=2em, itemsep=1pt, topsep=2pt, label=\arabic*.]
  \item No change, or background entirely unrelated to the prompt, or foreground also replaced/distorted such that the edit fails as a whole.
  \item Background only partially matches prompt content or style; major requested elements wrong or missing; or foreground noticeably altered.
  \item Main background concept matches but with missing/extra elements, wrong sub-style, or partial spill onto the subject.
  \item Requested background fully present and consistent with the prompt; only minor mismatches in tone, detail, or atmosphere.
  \item Background exactly matches the prompt in content, style, mood, and any specified dynamics; foreground untouched.
\end{enumerate}

\medskip
\noindent\textbf{Overall Visual Quality.} This dimension covers global image quality AND foreground-background harmonization. The lighting, color temperature, and shadows on the foreground must match the new background environment. For example, when the prompt changes the time of day (e.g.\ day to night, noon to sunset), keeping the original daytime lighting on the foreground while the background is dark is a major harmonization failure. The same applies to season, location, and style edits that imply different ambient light.
\begin{enumerate}[leftmargin=2em, itemsep=1pt, topsep=2pt, label=\arabic*.]
  \item Severe artefacts throughout (tearing, posterisation, color banding, heavy flicker), OR foreground lighting is grossly inconsistent with the new background (e.g.\ brightly lit subject against a night scene, conflicting light directions, no shadow adaptation).
  \item Clear visual degradation (persistent blur, noise, unstable colors), OR obvious lighting / color-temperature mismatch between foreground and background visible at first glance.
  \item Watchable but with visible flaws on closer look: occasional flicker, mild compression artefacts, soft regions, OR partial harmonization where the foreground tone is in the right direction but not fully matched to the background.
  \item Clean output with only minor issues when zoomed in or paused; foreground lighting and color grading are well aligned with the background, with only subtle discrepancies.
  \item Indistinguishable from real captured footage: sharp, stable, well-graded across the entire clip, with foreground lighting, color temperature, and shadows fully harmonized with the new background environment.
\end{enumerate}

\medskip
\noindent\textbf{Foreground Integrity}
\begin{enumerate}[leftmargin=2em, itemsep=1pt, topsep=2pt, label=\arabic*.]
  \item Foreground severely damaged: missing limbs/parts, large holes, replaced with a different subject, or shape collapsed.
  \item Noticeable foreground damage: partial erosion by background, distorted contours, identity drift across frames.
  \item Foreground mostly preserved but with visible defects: edge halos, slight shape deformation, occasional color bleed.
  \item Foreground well preserved with only minute edge artefacts; shape and identity stable throughout.
  \item Foreground perfectly preserved: every pixel of shape, texture, and identity intact across all frames.
\end{enumerate}

\medskip
\noindent\textbf{Foreground Motion Consistency}
\begin{enumerate}[leftmargin=2em, itemsep=1pt, topsep=2pt, label=\arabic*.]
  \item Foreground motion completely different from source: actions replaced, frozen, looped, or temporally scrambled.
  \item Major motion deviations: different gestures, dropped actions, or strong temporal jitter not present in source.
  \item Same general action is recognizable but with timing drift, trajectory shifts, or inconsistent speed versus source.
  \item Motion closely tracks the source with only minor temporal misalignment or subtle smoothing.
  \item Foreground motion is identical to the source video in trajectory, timing, and articulation, frame by frame.
\end{enumerate}

\medskip
\noindent\textbf{Background Dynamics (Liveness).} This dimension measures whether the background motion matches the intensity and character implied by the prompt. The bar is appropriateness to the prompt, not absolute amount of motion. A ``gentle swaying grass'' prompt rendered as subtle wind-like sway is fully correct and should receive a high score; the same subtle motion for a ``rushing waterfall'' prompt is severely under-rendered.
\begin{enumerate}[leftmargin=2em, itemsep=1pt, topsep=2pt, label=\arabic*.]
  \item Background motion contradicts the prompt: completely static when the prompt implies any motion, or wrong type/direction of motion (e.g.\ crashing waves rendered as a still pond).
  \item Motion intensity is far below what the prompt implies (e.g.\ a ``rushing river'' rendered as barely moving water), or required dynamics are largely absent.
  \item Motion type is in the right direction but noticeably under- or over-rendered, OR motion exists but feels stiff and unnatural.
  \item Motion intensity and character are well matched to the prompt, with only minor stiffness, small frozen patches, or slight over/under rendering.
  \item Background motion perfectly matches the prompt in both intensity and character, rendered naturally and continuously throughout the clip --- gentle prompts receive gentle motion, energetic prompts receive energetic motion.
\end{enumerate}
\textit{Special case:} if the prompt explicitly asks for a static background (e.g.\ ``still photo'', ``frozen scene'', ``no motion''), a faithfully static background scores 5 and any unwanted motion lowers the score accordingly.

\medskip
\noindent\textbf{Background Visual Quality}
\begin{enumerate}[leftmargin=2em, itemsep=1pt, topsep=2pt, label=\arabic*.]
  \item Background severely degraded: melting structures, broken geometry, heavy blur, or incoherent textures.
  \item Clear distortion or blur in major background regions; structures wobble or warp over time.
  \item Acceptable background with visible imperfections: soft textures, mild geometric inconsistency, minor temporal warping.
  \item High-quality background with only minor issues on close inspection; geometry and textures stable.
  \item Background is sharp, geometrically coherent, and temporally stable; on par with real footage.
\end{enumerate}

\medskip
\noindent\textbf{Constraints.} The scores for Overall Visual Quality, Foreground Integrity, Foreground Motion Consistency, Background Dynamics, and Background Visual Quality must not exceed the score for Instruction Compliance.

\medskip
\noindent\textbf{Example Response Format.}
\begin{itemize}[leftmargin=2em, itemsep=1pt, topsep=2pt, label=--]
  \item Brief reasoning: No more than 30 words.
  \item Instruction Compliance: 1--5.
  \item Overall Visual Quality: 1--5.
  \item Foreground Integrity: 1--5.
  \item Foreground Motion Consistency: 1--5.
  \item Background Dynamics: 1--5.
  \item Background Visual Quality: 1--5.
\end{itemize}

\medskip
Editing instruction is: \texttt{\{edit\_prompt\}}.

Below are the videos before and after editing:
\end{promptbox}

As outlined above, rather than merely outputting the dimensional scores, we intentionally prompt Gemini-2.5-Pro to generate a brief rationale via chain-of-thought reasoning prior to its final response, thereby yielding more accurate and reliable evaluation results.

\section{Additional Experiments}

\subsection{Theme-specific Results on \emph{Sparkle-Bench}}
\label{subsec:appendix_full_results_on_sparkle_bench}

\begin{table}[h]
\setlength{\tabcolsep}{8pt}
\centering
\caption{Scores on the \emph{Location} theme of \emph{Sparkle-Bench}.}
\resizebox{.9\columnwidth}{!}{
\begin{tabular}{lcccccccc}
\toprule
\multirow{2}{*}{\textbf{Model}} & \multirow{2}{*}{\textbf{Configuration}} & \multirow{2}{*}{\textbf{Overall}} & \multicolumn{2}{c}{\textbf{Global}} & \multicolumn{2}{c}{\textbf{Foreground}} & \multicolumn{2}{c}{\textbf{Background}} \\
\cmidrule(l){4-9}
                                &                                         &                                   & Ins.             & Vis.             & FgIn.              & FgMo.              & BgDy.              & BgVi.              \\
\midrule
InsViE~\cite{insvie}                          & 2B-25F@480P                             & 1.00                              & 1.00             & 1.00             & 1.00               & 1.00               & 1.00               & 1.00               \\
ICVE~\cite{icve}                            & 13B-81F@480P                            & 2.01                              & 2.28             & 1.59             & 2.18               & 2.27               & 1.71               & 2.05               \\
DITTO~\cite{ditto}                           & 14B-81F@480P                            & 2.00                              & 2.14             & 1.95             & 1.97               & 2.08               & 1.74               & 2.11               \\
OmniVideo2~\cite{omnivideo2}                      & A14B-41F@480P                           & 2.42                              & 2.67             & 2.23             & 2.47               & 2.56               & 1.98               & 2.63               \\
UniVideo~\cite{omnivideo2}                        & 13B-81F@480P                            & 2.20                              & 2.47             & 1.59             & 2.31               & 2.47               & 1.93               & 2.41               \\
Kiwi-Edit~\cite{kiwiedit}                       & 5B-81F@720P                             & 2.53                              & 2.92             & 2.12             & 2.88               & 2.90               & 1.51               & 2.86               \\
Lucy-Edit-1.1~\cite{lucyedit}                   & 5B-81F@720P                             & 2.86                              & 3.22             & 2.28             & 2.94               & 3.19               & 2.55               & 3.00               \\
\midrule
\rowcolor{shallow_blue}
Kiwi-Sparkle (Ours)                    & 5B-81F@720P                             & \textbf{3.84}                              & \textbf{4.10}             & \textbf{3.47}             & \textbf{3.83}               & \textbf{4.06}               & \textbf{3.57}               & \textbf{4.00}       \\
\bottomrule
\end{tabular}}
\label{tab:appendix_sparkle_bench_location}
\end{table}

\begin{table}[h]
\setlength{\tabcolsep}{8pt}
\centering
\caption{Scores on the \emph{Season} theme of \emph{Sparkle-Bench}.}
\resizebox{.9\columnwidth}{!}{
\begin{tabular}{lcccccccc}
\toprule
\multirow{2}{*}{\textbf{Model}} & \multirow{2}{*}{\textbf{Configuration}} & \multirow{2}{*}{\textbf{Overall}} & \multicolumn{2}{c}{\textbf{Global}} & \multicolumn{2}{c}{\textbf{Foreground}} & \multicolumn{2}{c}{\textbf{Background}} \\
\cmidrule(l){4-9}
                                &                                         &                                   & Ins.             & Vis.             & FgIn.              & FgMo.              & BgDy.              & BgVi.              \\
\midrule
InsViE~\cite{insvie}                          & 2B-25F@480P                             & 1.05                              & 1.07             & 1.02             & 1.05               & 1.07               & 1.04               & 1.04               \\
ICVE~\cite{icve}                            & 13B-81F@480P                            & 2.00                              & 2.37             & 1.46             & 2.22               & 2.36               & 1.52               & 2.06               \\
DITTO~\cite{ditto}                           & 14B-81F@480P                            & 2.00                              & 2.14             & 1.97             & 1.93               & 2.12               & 1.75               & 2.11               \\
OmniVideo2~\cite{omnivideo2}                      & A14B-41F@480P                           & 2.22                              & 2.44             & 2.04             & 2.25               & 2.33               & 1.82               & 2.41               \\
UniVideo~\cite{univideo}                        & 13B-81F@480P                            & 2.60                              & 3.01             & 1.83             & 2.83               & 3.01               & 2.12               & 2.83               \\
Kiwi-Edit~\cite{kiwiedit}                       & 5B-81F@720P                             & 2.55                              & 2.97             & 2.12             & 2.87               & 2.95               & 1.51               & 2.88               \\
Lucy-Edit-1.1~\cite{lucyedit}                   & 5B-81F@720P                             & 2.68                              & 3.07             & 2.15             & 2.67               & 3.04               & 2.41               & 2.76               \\
\midrule
\rowcolor{shallow_blue}
Kiwi-Sparkle (Ours)                   & 5B-81F@720P                             & \textbf{3.79}                              & \textbf{4.14}             & \textbf{3.34}             & \textbf{3.77}               & \textbf{4.06}               & \textbf{3.39}               & \textbf{4.03}   \\
\bottomrule
\end{tabular}}
\label{tab:appendix_sparkle_bench_season}
\end{table}

\begin{table}[h]
\setlength{\tabcolsep}{8pt}
\centering
\caption{Scores on the \emph{Time} theme of \emph{Sparkle-Bench}.}
\resizebox{.9\columnwidth}{!}{
\begin{tabular}{lcccccccc}
\toprule
\multirow{2}{*}{\textbf{Model}} & \multirow{2}{*}{\textbf{Configuration}} & \multirow{2}{*}{\textbf{Overall}} & \multicolumn{2}{c}{\textbf{Global}} & \multicolumn{2}{c}{\textbf{Foreground}} & \multicolumn{2}{c}{\textbf{Background}} \\
\cmidrule(l){4-9}
                                &                                         &                                   & Ins.             & Vis.             & FgIn.              & FgMo.              & BgDy.              & BgVi.              \\
\midrule
InsViE~\cite{insvie}                          & 2B-25F@480P                             & 1.13                              & 1.20             & 1.06             & 1.12               & 1.19               & 1.09               & 1.11               \\
ICVE~\cite{icve}                            & 13B-81F@480P                            & 1.87                              & 2.19             & 1.39             & 2.08               & 2.18               & 1.47               & 1.92               \\
DITTO~\cite{ditto}                           & 14B-81F@480P                            & 2.17                              & 2.38             & 1.96             & 2.16               & 2.34               & 1.84               & 2.34               \\
OmniVideo2~\cite{omnivideo2}                      & A14B-41F@480P                           & 2.20                              & 2.45             & 1.99             & 2.33               & 2.36               & 1.67               & 2.40               \\
UniVideo~\cite{univideo}                        & 13B-81F@480P                            & 2.33                              & 2.63             & 1.83             & 2.56               & 2.63               & 1.77               & 2.53               \\
Kiwi-Edit~\cite{kiwiedit}                       & 5B-81F@720P                             & 2.48                              & 2.89             & 2.08             & 2.83               & 2.88               & 1.48               & 2.73               \\
Lucy-Edit-1.1~\cite{lucyedit}                   & 5B-81F@720P                             & 2.62                              & 2.94             & 2.14             & 2.77               & 2.93               & 2.19               & 2.76               \\
\midrule
\rowcolor{shallow_blue}
Kiwi-Sparkle (Ours)                    & 5B-81F@720P                             & \textbf{3.69}                              & \textbf{4.01}             & \textbf{3.14}             & \textbf{3.66}               & \textbf{3.99}               & \textbf{3.47}               & \textbf{3.87}               \\
\bottomrule
\end{tabular}}
\label{tab:appendix_sparkle_bench_time}
\end{table}

\begin{table}[h]
\setlength{\tabcolsep}{8pt}
\centering
\caption{Scores on the \emph{Style} theme of \emph{Sparkle-Bench}.}
\resizebox{.9\columnwidth}{!}{
\begin{tabular}{lcccccccc}
\toprule
\multirow{2}{*}{\textbf{Model}} & \multirow{2}{*}{\textbf{Configuration}} & \multirow{2}{*}{\textbf{Overall}} & \multicolumn{2}{c}{\textbf{Global}} & \multicolumn{2}{c}{\textbf{Foreground}} & \multicolumn{2}{c}{\textbf{Background}} \\
\cmidrule(l){4-9}
                                &                                         &                                   & Ins.             & Vis.             & FgIn.              & FgMo.              & BgDy.              & BgVi.              \\
\midrule
InsViE~\cite{insvie}                 & 2B-25F@480P                    & 1.04                     & 1.05         & 1.04        & 1.04           & 1.05          & 1.04           & 1.05          \\
ICVE~\cite{icve}                   & 13B-81F@480P                   & 1.90                     & 2.15         & 1.50        & 2.05           & 2.14          & 1.63           & 1.95          \\
DITTO~\cite{ditto}                  & 14B-81F@480P                   & 1.85                     & 1.95         & 1.87        & 1.68           & 1.91          & 1.75           & 1.95          \\
OmniVideo2~\cite{omnivideo2}             & A14B-41F@480P                  & 2.57                     & 2.76         & 2.41        & 2.65           & 2.66          & 2.20           & 2.75          \\
UniVideo~\cite{univideo}               & 13B-81F@480P                   & 2.52                     & 2.75         & 2.02        & 2.65           & 2.75          & 2.27           & 2.71          \\
Kiwi-Edit~\cite{kiwiedit}              & 5B-81F@720P                    & 2.60                     & 2.92         & 2.29        & 2.85           & 2.88          & 1.80           & 2.88          \\
Lucy-Edit-1.1~\cite{lucyedit}          & 5B-81F@720P                    & 2.77                     & 3.02         & 2.35        & 2.74           & 3.00          & 2.71           & 2.82          \\
\midrule
\rowcolor{shallow_blue}
Kiwi-Sparkle (Ours)           & 5B-81F@720P                    & \textbf{3.92}                     & \textbf{4.15}         & \textbf{3.65}        & \textbf{3.80}           & \textbf{4.10}          & \textbf{3.75}           & \textbf{4.05}          \\
\bottomrule
\end{tabular}}
\label{tab:appendix_sparkle_bench_style}
\end{table}

In addition to the overall scores on \emph{Sparkle-Bench} (Table~\ref{tab:sparkle_bench}), we provide theme-specific results for \emph{Location}, \emph{Season}, \emph{Time}, and \emph{Style} in Tables~\ref{tab:appendix_sparkle_bench_location}, \ref{tab:appendix_sparkle_bench_season}, \ref{tab:appendix_sparkle_bench_time}, and \ref{tab:appendix_sparkle_bench_style}, respectively. Across all dimensions, \emph{Kiwi-Sparkle} demonstrates exceptional instruction-following capabilities, emerging as the only model to surpass 4.0/5.0. This indicates that almost all required elements are accurately synthesized during editing, a conclusion further supported by its high \emph{Background Dynamics} (BgDy) and \emph{Background Visual Quality} (BgVi) scores. These encouraging results firmly validate the high data quality of \emph{Sparkle}. Simultaneously, the foreground is well-preserved with consistent motion. The related metrics, namely \emph{Foreground Integrity} (FgIn) and \emph{Foreground Motion Consistency} (FgMo), remain close to 4.0. This strongly proves the effectiveness of our BAIT algorithm in imparting precise foreground knowledge to downstream models.

Among all themes, \emph{Time} proves to be the most challenging. Most models, including our \emph{Kiwi-Sparkle}, yield their lowest scores on this theme, indicating that light and shadow adjustments still leave room for improvement. Nevertheless, even in this challenging scenario, \emph{Kiwi-Sparkle} surpasses the SOTA model, \emph{i.e.,} Lucy-Edit-1.1, by approximately 41\%. This demonstrates that our rigorous data pipeline significantly contributes to achieving more harmonious edits. Conversely, most models achieve their highest scores on the \emph{Style} theme. This suggests that the knowledge acquired from abundant global style transfer data can somewhat generalize to style-oriented background editing, an observation that warrants future investigation. In summary, \emph{Sparkle} facilitates a balanced refinement across all themes, making it highly suitable as a post-training corpus to enhance background replacement capabilities.

\subsection{Human Evaluation and Cross-evaluator Validation}
\label{subsec:appendix_human_cross_validation}

\begin{table}[h]
\centering
\caption{The cross-evaluator validation results on \emph{Sparkle-Bench}.}
\resizebox{.9\columnwidth}{!}{
\begin{tabular}{llccccccc}
\toprule
\multicolumn{1}{c}{\multirow{2}{*}{\textbf{Judge}}} & \multicolumn{1}{c}{\multirow{2}{*}{\textbf{Model}}} & \multirow{2}{*}{\textbf{Overall}} & \multicolumn{2}{c}{\textbf{Global}} & \multicolumn{2}{c}{\textbf{Foreground}} & \multicolumn{2}{c}{\textbf{Background}} \\
\cmidrule(l){4-9}
\multicolumn{1}{c}{} & \multicolumn{1}{c}{} & & Ins. & Vis. & FgIn. & FgMo. & BgDy. & BgVi. \\
\midrule
\multirow{2}{*}{Gemini-2.5-Pro} & Kiwi-Edit & 2.54 & 2.92 & 2.15 & 2.86 & 2.90 & 1.57 & 2.84 \\
& Kiwi-Sparkle (Ours) & \textbf{3.81} & \textbf{4.10} & \textbf{3.40} & \textbf{3.77} & \textbf{4.05} & \textbf{3.54} & \textbf{3.99} \\
\midrule
\multirow{2}{*}{Gemini-3.6-Flash} & Kiwi-Edit & 3.08 & 3.33 & 3.13 & 3.24 & 3.17 & 2.32 & 3.30 \\
& Kiwi-Sparkle (Ours) & \textbf{4.36} & \textbf{4.53} & \textbf{4.22} & \textbf{4.30} & \textbf{4.20} & \textbf{4.39} & \textbf{4.51} \\
\bottomrule
\end{tabular}}
\label{tab:appendix_cross_validator}
\end{table}

Beyond the Gemini-2.5-Pro-based evaluations in our main experiments, we also conduct human evaluation and cross-evaluator validation to further strengthen our evaluation reliability.

\begin{wraptable}{r}{0.55\columnwidth}
\centering
\vspace{-1.5em}
\caption{Human evaluation results on \emph{Sparkle-Bench}.}
\resizebox{\linewidth}{!}{
\begin{tabular}{lccc}
\toprule
\textbf{Options}   & Kiwi-Edit & Kiwi-Sparkle & Tied \\
\midrule
\textbf{Selection Rate} & 5\%       & 93\%         & 2\%  \\
\bottomrule
\end{tabular}}
\label{tab:appendix_human_evaluation}
\vspace{-1em}
\end{wraptable}
To be specific, the human evaluation is conducted by randomly sampling two videos per \emph{Sparkle-Bench} theme, yielding 194 videos across 97 scenes. Five annotators compare Kiwi-Sparkle and Kiwi-Edit~\cite{kiwiedit}, select the best using three options: (\textbf{a}) Kiwi-Edit, (\textbf{b}) Kiwi-Sparkle, or (\textbf{c}) Tie. The average selection rates are listed in Table~\ref{tab:appendix_human_evaluation}. The strong human preference for Kiwi-Sparkle is consistent with Table~\ref{tab:sparkle_bench}, supporting our six-dimensional evaluation protocol and the reliability of Gemini-2.5-Pro.

For cross-evaluator validation, we replace Gemini-2.5-Pro with Gemini-3.6-Flash, as other advanced models such as GPT currently lack a native API for video input. The overall \emph{Sparkle-Bench} scores are reported in Table~\ref{tab:appendix_cross_validator}. Gemini-3.6-Flash yields the same performance trend as Gemini-2.5-Pro, further supporting the reliability of our evaluation protocol.

\subsection{Robustness of BAIT}

\begin{table}[h]
\setlength{\tabcolsep}{8pt}
\centering
\caption{Evaluation of BAIT's Robustness under 50 occlusion cases. \textbf{\textcolor{red!60}{Red}} numbers indicate the absolute gain. Ins, Cons, and VQ stand for \emph{Instruction Compliance}, \emph{Consistency \& Detail Fidelity}, and \emph{Visual Quality \& Stability}, respectively.}
\resizebox{.9\columnwidth}{!}{
\begin{tabular}{lccccc}
\toprule
\textbf{Tracking   Method} & \textbf{Selection Rate} & \textbf{Overall}     & \textbf{Ins.}        & \textbf{Cons.}        & \textbf{VQ.}         \\
\midrule
Single-Frame      & 6\%            & 2.44        & 2.90        & 2.31         & 2.12        \\
BAIT              & 94\%           & 2.64 \textcolor{red!60}{(+8\%)} & 3.05 \textcolor{red!60}{(+5\%)} & 2.58 \textcolor{red!60}{(+12\%)} & 2.29 \textcolor{red!60}{(+8\%)} \\
\bottomrule
\end{tabular}}
\label{tab:appendix_bait_robustness}
\end{table}

During data curation of \emph{Sparkle}, we neither assume a single foreground object per video nor apply such filtering, and many videos contain multiple independently moving foreground objects with mutual occlusions. To prove BAIT's robustness under these cases, we further provide quantitative evidence through \emph{human evaluation} and \emph{Gemini scoring}. From the 500 videos used in the OpenVE-3M ablations in Table~\ref{tab:data_quality}, we select 50 samples containing occlusions. We then: (\textbf{i}) ask five annotators to compare masks produced by BAIT and first-frame tracking; and (\textbf{ii}) copy the extracted foregrounds and paste them onto the corresponding backgrounds and ask Gemini to assess their quality (same as the Copy-and-Paste setting in Table~\ref{tab:ab_cp}). 

The results are reported in Table~\ref{tab:appendix_bait_robustness}. BAIT consistently outperforms first-frame tracking. In contrast, first-frame tracking degrades substantially, producing unstable masks, failing to suppress artifacts like temporal glitches, and occasionally missing foreground regions, as shown in Figure~\ref{fig:bait}, leading to lower human selection rates and Gemini scores. These results suggest that BAIT remains robust under occlusion, as it uses timestamp-specific bounding boxes generated by Qwen3-VL-32B, which can naturally localize multiple foreground objects, \emph{e.g.,} the fine-grained boxes around the woman in Figure~\ref{fig:framework}, Stage 3. Its majority voting improves robustness to occlusion: \emph{even if an object is occluded at some timestamps, masks from other temporal anchors recover the lost information.} It also suppresses unstable masks from single-frame tracking, since anchors at different timestamps capture complementary object shapes and improve temporal consistency.

\subsection{Sensitivity of EditScore Threshold across Stages}

\begin{table}[h]
\setlength{\tabcolsep}{8pt}
\centering
\caption{The bad-edit rates \emph{w.r.t.} different EditScore filtering thresholds across stages. BG and FG refer to background and foreground, respectively. Default values are highlighted.}
\resizebox{.9\columnwidth}{!}{
\begin{tabular}{cccc}
\toprule
\textbf{Threshold} & \textbf{Stage 2:} \emph{BG Replacement} & \textbf{Stage 3:} \emph{FG Removal} & \textbf{Stage 5:} \emph{Video Synthesis} \\
\midrule
7.0       & 21\%                            & 30\%                        & 24\%                                             \\
7.5       & 14\%                            & 18\%                        & 15\%                                             \\
8.0       & \textbf{5\% (default)}                   & 9\%                         & \textbf{5\% (default)}                                    \\
8.5       & 3\%                             & \textbf{4\% (default)}               & 4\%     \\      
\bottomrule
\end{tabular}}
\label{tab:appendix_ab_editscore}
\end{table}

As illustrated in Figure~\ref{fig:framework}, we apply EditScore~\cite{editscore} to filter out unqualified samples across multiple stages, and the filtering threshold is crucial for controlling prompt alignment. To find the optimal threshold values, we conducted human evaluation ablations at each filtering stage involving EditScore. Specifically, at each stage, we first randomly sampled 100 edited images or videos with EditScore above 7.0, as lower-scoring samples typically contained severe artifacts. Then we evaluated thresholds at 0.5 intervals and manually measured the bad-edit rate after filtering. The detailed results are shown in Table~\ref{tab:appendix_ab_editscore}. We find that EditScore effectively captures overall edit quality. A threshold of 8.0 is sufficient for Stages 2 and 5, which mainly involve image- and video-level background replacement. However, partially successful removal may still receive a high score despite visible artifacts. We therefore adopt a stricter threshold of 8.5 in Stage 3 to ensure higher data quality. These values guarantee that the bad-edit rates are below 5\%.

\subsection{Longer-video Consistency}
Our curated Sparkle videos contain 65–81 frames, which matches the standard maximum length of current open-source video editors in Table~\ref{tab:openve_bench}. Longer-sequence training exceeds an 80GB H100, making 81 frames a practical setting. Still, we believe that longer-video consistency is an important research topic, and we conduct two frame-extension experiments covering both data generation and downstream evaluation as a preliminary exploration.

Specifically, for data generation, we slow down the 500 videos used in the Location-subset ablations to 101/121 frames by repeating some frames, then process them with the Sparkle pipeline. Longer sequences will exceed the memory capacity of an 80GB H100. We evaluate the outputs using the same Gemini protocol as in Table~\ref{tab:data_quality}. For downstream evaluation, we use the same approach to extend the 108 videos in the Sparkle-Bench Location theme to 101/121 frames and test Kiwi-Sparkle's editing performance. 

\begin{wraptable}{r}{0.5\columnwidth}
\setlength{\tabcolsep}{8pt}
\centering
\vspace{-2em}
\caption{The longer-video consistency results \emph{w.r.t} data generation and downstream evaluation.}
\resizebox{\linewidth}{!}{
\begin{tabular}{lcc}
\toprule
\textbf{Stage}                            & \textbf{Frames} & \textbf{Overall}     \\
\midrule
\multirow{3}{*}{Data Generation} & 81     & 3.88        \\
                                 & 101    & 3.81 (-2\%) \\
                                 & 121    & 3.73 (-4\%) \\
\midrule
\multirow{3}{*}{Downstream Evaluation}      & 81     & 3.84        \\
                                 & 101    & 3.71 (-3\%) \\
                                 & 121    & 3.58 (-7\%) \\
\bottomrule
\end{tabular}}
\label{tab:appendix_longer_video}
\vspace{-1em}
\end{wraptable}
The aggregated results are listed in Table~\ref{tab:appendix_longer_video}. Extending data generation to 121 frames causes a moderate drop, likely because Wan2.2 was trained on similar lengths. Kiwi-Sparkle degrades more, as it was trained on at most 81 frames, but the decline remains moderate. Longer experiments are currently limited by Wan2.2 and GPU memory. Importantly, \emph{Sparkle’s decoupled guidance generation pipeline is not inherently length-limited}; the constraint comes from the current Stage 3/5 generator, \emph{i.e.,} Wan2.2, whose recommended setting is 81 frames. Future long-video generators can serve as drop-in replacements.

\clearpage

\section{Visualization}
\label{sec:appendix_visualization}

\subsection{Comparison between OpenVE-3M and \emph{Sparkle}}

\begin{figure}[h]
    \centering
    \includegraphics[width=1.\columnwidth]{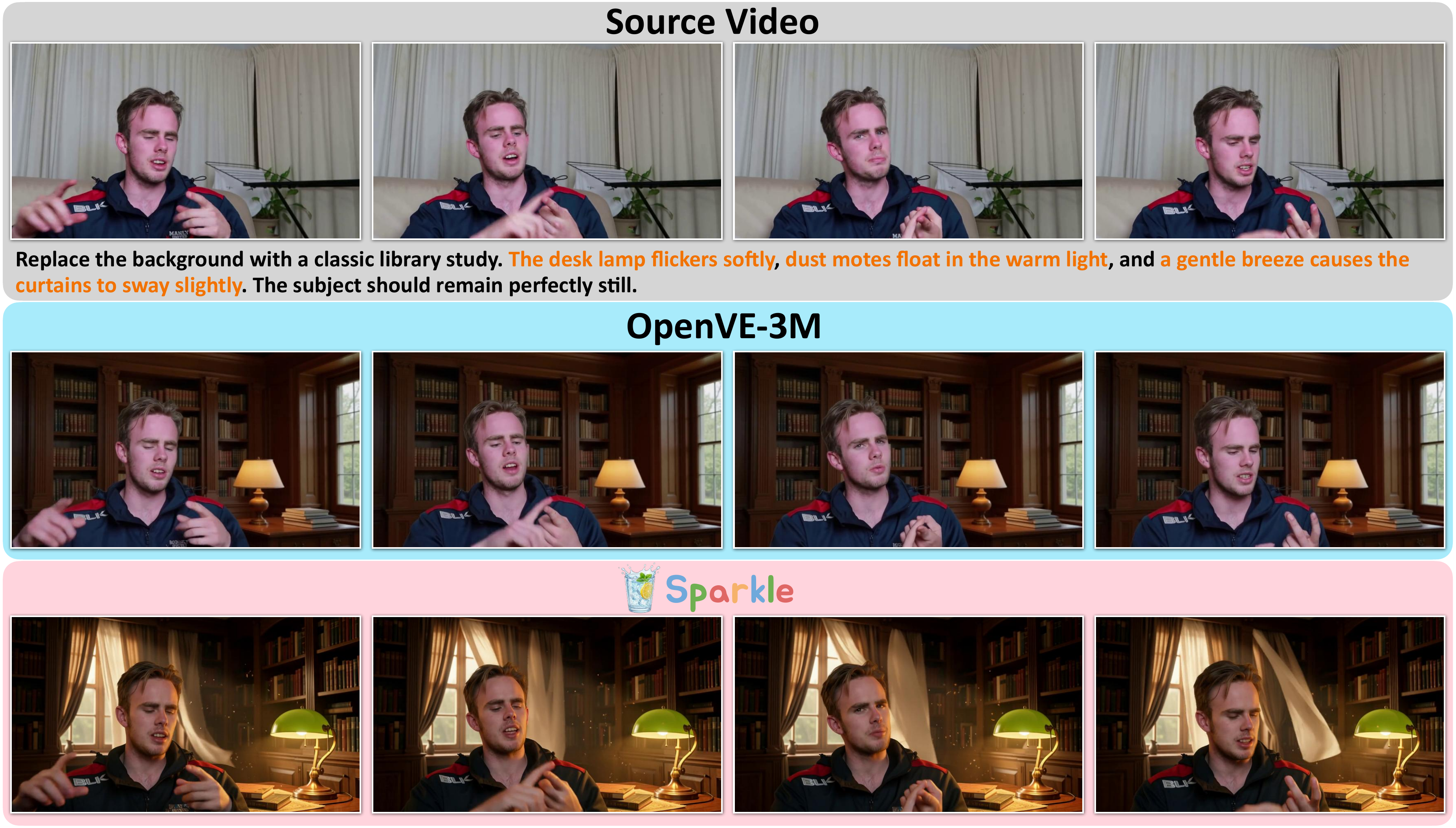}
    \caption{Data comparison between OpenVE-3M~\cite{openve3m} and our proposed \emph{Sparkle}-Part1.}
    \label{fig:appendix_data_compare_openve3m_sparkle_p1}
\end{figure}

\begin{figure}[h]
    \centering
    \includegraphics[width=1.\columnwidth]{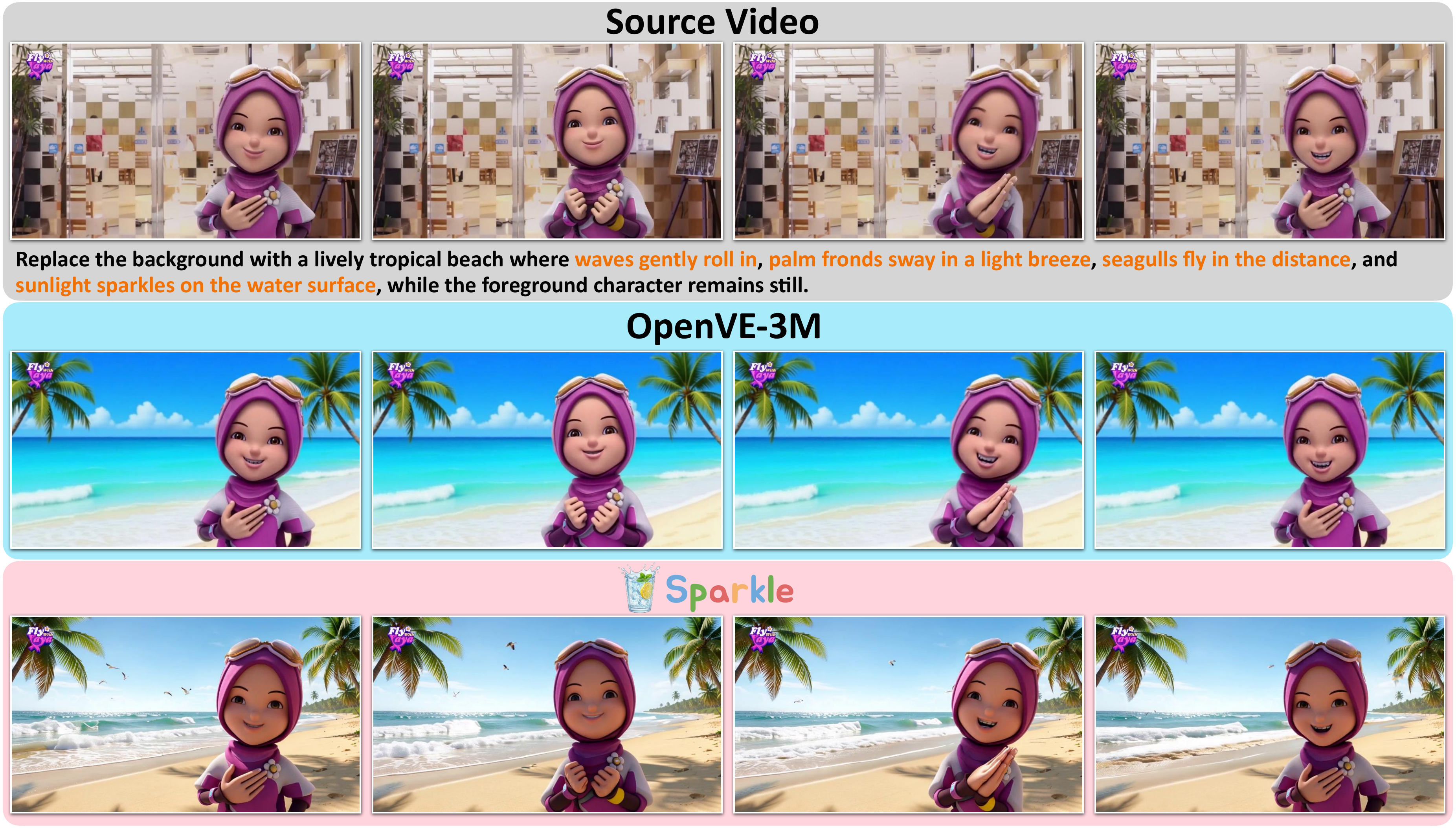}
    \caption{Data comparison between OpenVE-3M~\cite{openve3m} and our proposed \emph{Sparkle}-Part2.}
    \label{fig:appendix_data_compare_openve3m_sparkle_p2}
\end{figure}

\begin{figure}[h]
    \centering
    \includegraphics[width=1.\columnwidth]{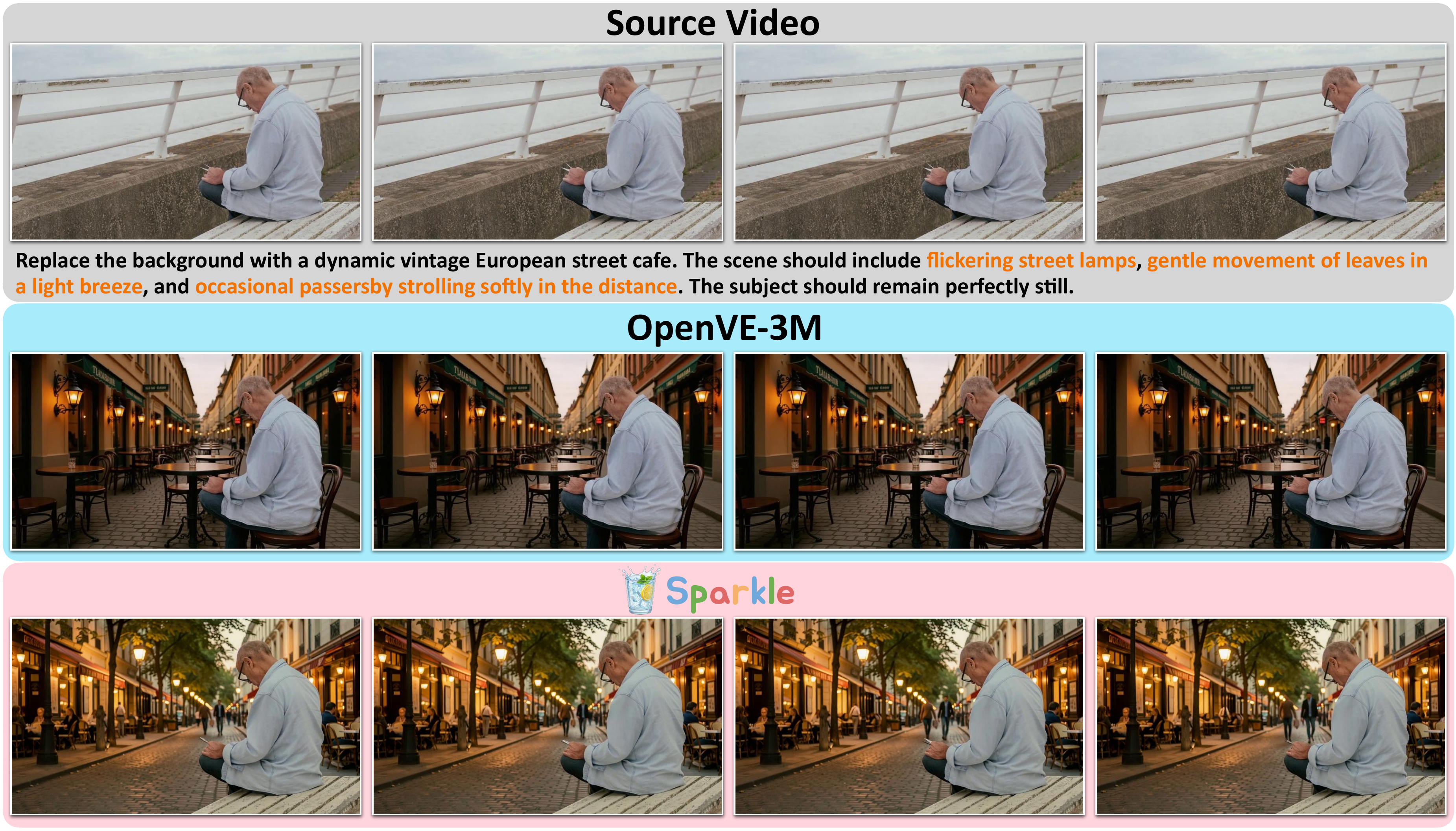}
    \caption{Data comparison between OpenVE-3M~\cite{openve3m} and our proposed \emph{Sparkle}-Part3.}
    \label{fig:appendix_data_compare_openve3m_sparkle_p3}
\end{figure}

\begin{figure}[h]
    \centering
    \includegraphics[width=1.\columnwidth]{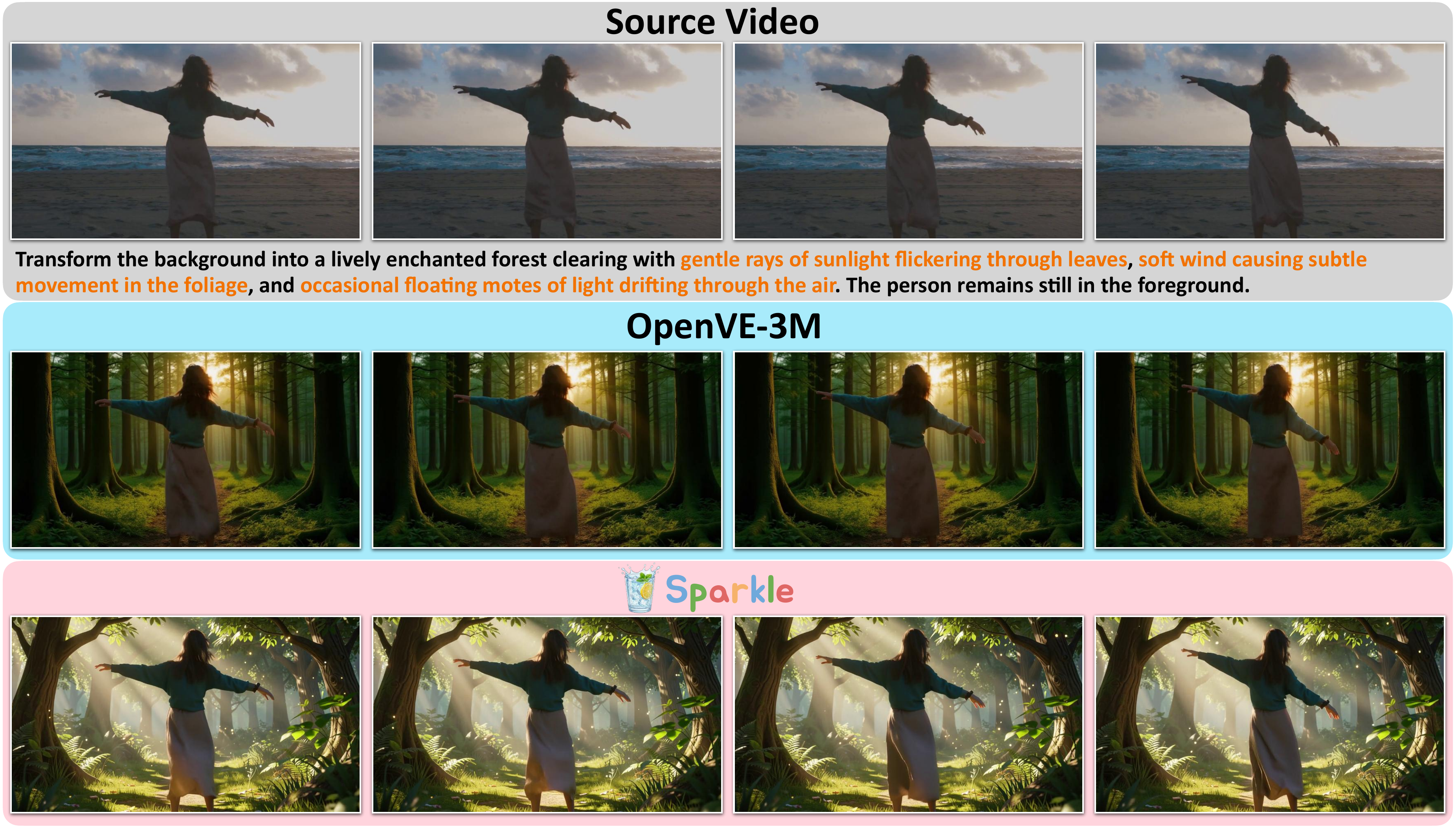}
    \caption{Data comparison between OpenVE-3M~\cite{openve3m} and our proposed \emph{Sparkle}-Part4.}
    \label{fig:appendix_data_compare_openve3m_sparkle_p4}
\end{figure}

Beyond the statistical data quality comparison in Table~\ref{tab:data_quality}, we provide intuitive visual comparisons of edits derived from identical source videos and prompts in Figures~\ref{fig:appendix_data_compare_openve3m_sparkle_p1}, \ref{fig:appendix_data_compare_openve3m_sparkle_p2}, \ref{fig:appendix_data_compare_openve3m_sparkle_p3}, and \ref{fig:appendix_data_compare_openve3m_sparkle_p4}. We clearly observe that OpenVE-3M suffers severely from \emph{Prompt Misalignment}. For instance, crucial elements such as the swaying curtains (Figure~\ref{fig:appendix_data_compare_openve3m_sparkle_p1}), flying seagulls (Figure~\ref{fig:appendix_data_compare_openve3m_sparkle_p2}), strolling passersby (Figure~\ref{fig:appendix_data_compare_openve3m_sparkle_p3}), and floating motes (Figure~\ref{fig:appendix_data_compare_openve3m_sparkle_p4}) are entirely missing. Furthermore, the backgrounds in the OpenVE-3M videos remain unnaturally static, indicating that relying solely on foreground guidance often fails to generate proper dynamics. In contrast, benefiting from our novel decoupled generation paradigm and rigorous quality control, all requested elements are faithfully rendered in our edits. Simultaneously, the backgrounds maintain dynamic realism, such as rolling waves, in a harmonious manner, significantly boosting the overall data quality.

\clearpage

\subsection{Comparison with Videos Synthesized by Copy-and-Paste} 

\begin{figure}[h]
    \centering
    \includegraphics[width=1.\columnwidth]{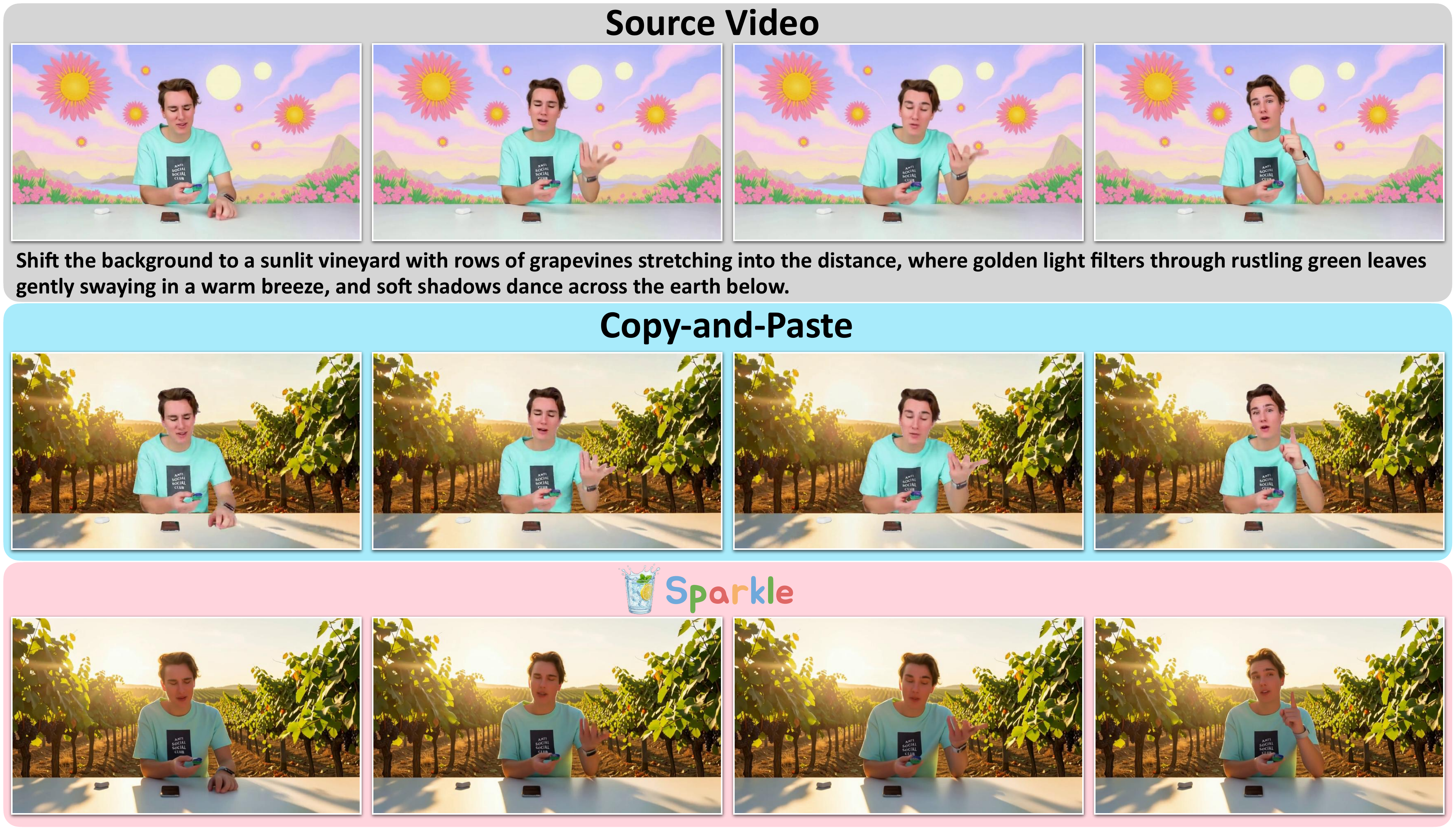}
    \caption{Data comparison between Copy-and-Paste and our proposed \emph{Sparkle}. The theme, subtheme, and scene are ``Location-rural-vineyard rows with rustling leaves''.}
    \label{fig:appendix_data_compare_copy_and_paste_location}
\end{figure}

\begin{figure}[h]
    \centering
    \includegraphics[width=1.\columnwidth]{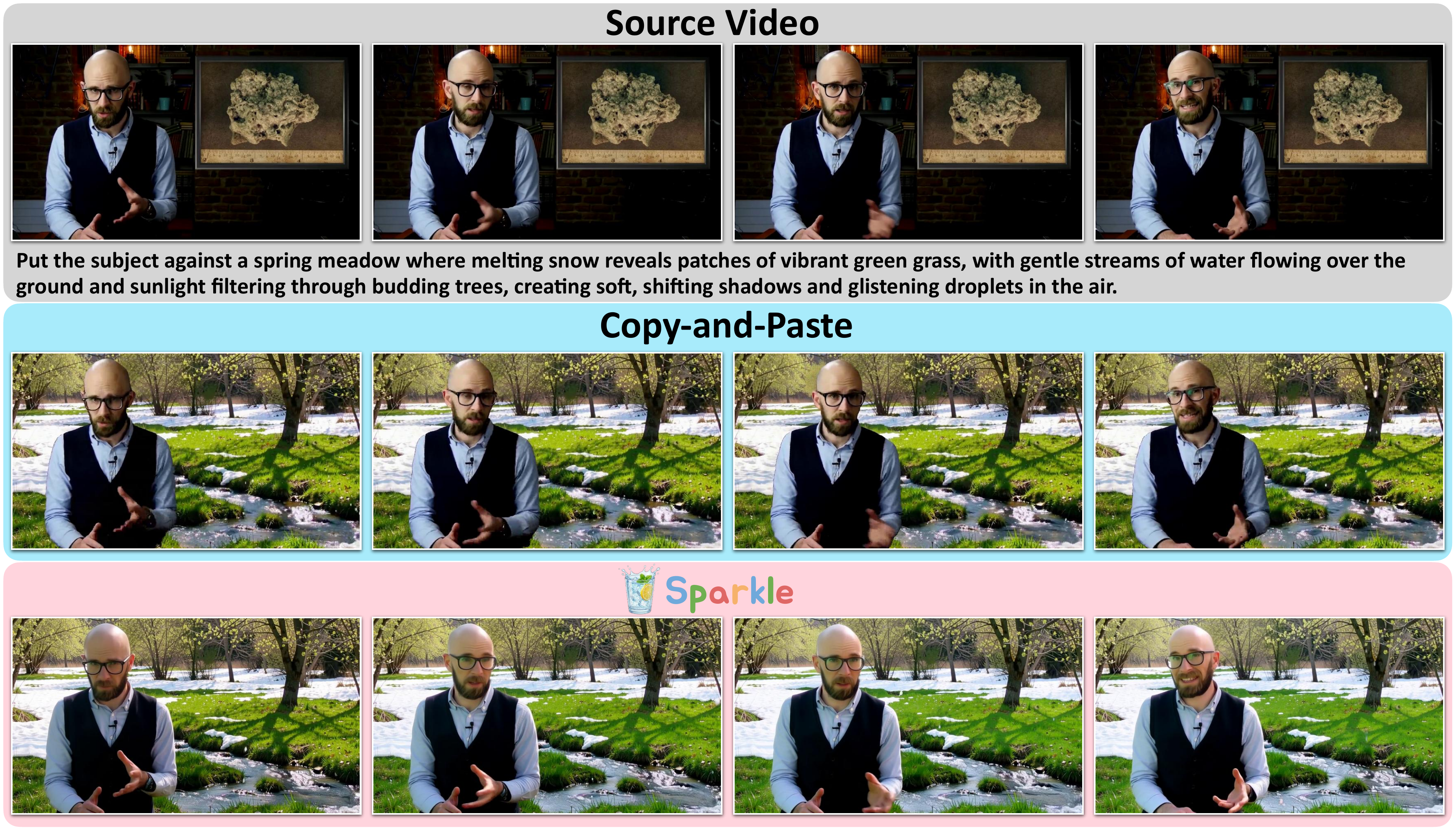}
    \caption{Data comparison between Copy-and-Paste and our proposed \emph{Sparkle}. The theme, subtheme, and scene are ``Season-spring-melting snow revealing grass''.}
    \label{fig:appendix_data_compare_copy_and_paste_season}
\end{figure}

\begin{figure}[h]
    \centering
    \includegraphics[width=1.\columnwidth]{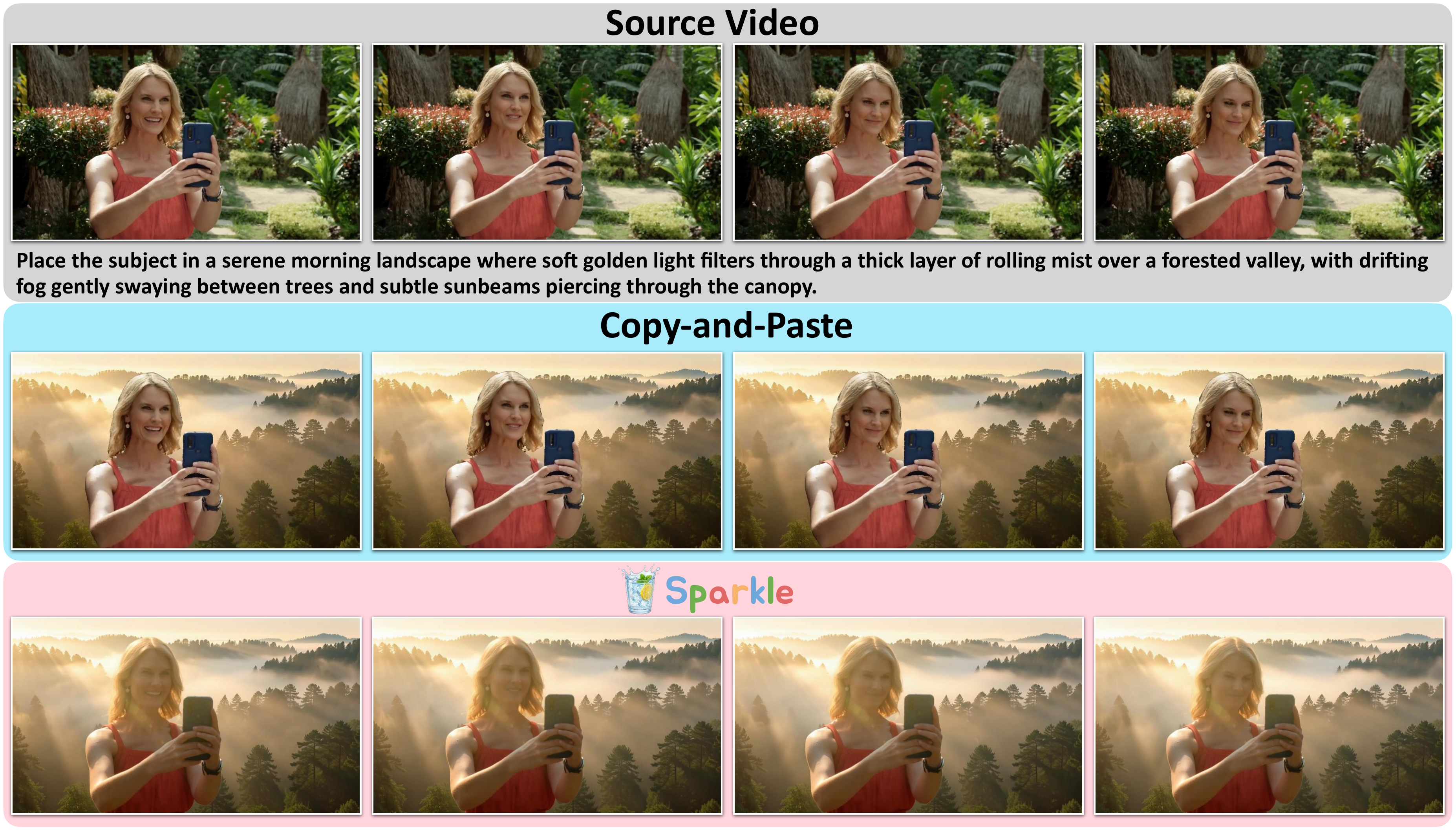}
    \caption{Data comparison between Copy-and-Paste and our proposed \emph{Sparkle}. The theme, subtheme, and scene are ``Time-dawn-morning mist rolling over terrain''.}
    \label{fig:appendix_data_compare_copy_and_paste_time}
\end{figure}

\begin{figure}[h]
    \centering
    \includegraphics[width=1.\columnwidth]{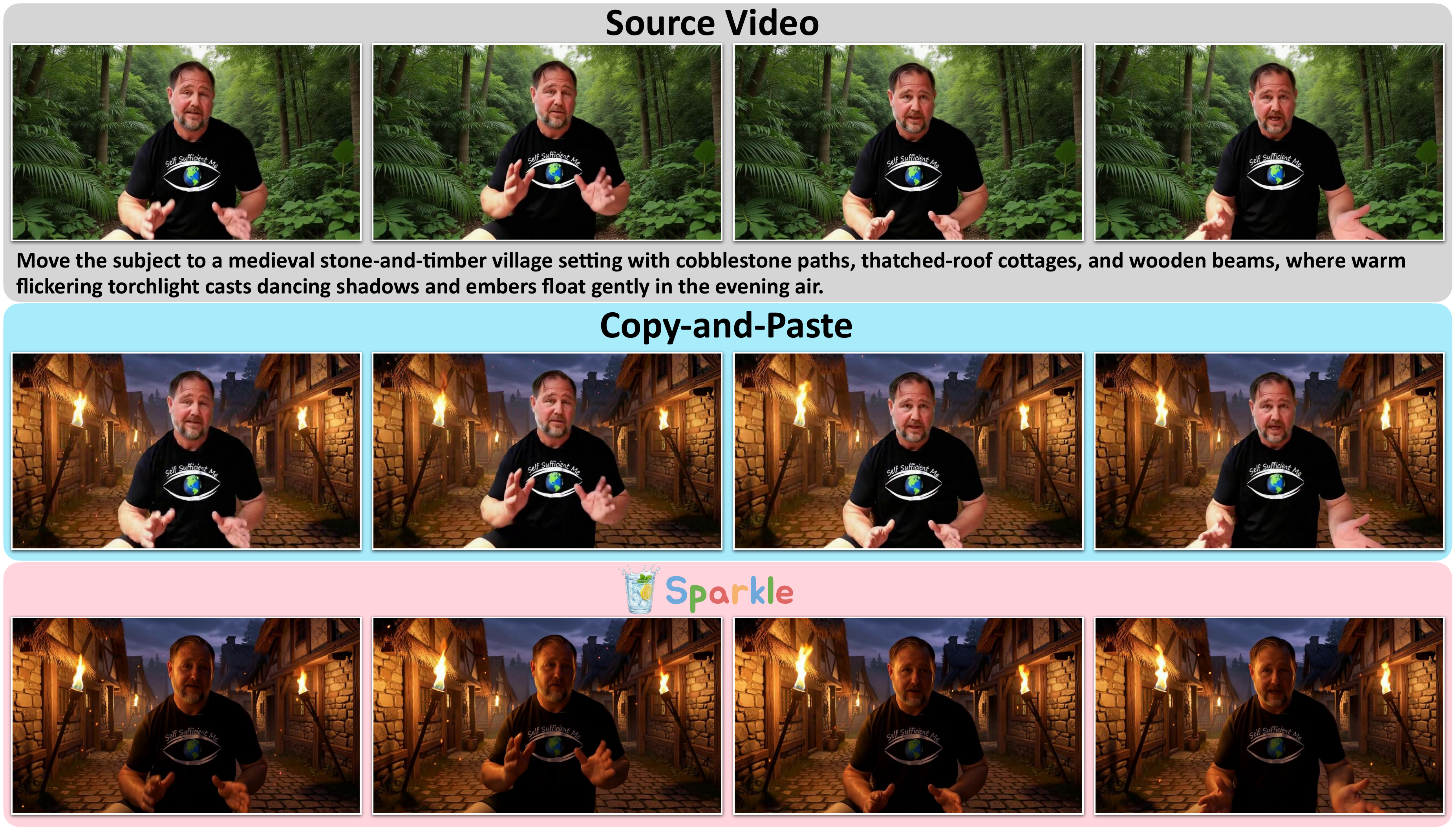}
    \caption{Data comparison between Copy-and-Paste and our proposed \emph{Sparkle}. The theme, subtheme, and scene are ``Style-era-medieval stone-and-timber village setting''.}
    \label{fig:appendix_data_compare_copy_and_paste_style}
\end{figure}

Figures~\ref{fig:appendix_data_compare_copy_and_paste_location}, \ref{fig:appendix_data_compare_copy_and_paste_season}, \ref{fig:appendix_data_compare_copy_and_paste_time}, and \ref{fig:appendix_data_compare_copy_and_paste_style} illustrate the low-quality synthesized videos by the Copy-and-Paste paradigm across the four themes, respectively. As shown in these figures, harsh contours are clearly visible, as current segmentation models are incapable of entirely eliminating contour noise. A more significant issue lies in lighting and shadow adjustments. For instance, in Figure~\ref{fig:appendix_data_compare_copy_and_paste_location}, the sunlight originates from behind the man. Therefore, maintaining the uniform lighting of the source figure creates unnatural artifacts. In contrast, the edits produced by \emph{Sparkle} not only adjust the lighting of the figure appropriately but also simulate the shadow on the table, an effect impossible to achieve with the Copy-and-Paste paradigm. Similarly, in Figure~\ref{fig:appendix_data_compare_copy_and_paste_time}, our \emph{Sparkle} edits vividly model the light reflections on the camera lens. This makes the results far more realistic than their rigidly pasted counterparts, demonstrating the high reliability of our full-video regeneration driven by decoupled guidance.

\clearpage

\subsection{Comparison with Foreground-Only Guidance} 

\begin{figure}[h]
    \centering
    \includegraphics[width=1.\columnwidth]{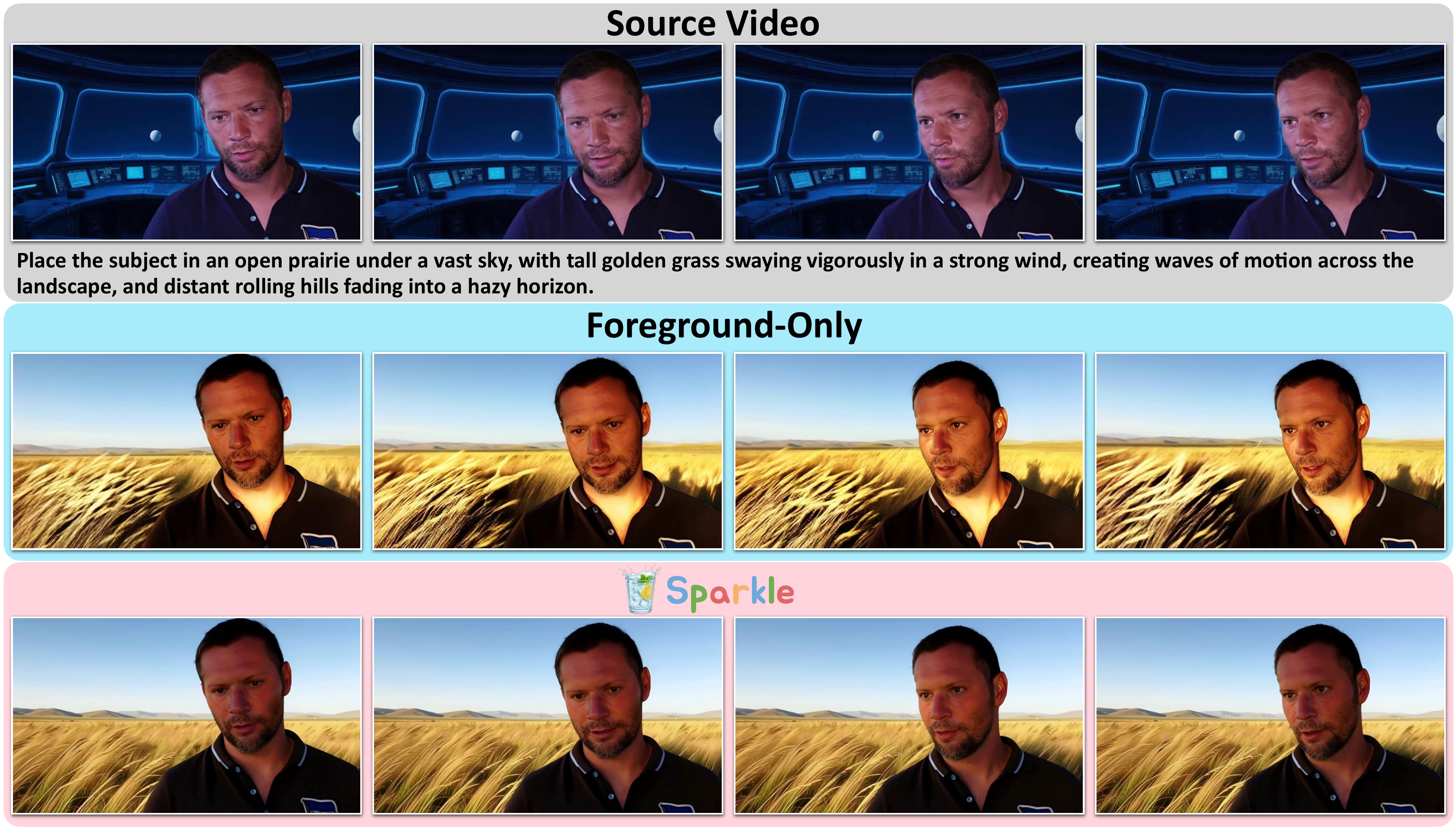}
    \caption{Data comparison between Foreground-Only and our proposed \emph{Sparkle}. The theme, subtheme, and scene are ``Location-rural-open prairie with tall grass waving''.}
    \label{fig:appendix_data_compare_foreground_only_location}
\end{figure}

\begin{figure}[h]
    \centering
    \includegraphics[width=1.\columnwidth]{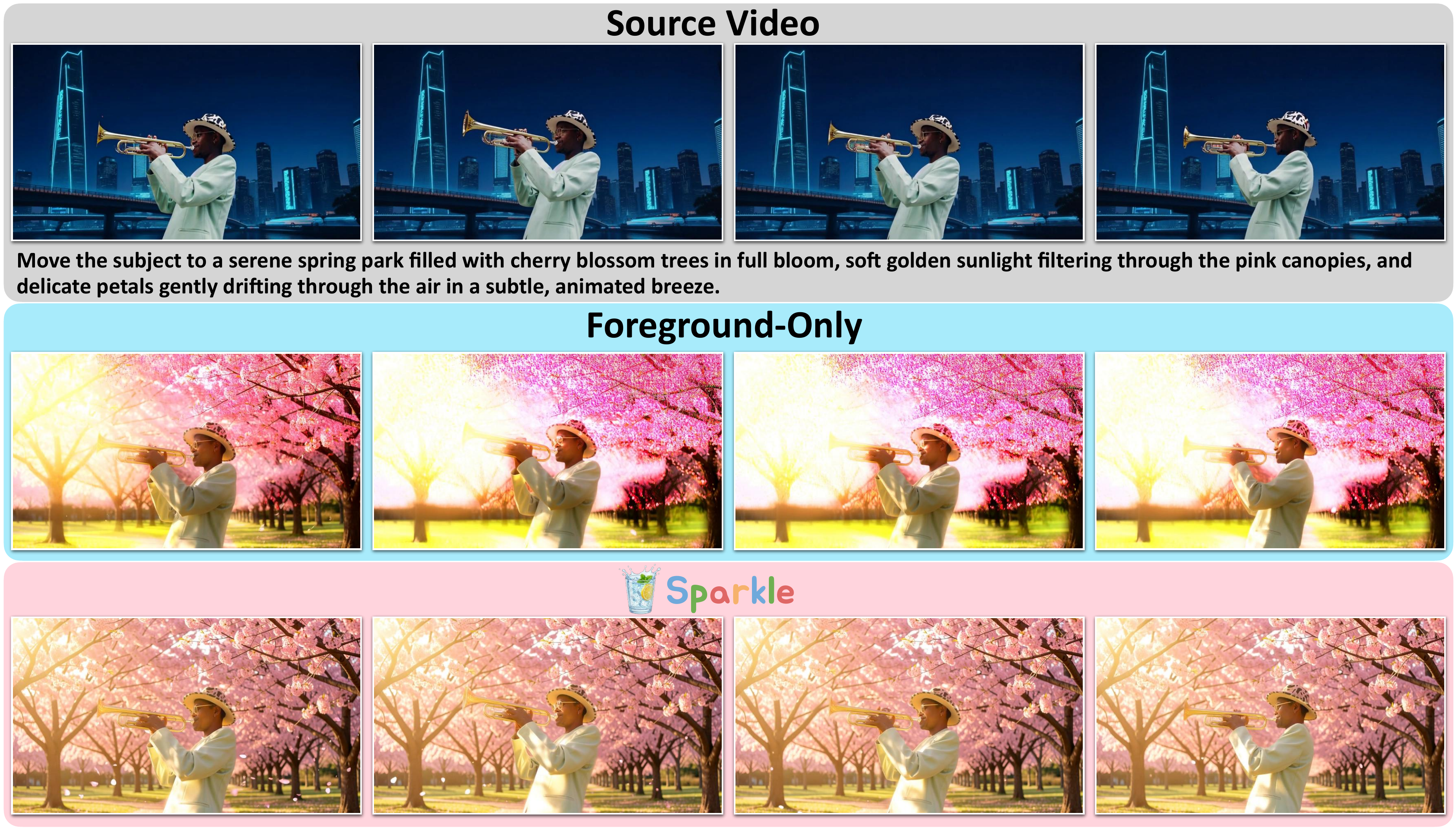}
    \caption{Data comparison between Foreground-Only and our proposed \emph{Sparkle}. The theme, subtheme, and scene are ``Season-spring-cherry blossoms in full bloom''.}
    \label{fig:appendix_data_compare_foreground_only_season}
\end{figure}

\begin{figure}[h]
    \centering
    \includegraphics[width=1.\columnwidth]{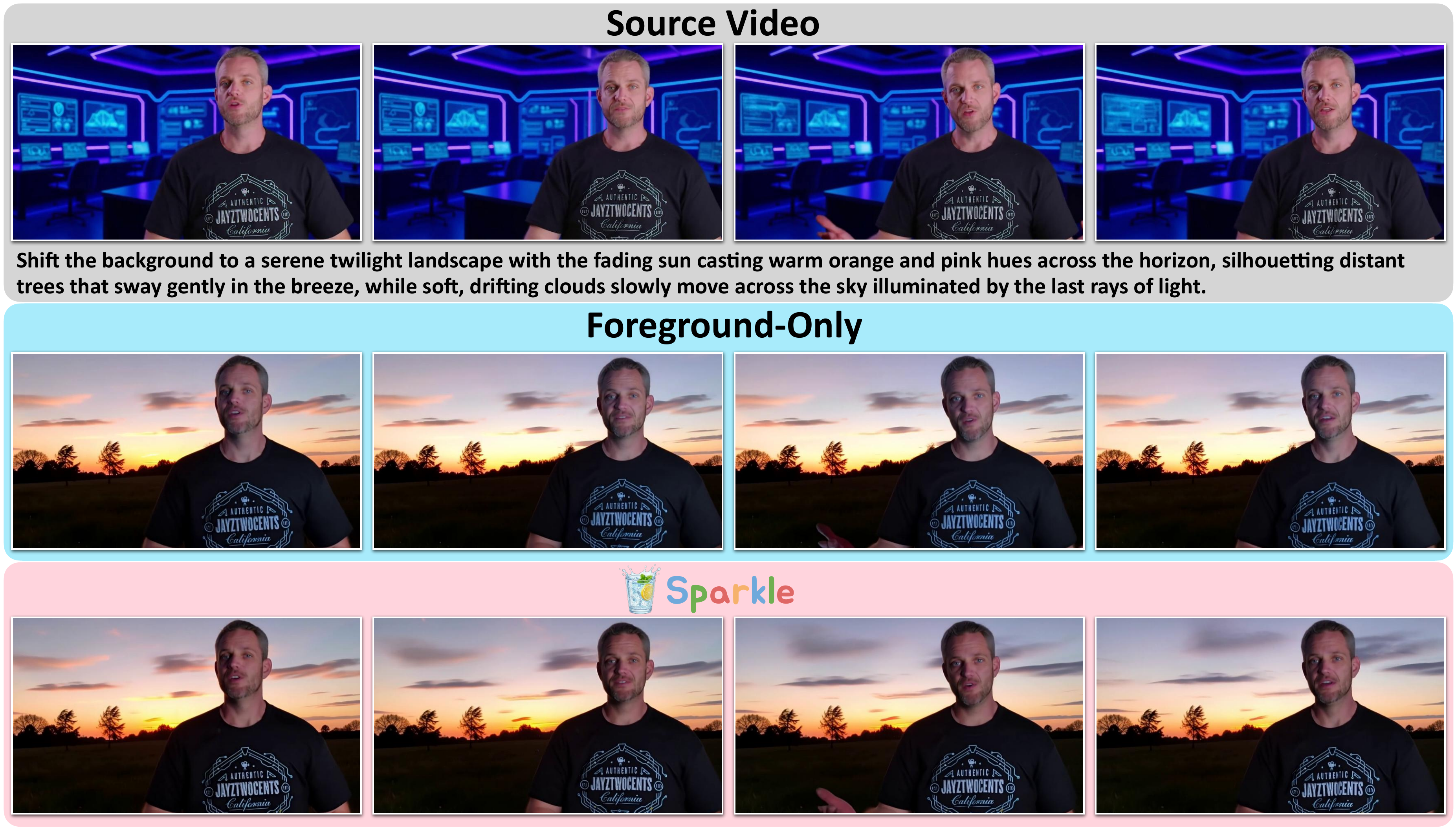}
    \caption{Data comparison between Foreground-Only and our proposed \emph{Sparkle}. The theme, subtheme, and scene are ``Time-dusk-silhouette lighting against fading sun''.}
    \label{fig:appendix_data_compare_foreground_only_time}
\end{figure}

\begin{figure}[h]
    \centering
    \includegraphics[width=1.\columnwidth]{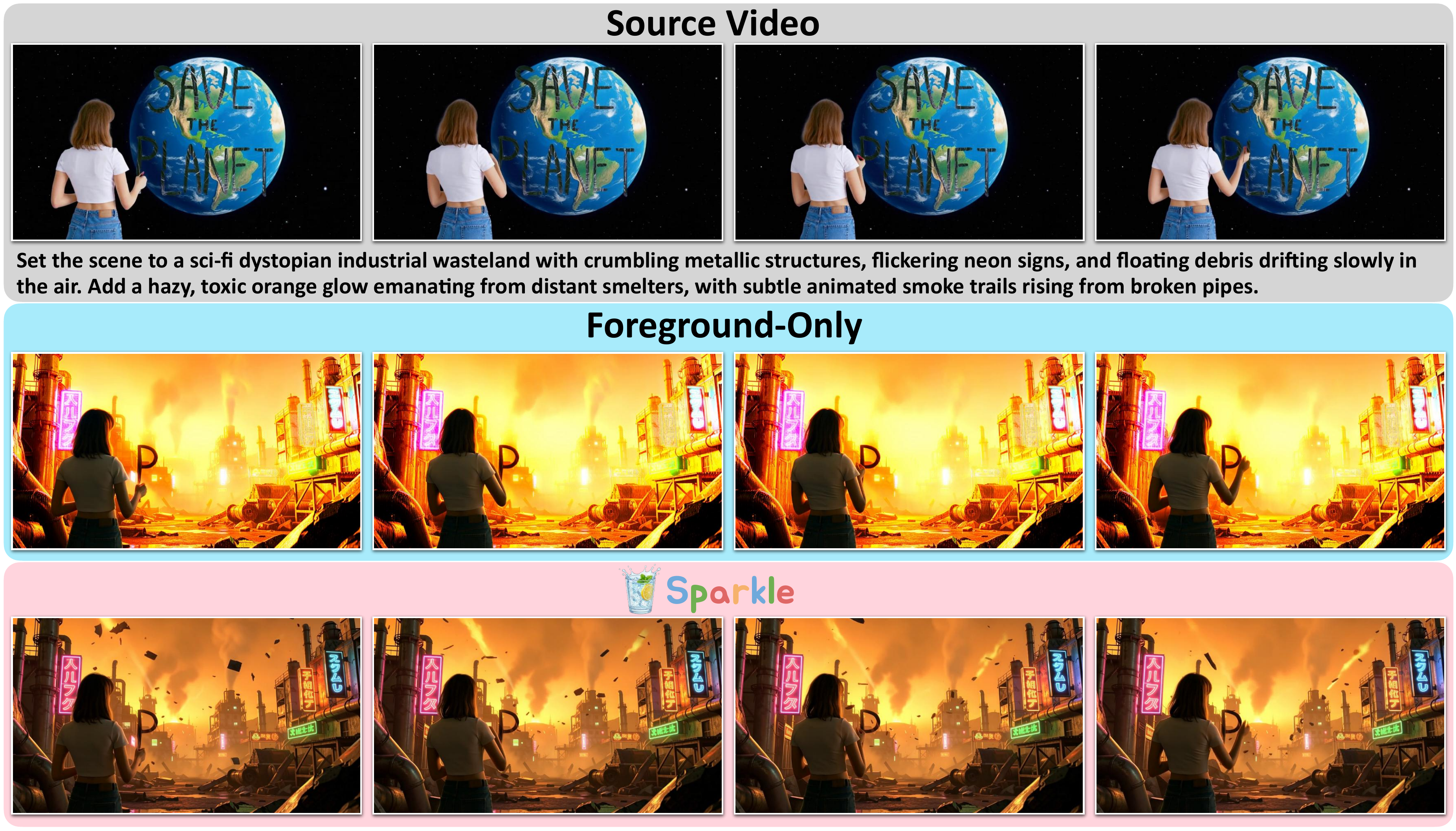}
    \caption{Data comparison between Foreground-Only and our proposed \emph{Sparkle}. The theme, subtheme, and scene are ``Style-cinematic-sci-fi dystopian industrial wasteland''.}
    \label{fig:appendix_data_compare_foreground_only_style}
\end{figure}

Because we utilize a different set of toolkits for data creation compared to OpenVE-3M, we conduct a more rigorous comparison by using only the BAIT-detected foreground to synthesize the final video in Stage 5. Under this setting, the sole variable is the presence of background guidance. The results across the four themes are presented in Figures~\ref{fig:appendix_data_compare_foreground_only_location}, \ref{fig:appendix_data_compare_foreground_only_season}, \ref{fig:appendix_data_compare_foreground_only_time}, and \ref{fig:appendix_data_compare_foreground_only_style}, respectively. Although our BAIT algorithm ensures accurate foreground preservation, the complete absence of background guidance inevitably leads to severe structural collapse. The most frequent issue, as shown in Figures~\ref{fig:appendix_data_compare_foreground_only_location} and \ref{fig:appendix_data_compare_foreground_only_season}, is the loss of high-frequency textures (such as yellow grass and blooming flowers). Furthermore, lighting control becomes highly unstable. For example, in Figure~\ref{fig:appendix_data_compare_foreground_only_style}, the frames suddenly become extremely overexposed. The model completely loses lighting control due to the difficulty of modeling motion without background guidance. Additionally, the unnatural static background issue observed in OpenVE-3M also occurs in Figure~\ref{fig:appendix_data_compare_foreground_only_time}. Conversely, with sufficient decoupled background guidance, our \emph{Sparkle}-created videos maintain excellent structural integrity. These results firmly validate that our observation is universal rather than specific to a particular toolkit, thoroughly justifying the necessity of introducing decoupled background guidance during data generation, as implemented in \emph{Sparkle}.

\clearpage

\subsection{Evaluation Results on OpenVE-Bench}

\begin{figure}[h]
    \centering
    \includegraphics[width=1.\columnwidth]{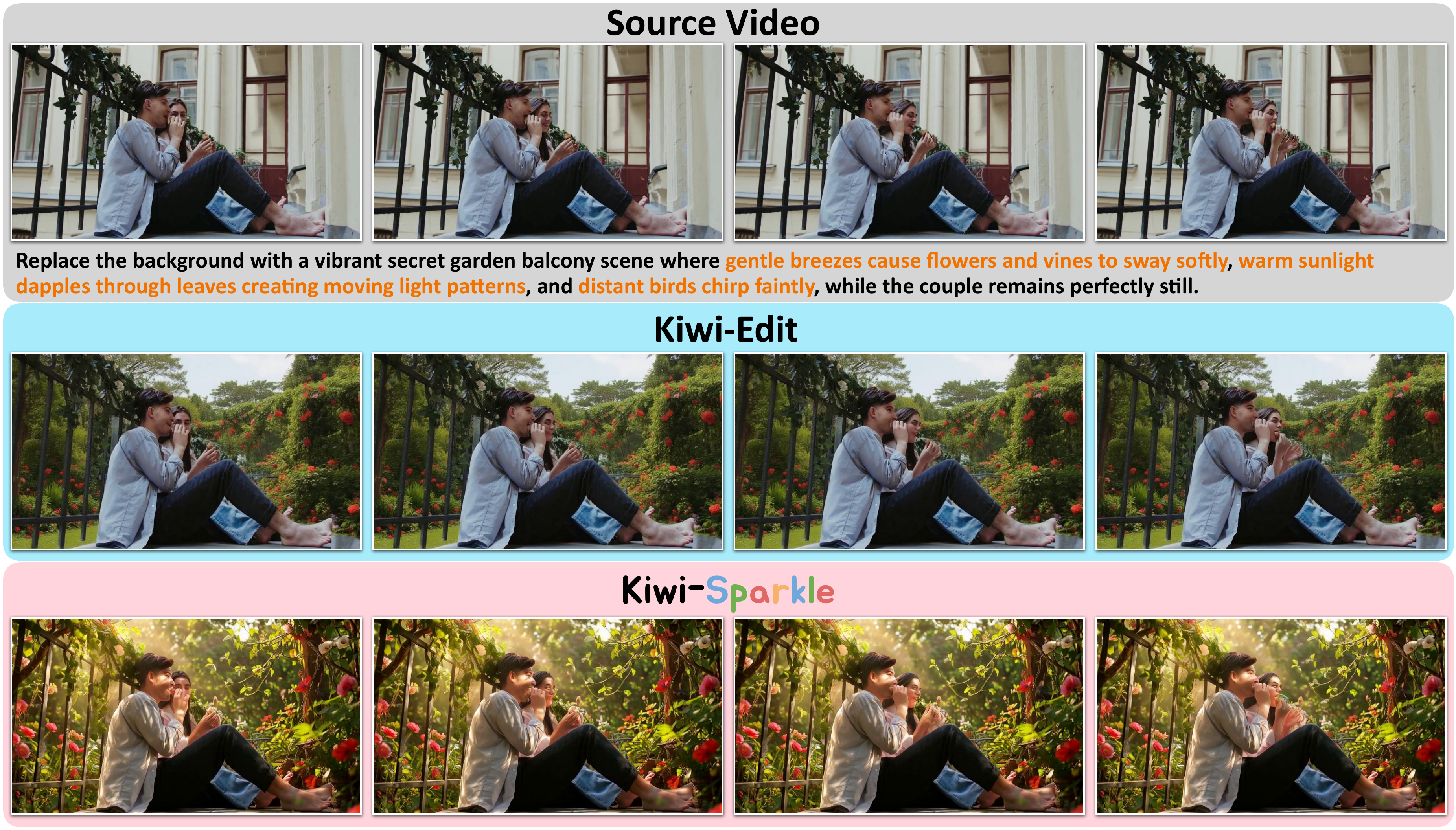}
    \caption{Edited video comparison between Kiwi-Edit and \emph{Kiwi-Sparkle} on OpenVE-Bench-Part1.}
    \label{fig:appendix_eval_results_openve_bench_wide_p1}
\end{figure}

\begin{figure}[h]
    \centering
    \includegraphics[width=1.\columnwidth]{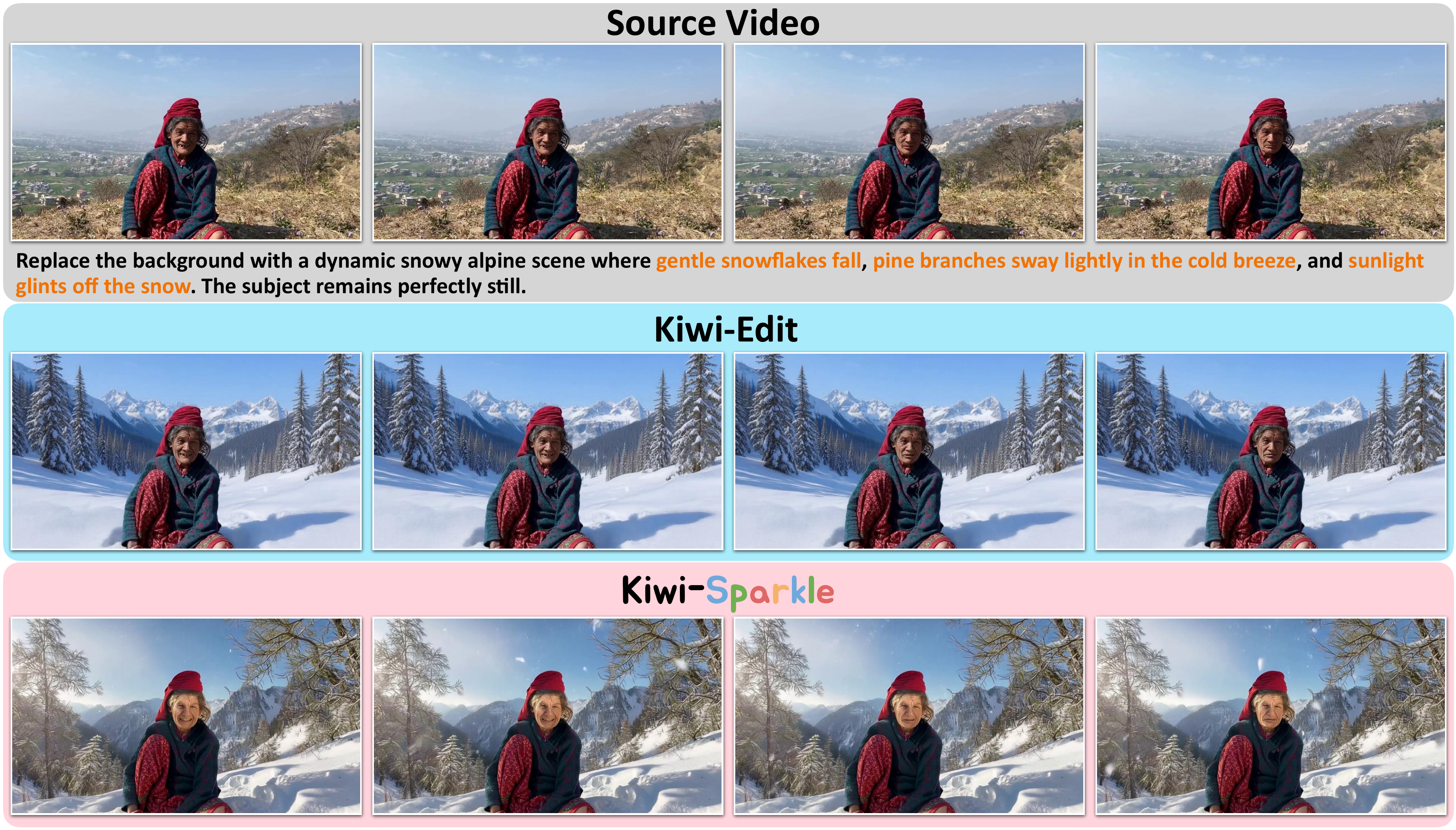}
    \caption{Edited video comparison between Kiwi-Edit and \emph{Kiwi-Sparkle} on OpenVE-Bench-Part2.}
    \label{fig:appendix_eval_results_openve_bench_wide_p2}
\end{figure}

\begin{figure}[h]
    \centering
    \includegraphics[width=1.\columnwidth]{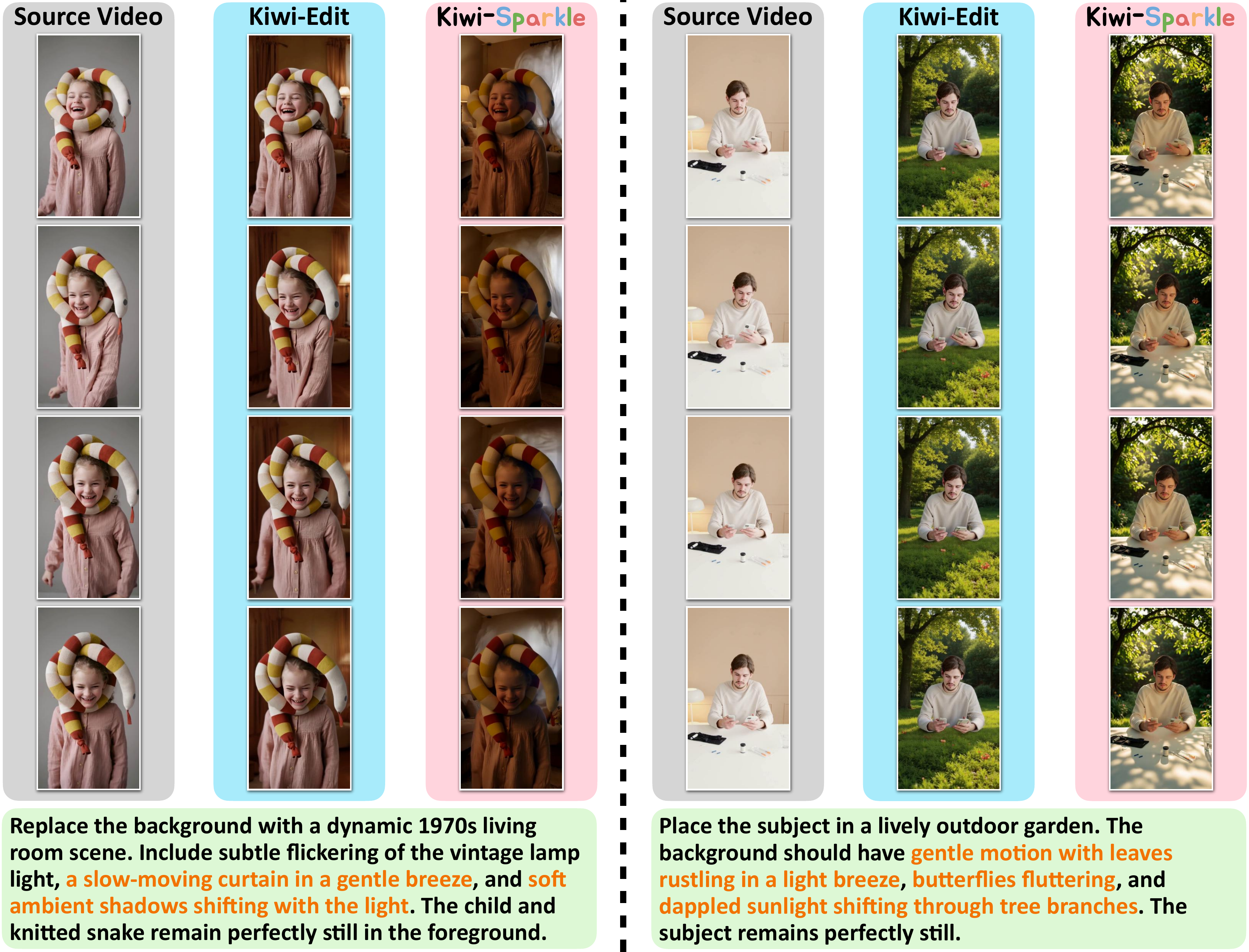}
    \caption{Edited video comparison between Kiwi-Edit and \emph{Kiwi-Sparkle} on OpenVE-Bench-Part3.}
    \label{fig:appendix_eval_results_openve_bench_portrait}
\end{figure}

Beyond evaluating the data itself, we illustrate the visual edits on OpenVE-Bench produced by the vanilla Kiwi-Edit and our \emph{Sparkle}-tuned version, \emph{Kiwi-Sparkle}, in Figures~\ref{fig:appendix_eval_results_openve_bench_wide_p1}, \ref{fig:appendix_eval_results_openve_bench_wide_p2}, and \ref{fig:appendix_eval_results_openve_bench_portrait}. We observe that Kiwi-Edit inherits the drawbacks of OpenVE-3M, consistently producing suboptimal static backgrounds. It also fails to make proper lighting adjustments, merely pasting the foreground onto the static background inharmoniously. Consequently, required dynamic elements, such as the ``warm sunlight'' in Figure~\ref{fig:appendix_eval_results_openve_bench_wide_p1} and the ``falling snowflakes'' in Figure~\ref{fig:appendix_eval_results_openve_bench_wide_p2}, are entirely missing. After fine-tuning on \emph{Sparkle}, these issues are resolved to a great extent. The edited videos become significantly more vibrant and lively, featuring harmonious lighting and motion without disturbing the foreground. This indicates that our high-quality \emph{Sparkle} dataset plays a vital role in infusing liveness into foundational background replacement capabilities following large-scale but noisy pre-training.

\clearpage

\subsection{Evaluation Results on \emph{Sparkle-Bench}} 

\begin{figure}[h]
    \centering
    \includegraphics[width=1.\columnwidth]{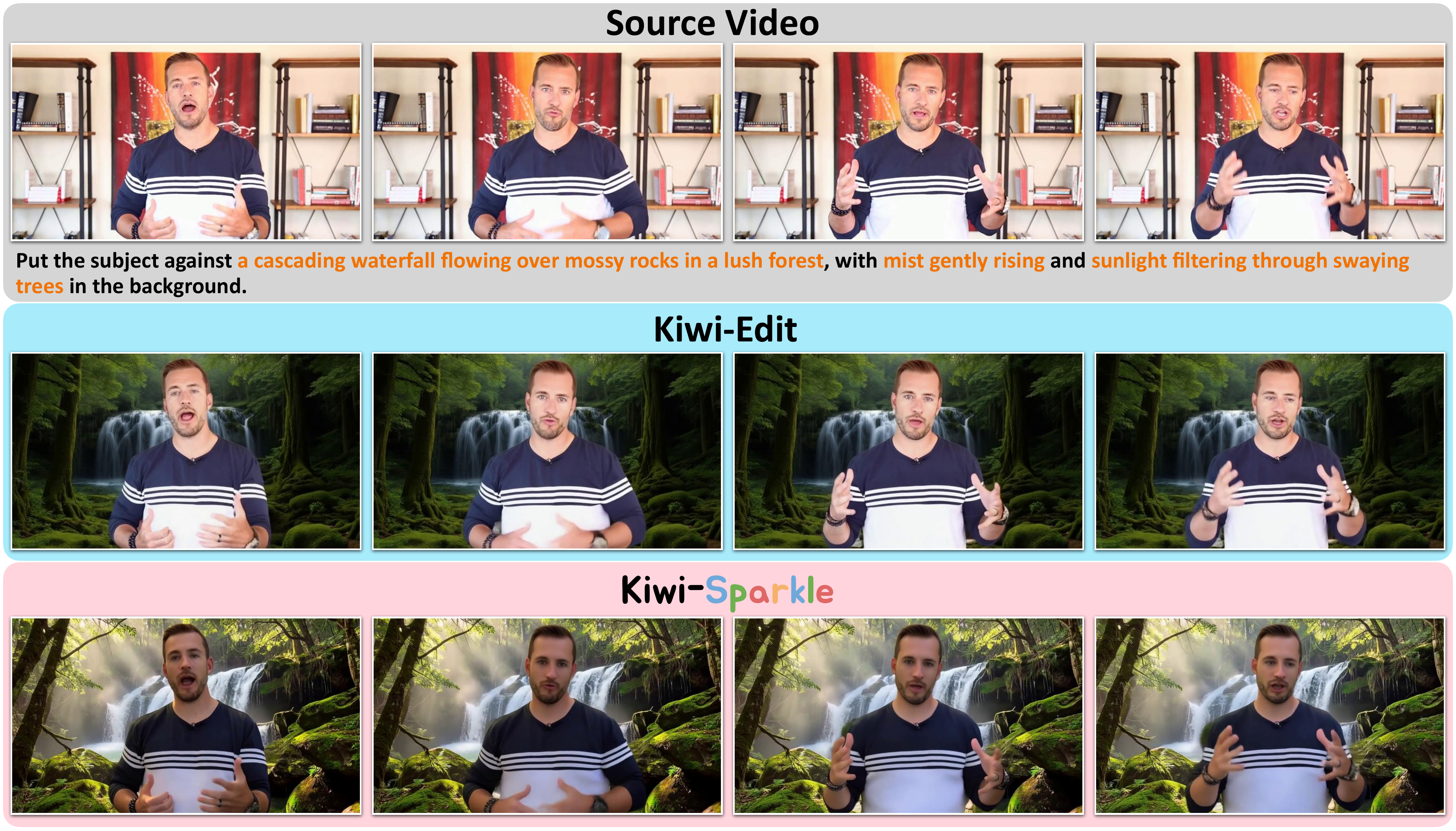}
    \caption{Edited video comparison between Kiwi-Edit and \emph{Kiwi-Sparkle} on \emph{Sparkle-Bench}. The theme, subtheme, and scene are ``Location-nature-waterfall cascading over mossy rocks''.}
    \label{fig:appendix_eval_results_sparkle_bench_location}
\end{figure}

\begin{figure}[h]
    \centering
    \includegraphics[width=1.\columnwidth]{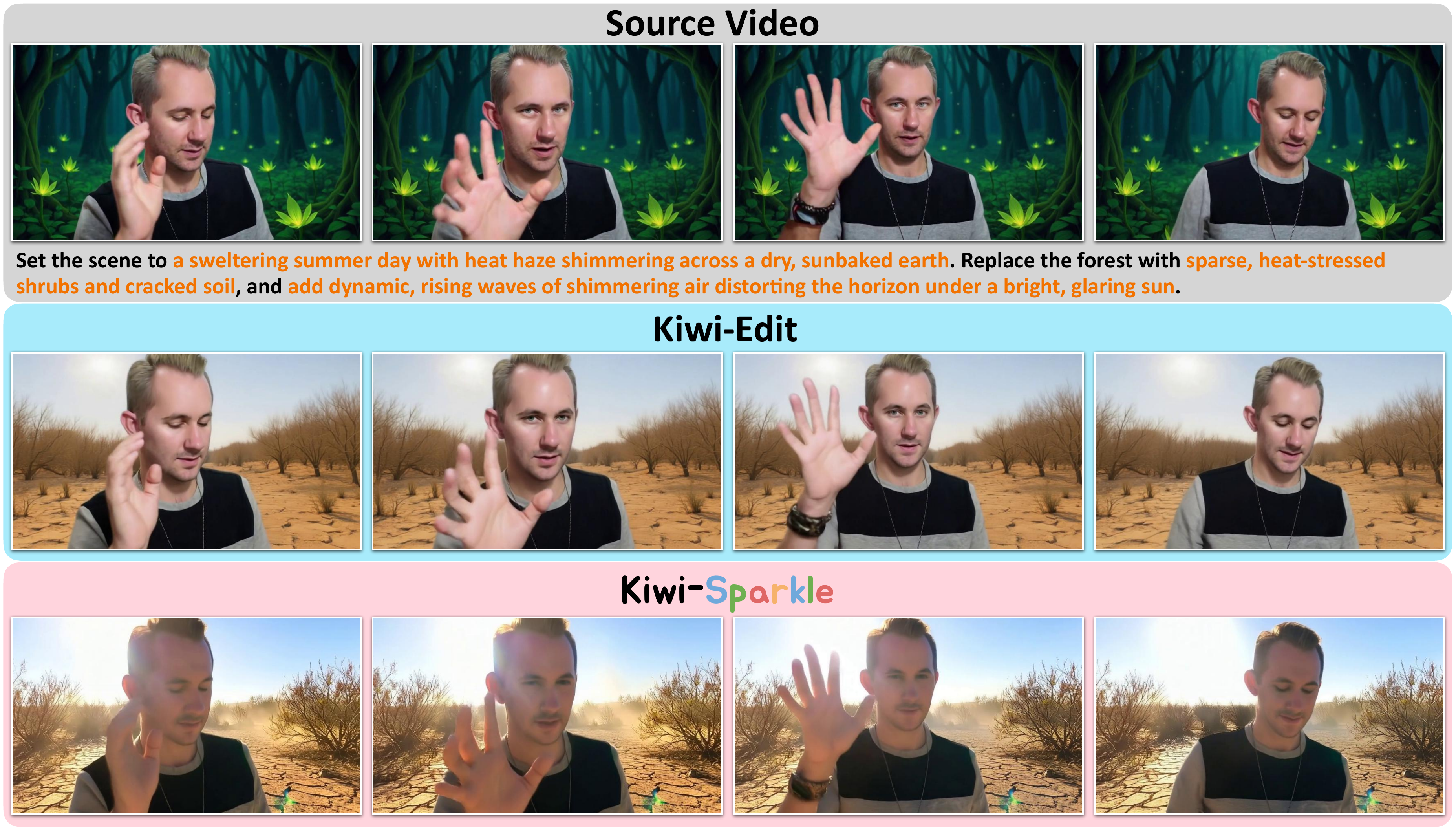}
    \caption{Edited video comparison between Kiwi-Edit and \emph{Kiwi-Sparkle} on \emph{Sparkle-Bench}. The theme, subtheme, and scene are ``Season-summer-heat haze shimmering on ground''.}
    \label{fig:appendix_eval_results_sparkle_bench_season}
\end{figure}

\begin{figure}[h]
    \centering
    \includegraphics[width=1.\columnwidth]{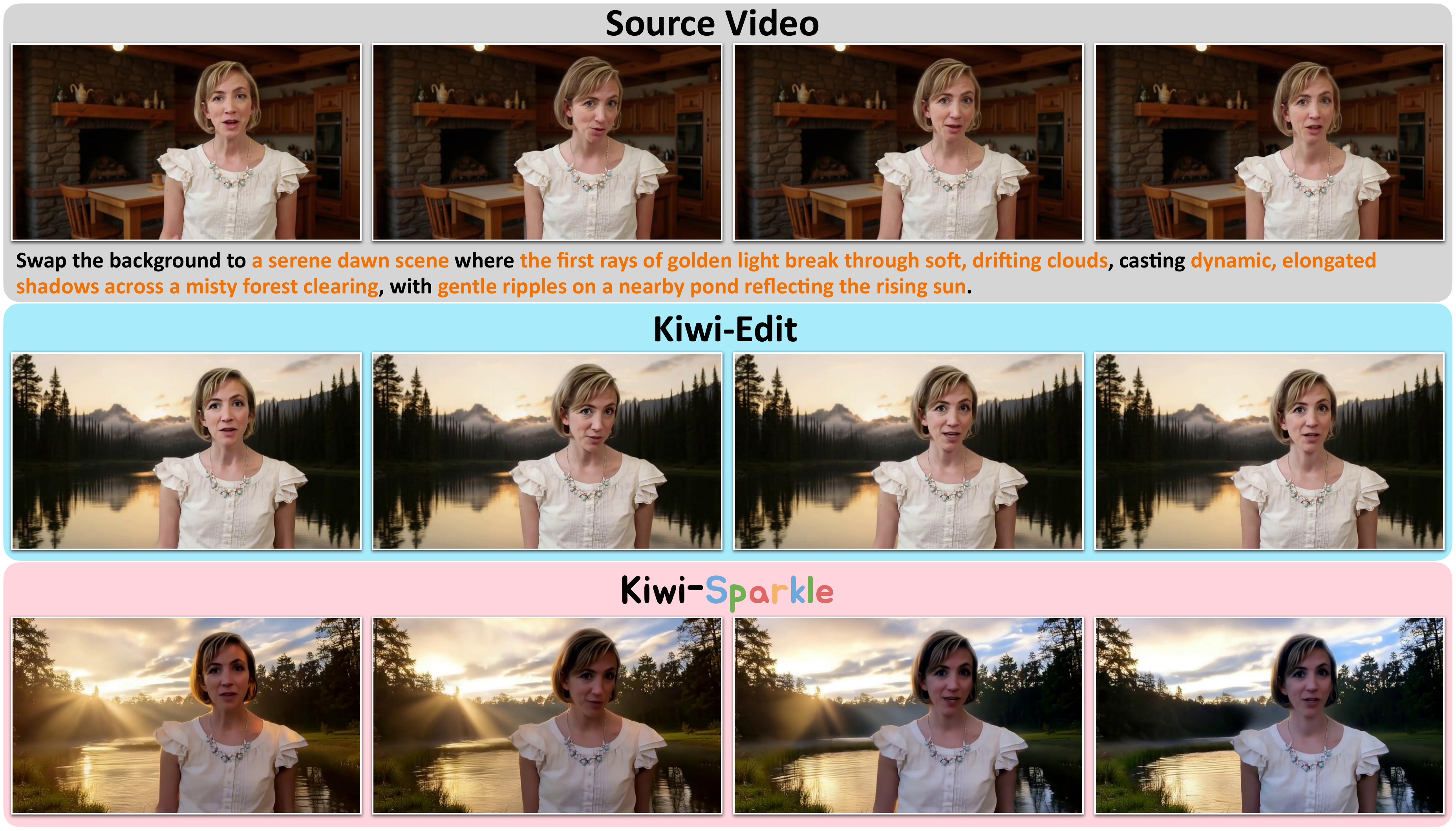}
    \caption{Edited video comparison between Kiwi-Edit and \emph{Kiwi-Sparkle} on \emph{Sparkle-Bench}. The theme, subtheme, and scene are ``Time-dawn-first rays of light breaking through''.}
    \label{fig:appendix_eval_results_sparkle_bench_time}
\end{figure}

\begin{figure}[h]
    \centering
    \includegraphics[width=1.\columnwidth]{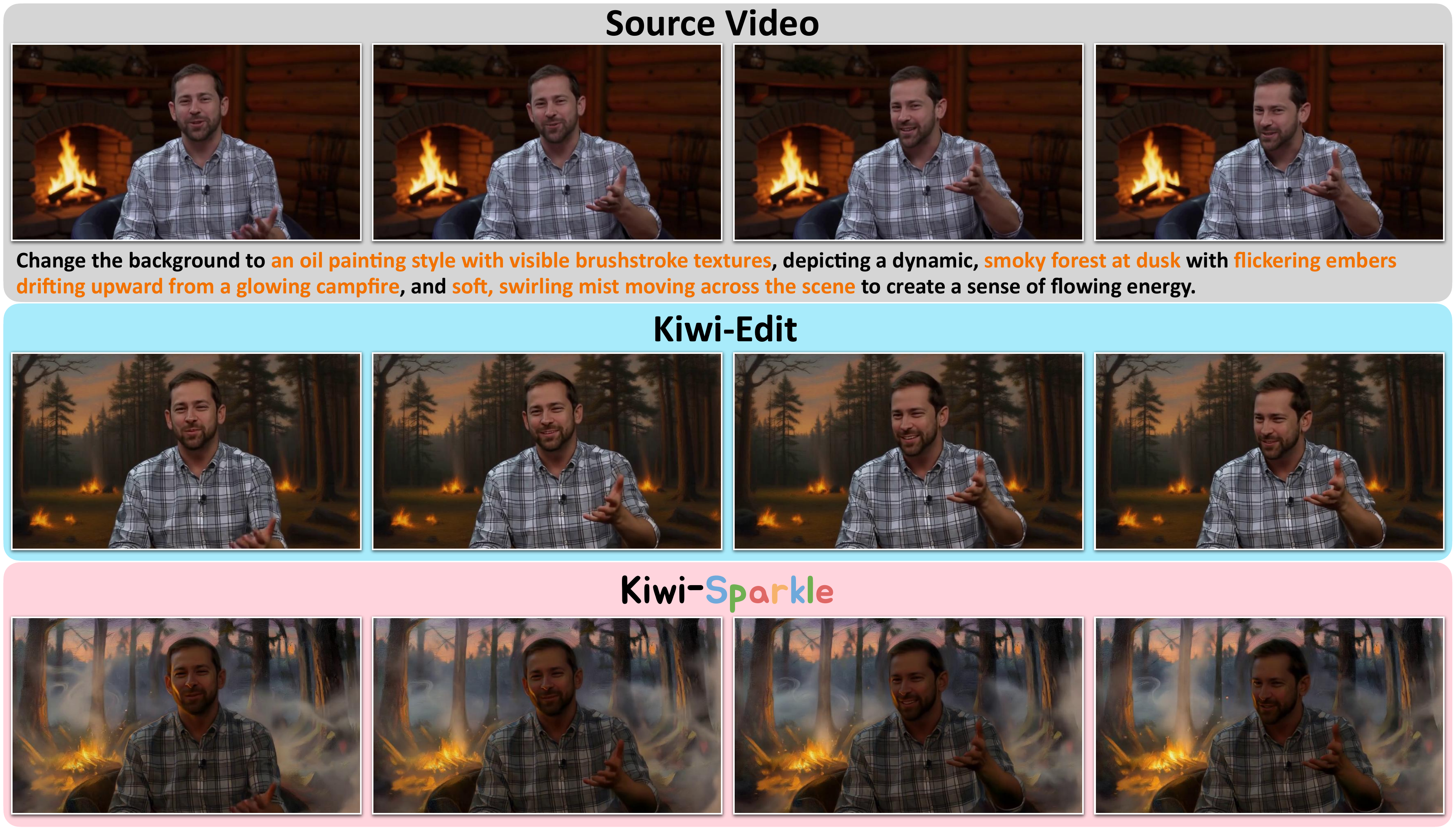}
    \caption{Edited video comparison between Kiwi-Edit and \emph{Kiwi-Sparkle} on \emph{Sparkle-Bench}. The theme, subtheme, and scene are ``Style-art style-oil painting style with visible brushstroke textures''.}
    \label{fig:appendix_eval_results_sparkle_bench_style}
\end{figure}

Figures~\ref{fig:appendix_eval_results_sparkle_bench_location}, \ref{fig:appendix_eval_results_sparkle_bench_season}, \ref{fig:appendix_eval_results_sparkle_bench_time}, and \ref{fig:appendix_eval_results_sparkle_bench_style} illustrate the evaluation results across the four themes on \emph{Sparkle-Bench} for Kiwi-Edit and \emph{Kiwi-Sparkle}. Similar to the results on OpenVE-Bench, Kiwi-Edit consistently produces suboptimal static or light-inconsistent edits, demonstrating that its low \emph{Background Dynamics} (BgDy) and \emph{Background Visual Quality} (BgVi) scores under our proposed evaluation metrics are well-justified. In contrast, our \emph{Kiwi-Sparkle} yields significantly higher-quality results, accurately modeling subtle motions such as ``heat haze'' in Figure~\ref{fig:appendix_eval_results_sparkle_bench_season} and ``gentle ripples'' in Figure~\ref{fig:appendix_eval_results_sparkle_bench_time}. These successful edits prove that the knowledge embedded within \emph{Sparkle} is highly suitable for general models to absorb, even across a broad range of scenes beyond the OpenVE-3M distribution.

\clearpage

\subsection{\emph{Kiwi-Sparkle} as an Effective Foreground Tracker}

\begin{figure}[h]
    \centering
    \includegraphics[width=1.\columnwidth]{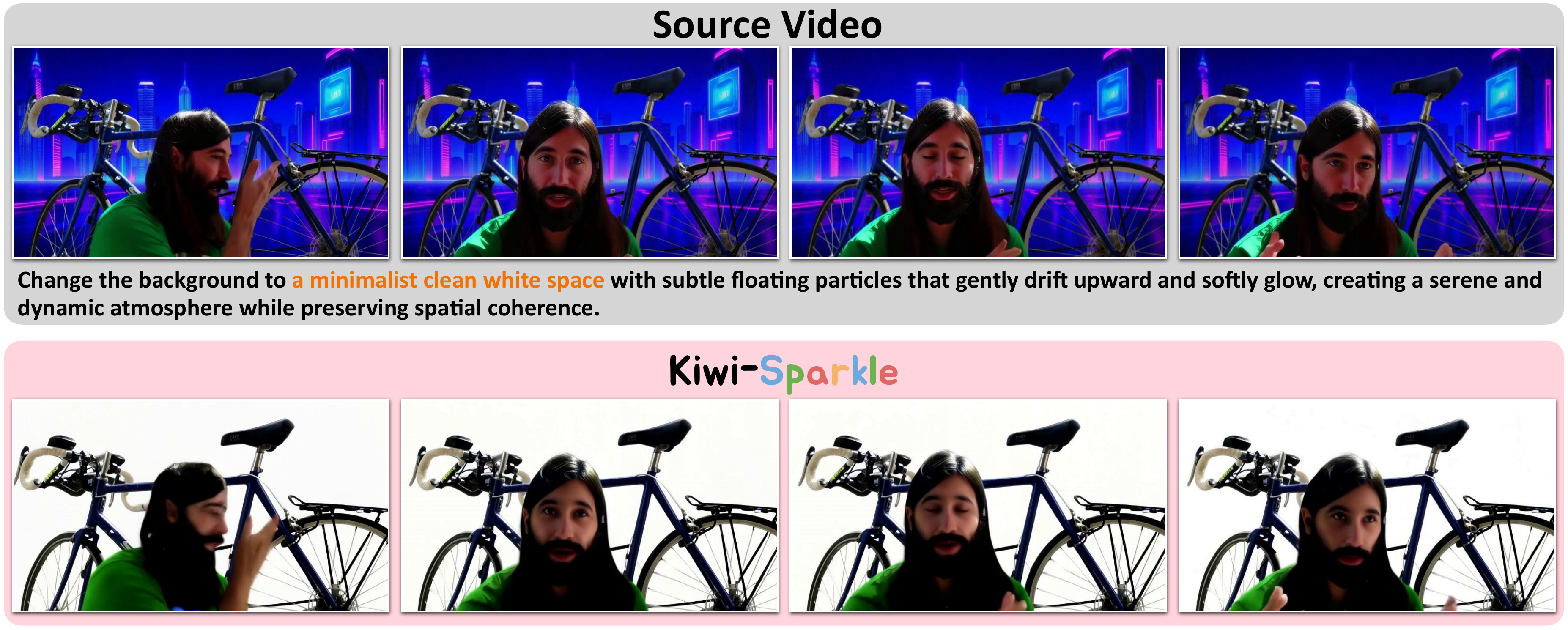}
    \caption{\emph{Kiwi-Sparkle} as an effective foreground tracker by using the trigger phrase ``a minimalist clean white space''-Part1.}
    \label{fig:appendix_kiwi_sparkle_as_foreground_tracker_p1}
\end{figure}

\begin{figure}[h]
    \centering
    \includegraphics[width=1.\columnwidth]{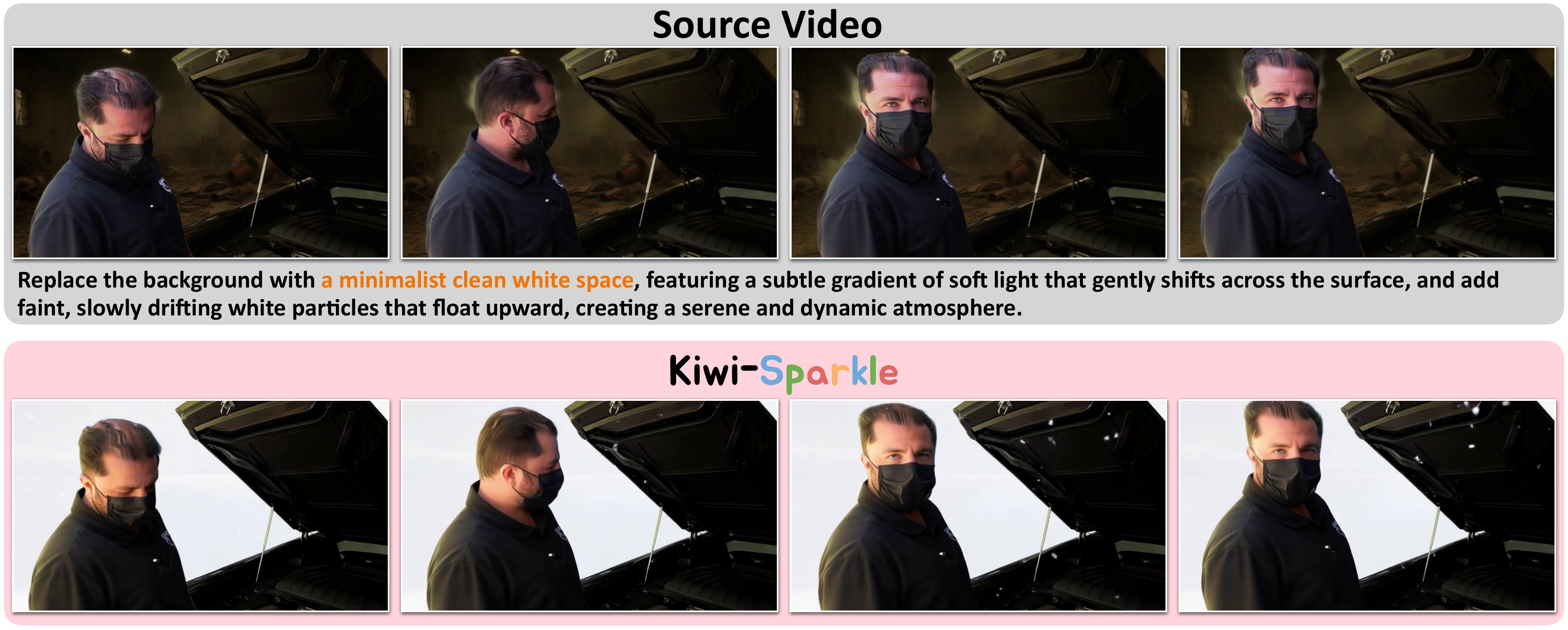}
    \caption{\emph{Kiwi-Sparkle} as an effective foreground tracker by using the trigger phrase ``a minimalist clean white space''-Part2.}
    \label{fig:appendix_kiwi_sparkle_as_foreground_tracker_p2}
\end{figure}

\begin{figure}[h]
    \centering
    \includegraphics[width=1.\columnwidth]{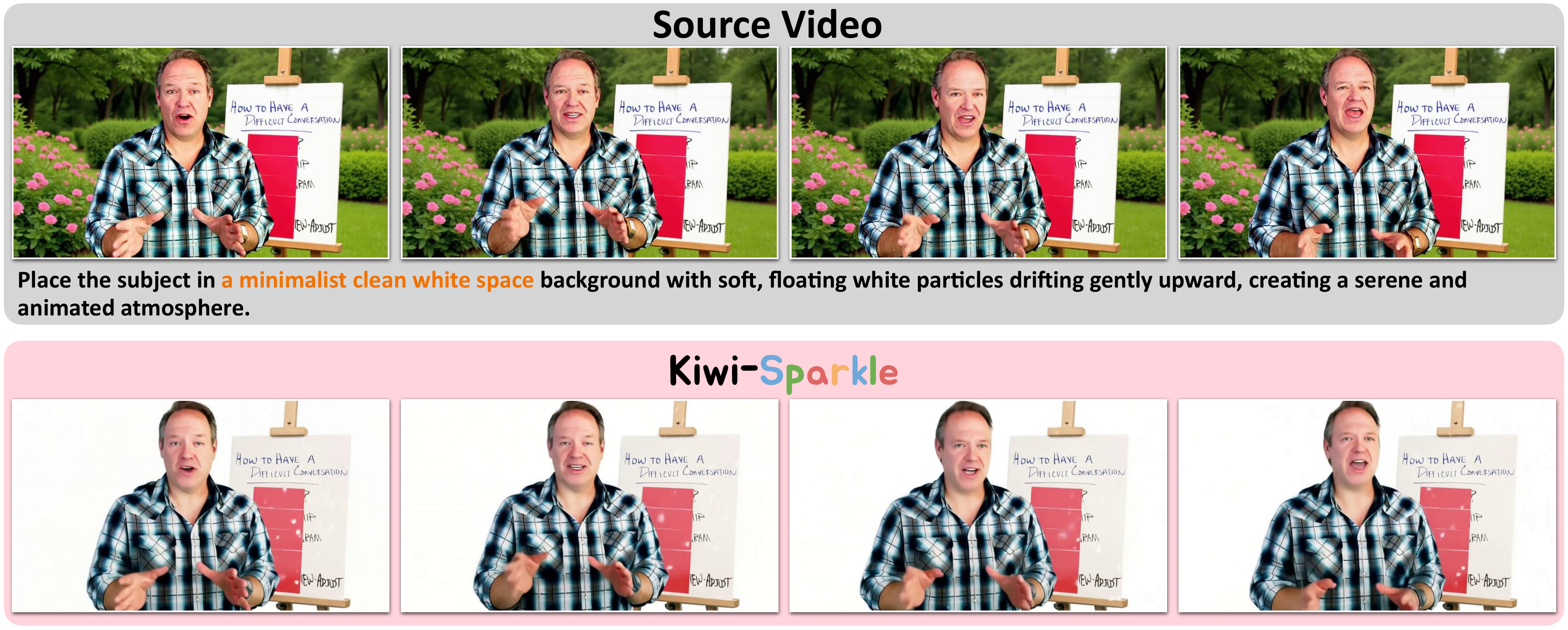}
    \caption{\emph{Kiwi-Sparkle} as an effective foreground tracker by using the trigger phrase ``a minimalist clean white space''-Part3.}
    \label{fig:appendix_kiwi_sparkle_as_foreground_tracker_p3}
\end{figure}

\begin{figure}[h]
    \centering
    \includegraphics[width=1.\columnwidth]{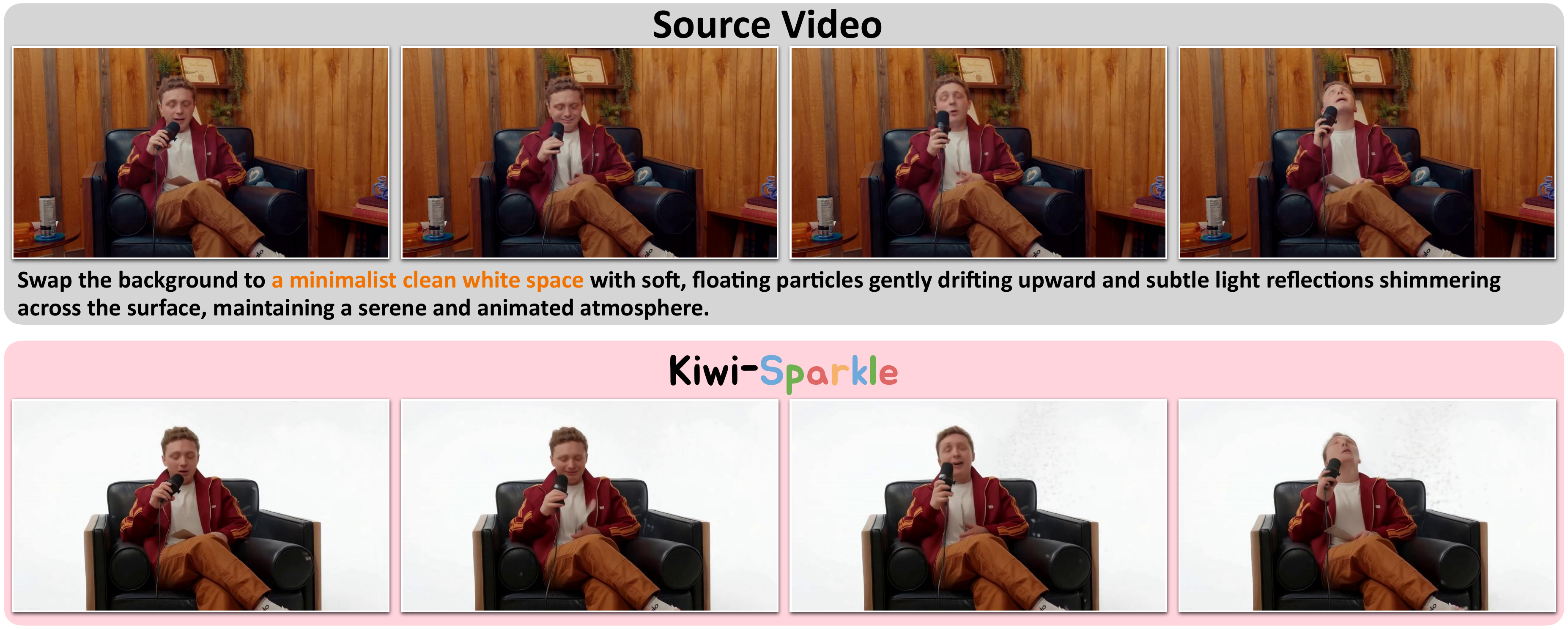}
    \caption{\emph{Kiwi-Sparkle} as an effective foreground tracker by using the trigger phrase ``a minimalist clean white space''-Part4.}
    \label{fig:appendix_kiwi_sparkle_as_foreground_tracker_p4}
\end{figure}

Beyond visual comparisons of the data and models, we demonstrate that \emph{Kiwi-Sparkle} possesses strong foreground tracking capabilities inherited from the proposed BAIT algorithm, alongside robust instruction-following skills. We validate this by introducing a specific scene description, ``a minimalist clean white space,'' as an editing category within the \emph{Style} theme. By applying this trigger phrase, \emph{Kiwi-Sparkle} accurately isolates foreground subjects from their original scenes onto a new white background. As illustrated in Figures~\ref{fig:appendix_kiwi_sparkle_as_foreground_tracker_p1}, \ref{fig:appendix_kiwi_sparkle_as_foreground_tracker_p2}, \ref{fig:appendix_kiwi_sparkle_as_foreground_tracker_p3}, and \ref{fig:appendix_kiwi_sparkle_as_foreground_tracker_p4}, even complex or large-scale foregrounds, such as the bicycle (Figure~\ref{fig:appendix_kiwi_sparkle_as_foreground_tracker_p1}) and the car (Figure~\ref{fig:appendix_kiwi_sparkle_as_foreground_tracker_p2}), can be seamlessly detached by \emph{Kiwi-Sparkle}. This compelling application not only solidifies our BAIT contribution but also sheds light on a potential editing-oriented object segmentation paradigm, a promising direction we leave for future research.

\clearpage

\subsection{Common Failure Modes of \emph{Kiwi-Sparkle}}
\vspace{-0.5em}

\begin{figure}[h]
    \centering
    \includegraphics[width=.9\columnwidth]{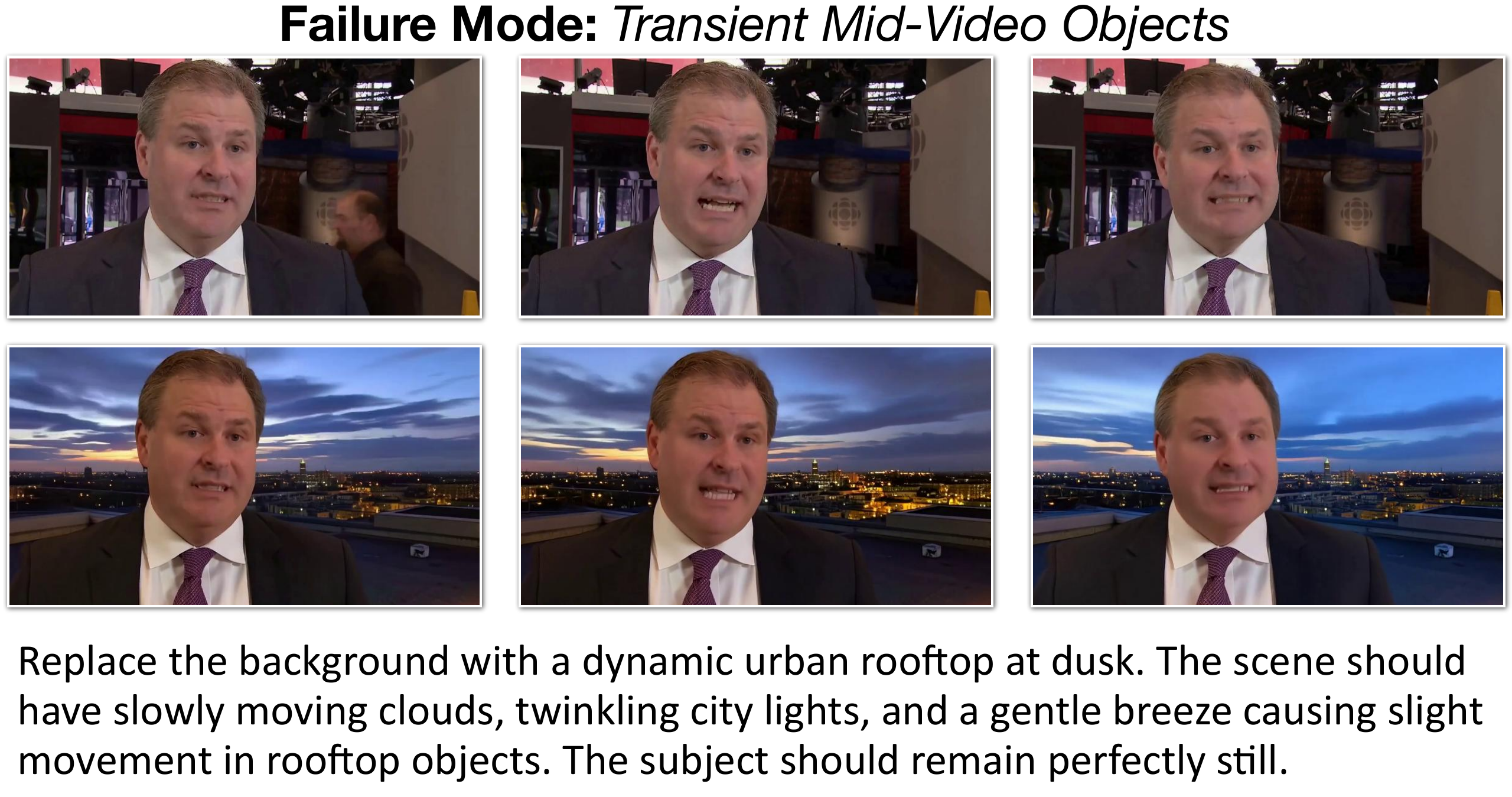}
    \caption{Sample illustration of \emph{Kiwi-Sparkle}'s common failure mode: \emph{Transient Mid-Video Objects}.}
    \label{fig:appendix_failure_p1}
\vspace{-1em}
\end{figure}

\begin{figure}[h]
    \centering
    \includegraphics[width=.9\columnwidth]{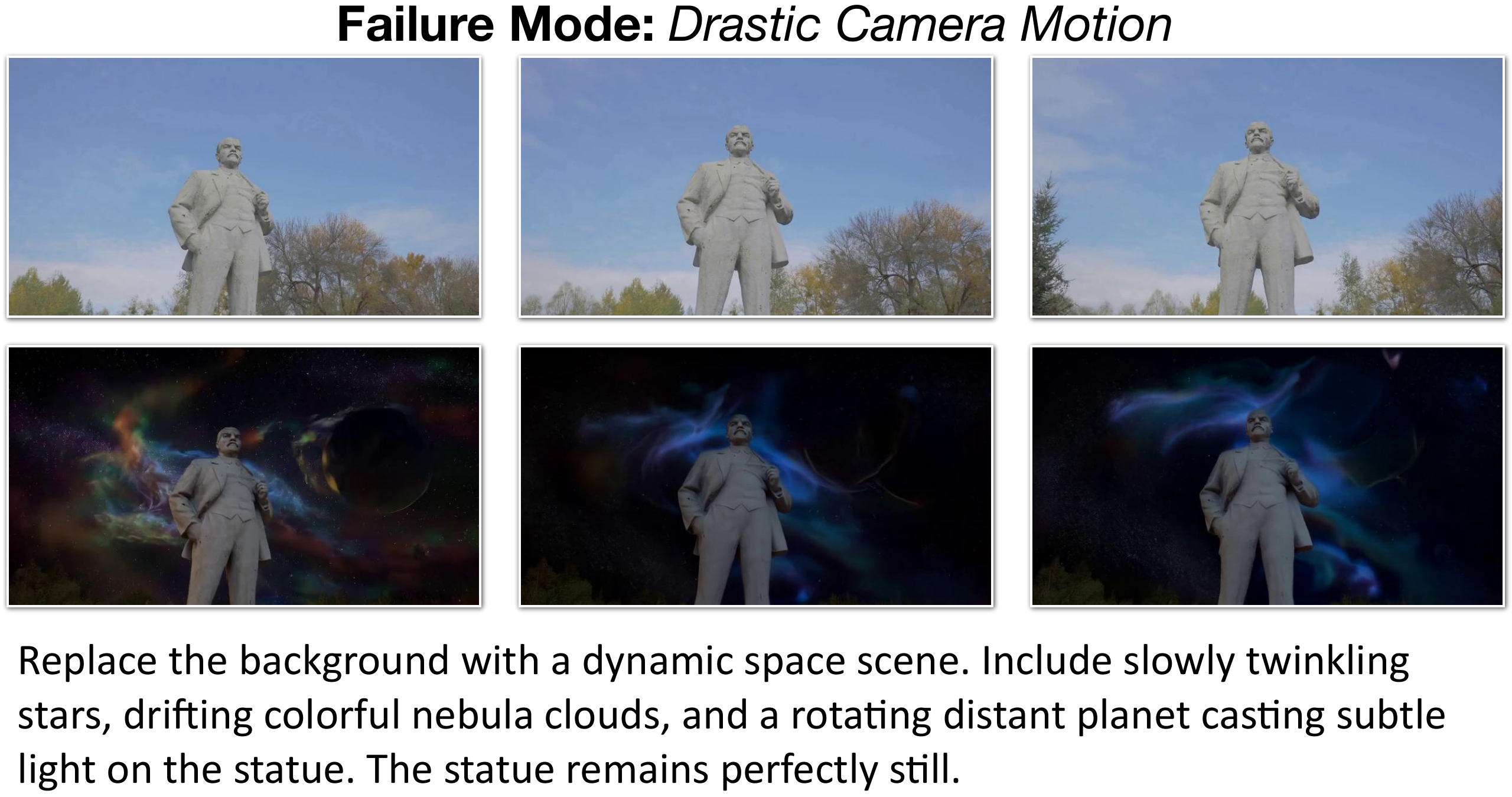}
    \caption{Sample illustration of \emph{Kiwi-Sparkle}'s common failure mode: \emph{Drastic Camera Motion}.}
    \label{fig:appendix_failure_p2}
\end{figure}

To characterize \emph{Kiwi-Sparkle}'s capability boundaries, we identify three representative failure modes:
\begin{itemize}[leftmargin=1em,itemsep=0.2ex,topsep=0pt]
\item \emph{Transient foreground objects appearing briefly mid-video.} As shown in Figure~\ref{fig:appendix_failure_p1}, the man who rapidly walks leftward and disappears midway through the source video is missing from the edited video. Such cases are inherently difficult for sparse frame sampling. Without targeted data curation, \emph{Kiwi-Sparkle} may fail to handle objects that rapidly appear and disappear between sampled frames. Salient-frame sampling could help address this issue, which we leave for future work to explore.
\item \emph{Drastic camera motion.} Due to Sparkle’s fixed-camera distribution, background replacement under drastic camera motion, \emph{e.g.,} a $60^\circ$ rotation, remains challenging. As shown in Figure~\ref{fig:appendix_failure_p2}, the rotating viewpoint leads to incomplete background replacement, leaving trees behind the statue visible in the edited video. We consider this an important direction for future work. \item \emph{Implicit background replacement requiring contextual reasoning.} Such instructions do not fully specify the target background. For example, ``replace the background with a forest in Singapore'' requires the model to infer region-specific visual characteristics before editing. \emph{Kiwi-Sparkle} performs less reliably on such cases due to limited implicit-reasoning data. As this is a common issue across existing video editors, we omit visualizations for simplicity, and view reasoning-intensive background replacement as an important future direction, with Sparkle providing a foundation for developing this capability.
\end{itemize}

\section{Limitations and Broader Impact}
\label{subsec:appendix_limitation_and_broader_imapct}

As discussed in Section~\ref{subsec:source_video_collection}, it remains infeasible for current open-source models to synchronize camera movement before and after background replacement. Therefore, we restrict the source video scope to static-camera sequences. Consequently, our approach may exhibit unstable performance when applied to videos featuring drastic camera movements. However, when the base model already possesses camera-control capability, as in Kiwi-Edit, this ability is preserved to some extent after Sparkle training under moderate camera motion. This is supported by the substantial gain on OpenVE-Bench, which includes both fixed-camera videos and videos with moderate camera motion. The improvement indicates that Sparkle adds complementary dynamic background-replacement capability without severely impairing existing camera-control ability. Although Sparkle alone remains limited under drastic camera motion due to the large distribution gap, a joint training with camera-control data could possibly activate both capabilities. We leave this exploration to future work.

Additionally, because our initial frame editor relies on FLUX.2-klein-9B~\cite{flux2klein}, the diversity and knowledge scope are inherently bounded by its training data, which could potentially introduce biases. This limitation similarly applies to the foundation model of \emph{Kiwi-Sparkle}, \emph{i.e.,} Kiwi-Edit~\cite{kiwiedit}. Being trained on massive corpora including GPT-Image-Edit-1.5M~\cite{gptimageedit}, NHR-Edit~\cite{nhredit}, Ditto-1M~\cite{ditto}, ReCo~\cite{reco}, and OpenVE-3M~\cite{openve3m}, it potentially inherits the biases present within these datasets. Therefore, though our model is powerful in background replacement, we urge users to exercise caution when utilizing our data and model to strictly avoid any potential misuse. Furthermore, generating the complete dataset required approximately 7 days on 128 H100 GPUs, a process that could potentially be accelerated by employing more efficient models in the future. As dataset creation is fundamentally a one-time effort, this does not impose a sustained computational burden on the community, unless users specifically intend to synthesize a new dataset of comparable scale.

\section{License}
\label{subsec:appendix_license}

The proposed dataset (\emph{Sparkle}), benchmark (\emph{Sparkle-Bench}), and model (\emph{Kiwi-Sparkle}) are all publicly released under the \textbf{CC-BY-4.0} license. The code is released under the \textbf{Apache-2.0} license. Our source videos come from OpenVE-3M and ultimately trace to OpenSora-Plan~\cite{opensora_plan}, OpenViD-HD~\cite{openvid}, and UltraVideo~\cite{ultravideo}, whose publications discuss diversity, identity, consent, and redistribution risks. We apply NSFW filtering and random human inspection to further mitigate safety risks. Our use of source videos from OpenVE-3M strictly adheres to their original license, and the OpenVE-3M authors retain all original rights to those videos.

\end{document}